\documentclass[10pt,twocolumn,letterpaper]{article}
\usepackage[pagenumbers]{cvpr} 
\usepackage[parfill]{parskip}
\usepackage{multirow}
\usepackage[export]{adjustbox}
\usepackage{array}

\usepackage[dvipsnames]{xcolor}

\definecolor{cvprblue}{rgb}{0.21,0.49,0.74}
\usepackage[pagebackref,breaklinks,colorlinks,citecolor=cvprblue]{hyperref}

\def\paperID{326} 
\def\confName{3DV\xspace}
\def\confYear{2024\xspace}

\title{fMPI: Fast Novel View Synthesis in the Wild with Layered Scene Representations}

\author{Jonas Kohler*\\
GenAI, Meta\\
{\tt\small jonaskohler@meta.com}
\and
Nicolas Griffiths Sanchez*, Luca Cavalli, Catherine Herold\\
Meta Reality Labs\\
\and
Alberto Garcia Garcia\\
Google\\
\and
Albert Pumerola, Ali Thabet\\
GenAI, Meta\\
}

\begin{document}
\maketitle

\begin{abstract}
In this study, we propose two novel input processing paradigms for novel view synthesis (NVS) methods based on layered scene representations that significantly improve their runtime without compromising quality. Our approach identifies and mitigates the two most time-consuming aspects of traditional pipelines: building and processing the so-called plane sweep volume (PSV), which is a high-dimensional tensor of planar re-projections of the input camera views. In particular, we propose processing this tensor in parallel groups for improved compute efficiency as well as super-sampling adjacent input planes to generate denser, and hence more accurate scene representation. The proposed enhancements offer significant flexibility, allowing for a balance between performance and speed, thus making substantial steps toward real-time applications. Furthermore, they are very general in the sense that any PSV-based method can make use of them, including methods that employ multiplane images, multisphere images, and layered depth images. In a comprehensive set of experiments, we demonstrate that our proposed paradigms enable the design of an NVS method that achieves state-of-the-art on public benchmarks while being up to $50x$ faster than existing state-of-the-art methods. It also beats the current forerunner in terms of speed by over $3x$, while achieving significantly better rendering quality.
\end{abstract}

{\let\thefootnote\relax\footnote{{* Equal contribution.}}}
\section{Introduction}\label{sec:intro}
\begin{figure}[t]
  \centering
  \begin{tabular}{ccccc}
    \parbox{0.14\linewidth}{\centering\scriptsize Source views} & 
    \parbox{0.22\linewidth}{\centering\scriptsize grouped PSV} & 
    \parbox{0.02\linewidth}{} & 
    \parbox{0.25\linewidth}{\centering\scriptsize super-sampled \\ MPI} & 
    \parbox{0.17\linewidth}{\centering\scriptsize Target view \\ \& Depthmap}
  \end{tabular}
    \includegraphics[width=\linewidth]{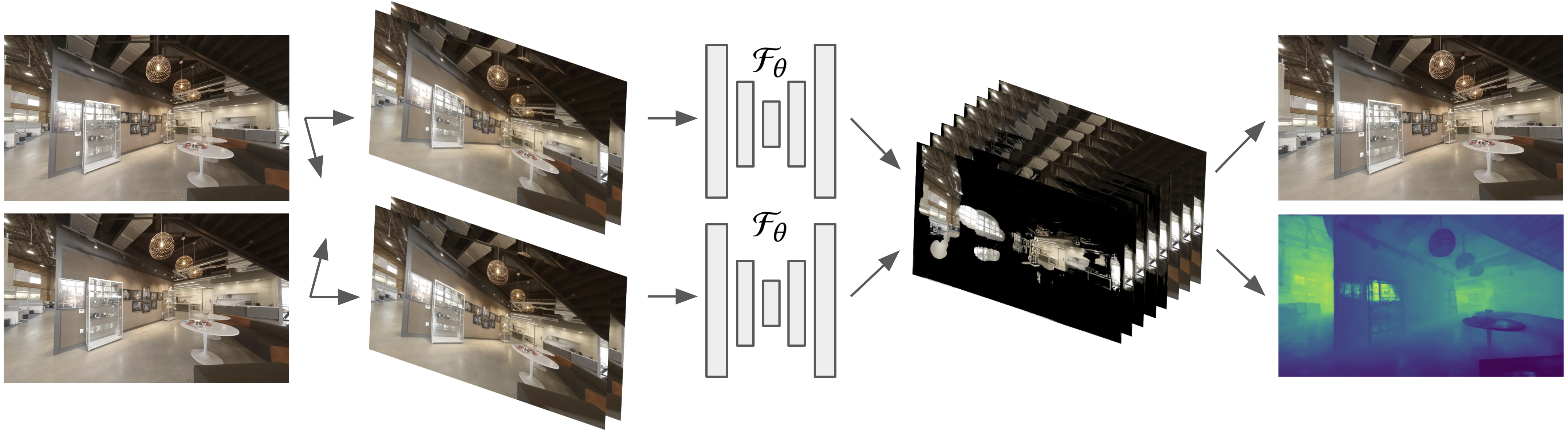}
    \makebox[0.24\linewidth]{\scriptsize DeepView (50sec) }\hfill
    \makebox[0.24\linewidth]{\scriptsize SIMPLI (2.9sec)}\hfill
    \makebox[0.24\linewidth]{\scriptsize fMPI-L (1.1sec)}\hfill
    \makebox[0.24\linewidth]{\scriptsize fMPI-S (0.003sec)}
    \includegraphics[width=0.24\linewidth]{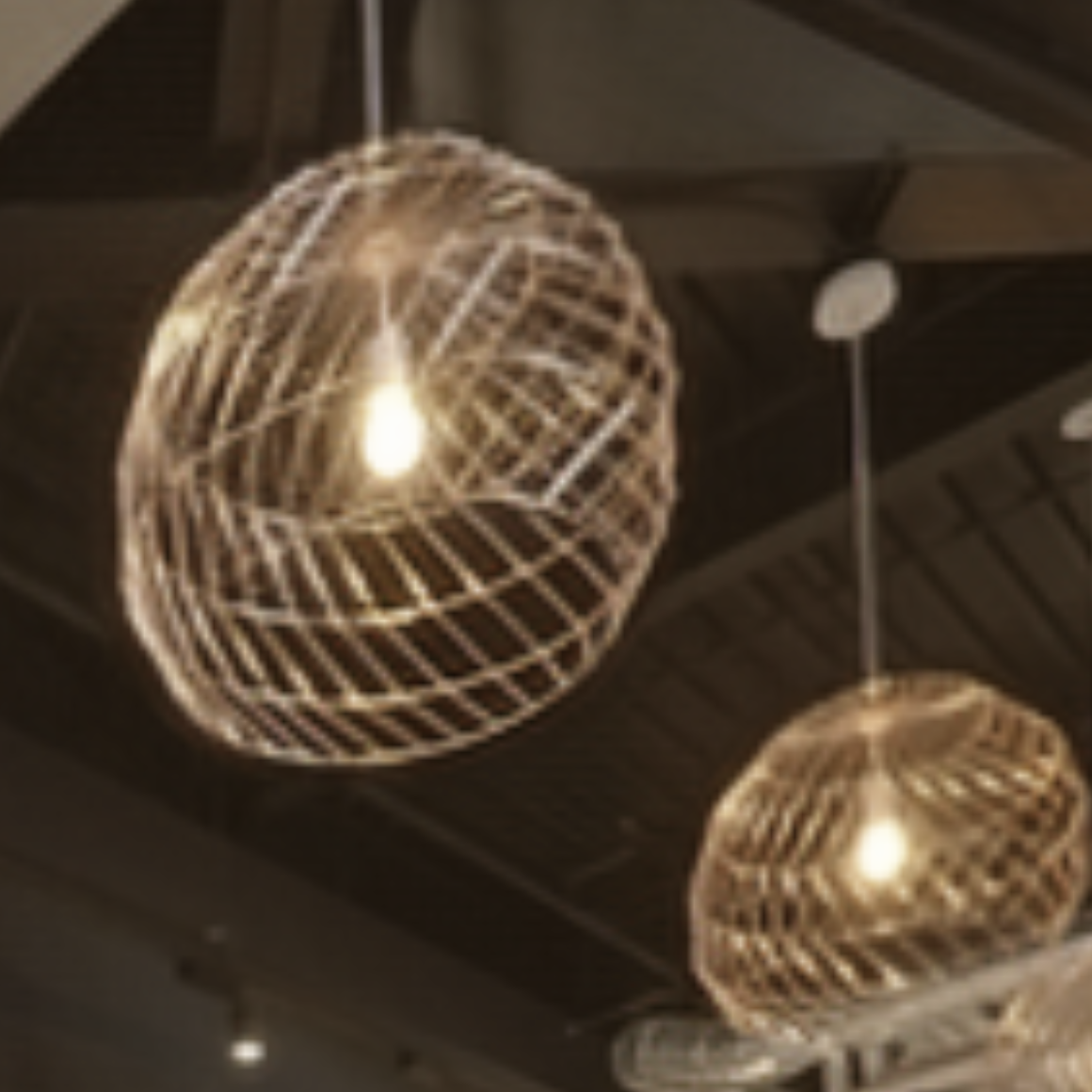}\hfill
    \includegraphics[width=0.24\linewidth]{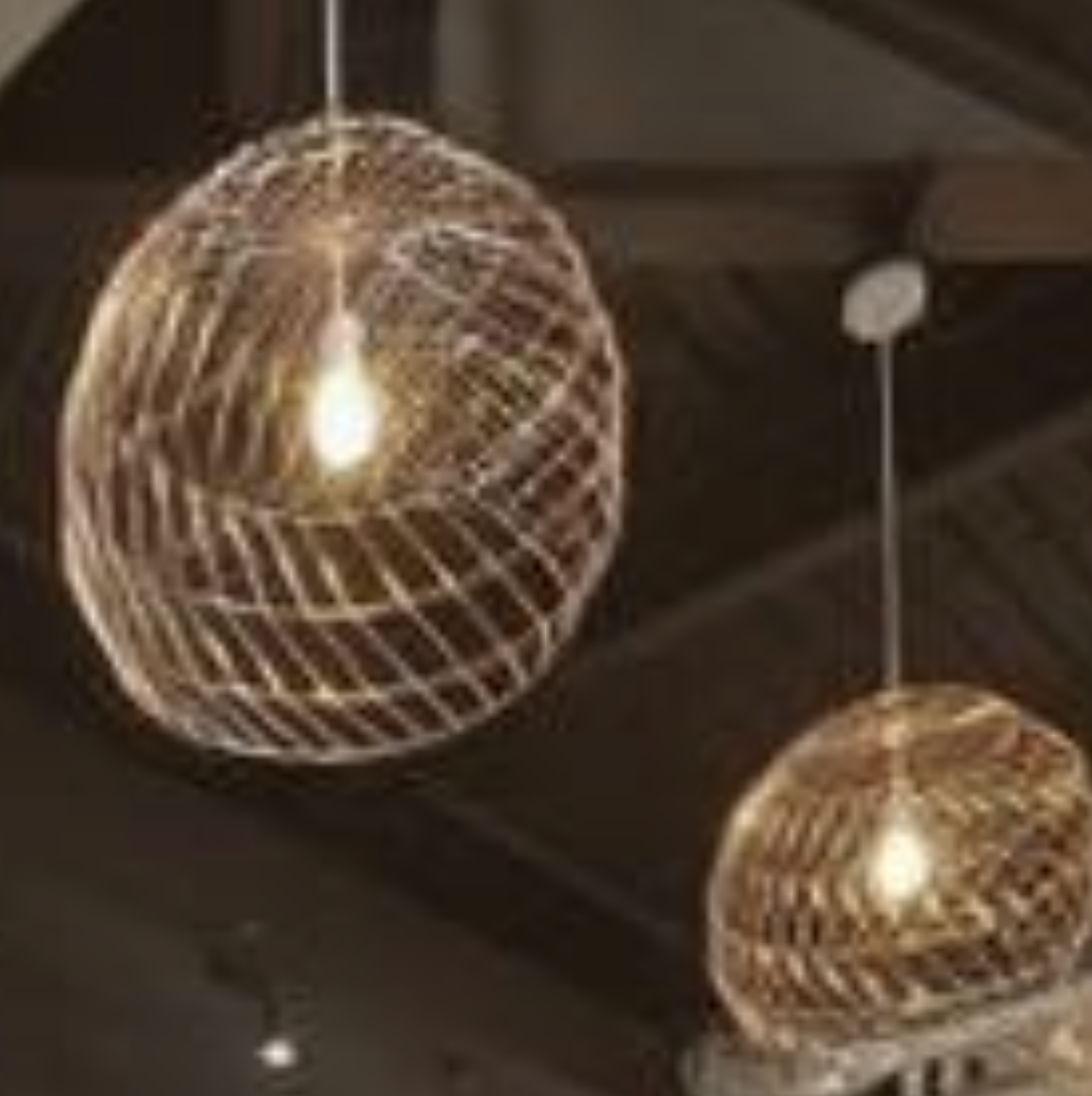}\hfill
    \includegraphics[width=0.24\linewidth]{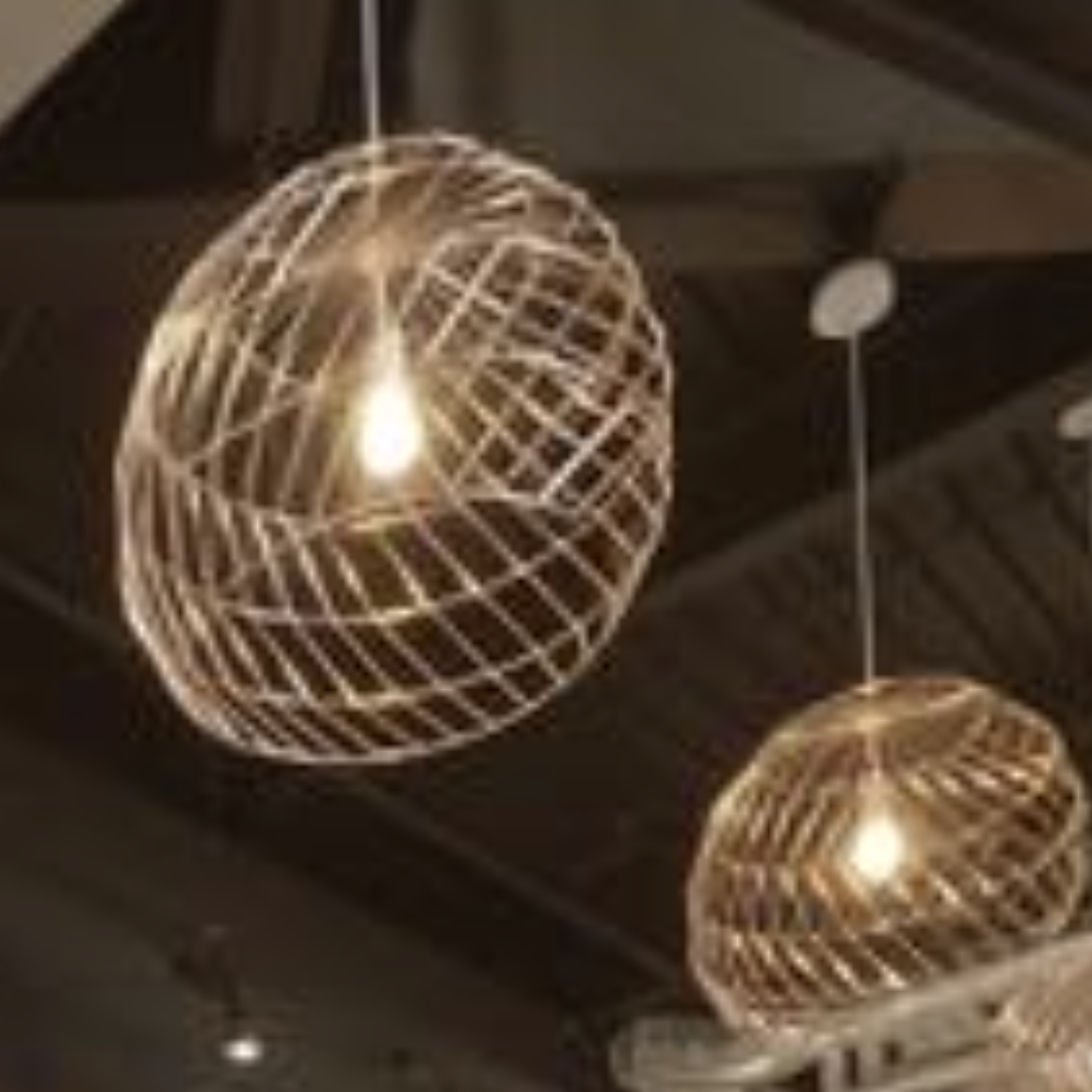}\hfill
    \includegraphics[width=0.24\linewidth]{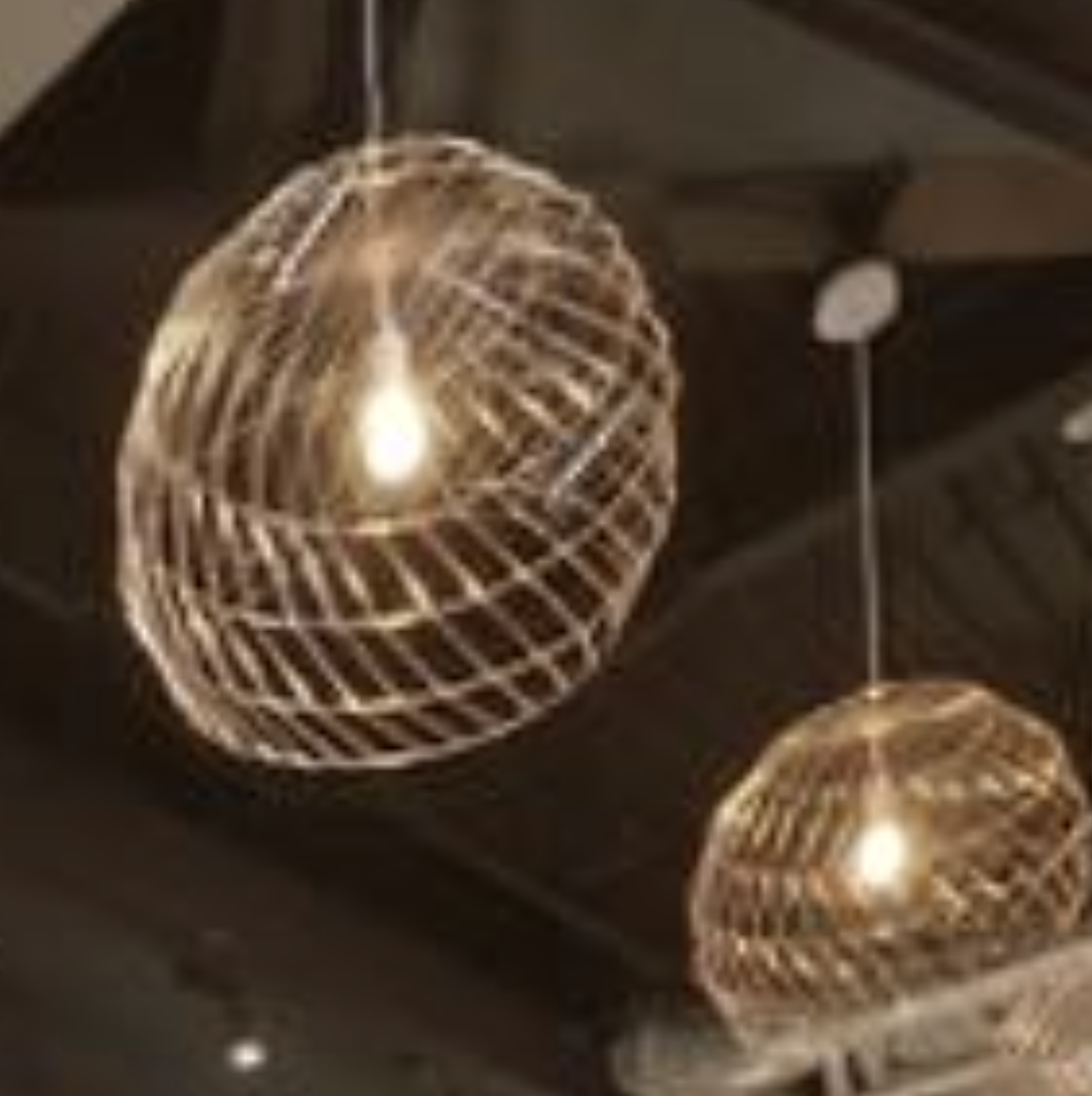}\vspace{-0.05cm}
  \caption{\textbf{Top:} Method Overview. (i) PSV is built by warping input views to planes at different depths. (ii) PSV is split into sets by \textit{grouping} consecutive planes. (iii) Each group is processed \textit{independently} by a neural network, which predicts a \textit{super-sampled} MPI with twice the number of output planes. (iv) Novel views/depthmaps are rendered via over-operation.
    \textbf{Bottom:} Qualitative comparison on the \textit{Spaces} evaluation dataset (resolution $464 \times 800$), as well as runtimes on an NVIDIA A100 GPU. Visual quality remains similar despite the dramatic runtime reductions of fMPI-L and fMPI-S.
    \label{fig:1}}
\end{figure}

Due to the growing prevalence of mobile phones and virtual reality (VR) headsets with stereo cameras, novel view synthesis (NVS) in the wild is becoming increasingly popular. This task presents an interesting challenge for computer vision research, as it requires the generation of high-quality novel views with spatial and temporal consistency in dynamic environments. Its complexity stems from the fact that application environments are often unfamiliar, potentially involving dynamic content, occlusions, and disocclusions as well as varying lighting conditions.

Despite these challenges, substantial progress has been made in recent years, targeting applications such as stereo magnification \cite{zhou2018stereo}, passthrough for VR headsets \cite{xiao2022neuralpassthrough}, interactive 3D photography (e.g, \cite{shih20203d}) and VR videos with 6DoF NVS in dynamic scenes (e.g. \cite{broxton2020immersive}). The most suitable methods for NVS in the wild work on top of multi-layer representations of scenes, such as multiplane images (MPIs) \cite{zhou2018stereo,flynn2019deepview}, 
multisphere images (MSIs) \cite{broxton2020immersive} and layered depth images (LDIs) \cite{shih20203d,khakhulin2022stereo,solovev2023self}. In these methods, novel views are generated from a small set of images, which are transformed into a semitransparent layered representation by a neural network. Contrary to traditional view interpolation techniques with a \textit{single} depth map, layered representations can model complex appearance effects, such as transparency, lightning reflection, and complex structural patterns, even in scenes with high depth complexity. Furthermore, in contrast to neural field methods such as NeRF \cite{martin2021nerf}, these methods are particularly suitable for NVS in the wild because they do not require per-scene optimization and offer real-time rendering. 


However, NVS in dynamic scenes not only requires efficient \textit{rendering} but also on-the-fly \textit{generation} of such layered scene representations. In this regard, all available methods fall short of meeting this requirement due to their high computational complexity for scene understanding. For example, the current state-of-the-art MLI method SIMPLI \cite{solovev2023self} can render in $120$FPS but takes around $3$ seconds to generate an MLI at a relatively low resolution of $480$p. This hinders wider adaptation of layer-based methods in the aforementioned use cases\footnote{For example, passthrough on VR devices (generating realistic views from camera streams at the user's eyes) requires rendering rates above $30$FPS at $1080$p or higher.} and prevents the development of potential future applications like immersive remote control and virtual telepresence via 2.5D or stereoscopic video calling. 

In this paper, we take a step towards real-time NVS in the wild by presenting two contributions that dramatically speed up existing layered-based approaches. Towards this end, we identify the use of so-called plane sweep volumes (PSV) as the main source of computational complexity. Obtaining such volumes - which constitute the neural network input in existing pipelines -  requires multiple re-projections of the source views, which are expensive to obtain due to the need to compute projection matrices, map source to target coordinates, as well as to sample the source view multiple times. Furthermore, large PSVs necessitate the use of either 2D convolutional neural networks with a large number of channels or networks with three-dimensional convolutional kernels (e.g. \citep{zhou2018stereo,flynn2019deepview}), both of which have high computational cost. 
To address this issue, we introduce two novel paradigms for generating and processing plane sweep volumes, enabling significantly faster view synthesis times without compromising quality:
\begin{itemize}
    \item Plane grouping: We propose breaking the PSV into several disjunct groups which are processed in parallel. This allows for employing small (and fast) neural network backbones without introducing information bottlenecks. The number of groups presents a flexible hyper-parameter that adjusts the per plane compute budget, hence trading speed and performance.
    \item Super-sampling: We furthermore show that neural networks are able to leverage redundancies in the PSV to improve the performance of MPI methods without added computational costs. In particular, a sparse PSV can be super-sampled into a dense MPI by giving up on the usual one-to-one plane correspondence, which significantly reduces runtimes as creating PSV planes is an expensive process.
\end{itemize}

Finally, building on these two contributions, we set a new reference runtime for NVS in the wild with MPI-based methods. Through an extensive set of experiments, we demonstrate its ability to achieve state-of-the-art quality, while being $50x$ faster than the best existing MPI, and $10x$ faster than the best MLI method to date.
\section{Related Work}\label{sec:related}

Novel View Synthesis (NVS) is a long-standing problem at the intersection of computer graphics and computer vision with seminal works in the field dating back to the 1990s. Early methods focused on interpolating between corresponding pixels from input images \citep{chen1993view,mcmillan1995plenoptic,seitz1996view,szeliski1994image} or between rays in space \citep{gortler1996lumigraph,levoy1996light}.
In recent years, machine learning-based methods  have driven significant advancements in both the quality of rendered images and the range of possible new views. This progress has been primarily driven by huge developments in the field of deep learning, alongside the emergence of differentiable rendering methods, which allow for 3D scene understanding based solely on 2D supervision. In the following paragraphs, we summarize the two most common approaches for learned NVS.


\paragraph{Explicit, layered scene representations}
Multi-plane images (MPIs) present a perspective variant of volumetric representations (such as 3D voxel grids) that, instead of using a uniformly sampled volume, positions a layered scene representation in the view frustum corresponding to one of the input (or target) cameras with fronto-parallel planes along the optical axis (usually placed at uniform disparities) \cite{szeliski1999stereo,zhou2018stereo}. Traditionally, each plane element contains color and opacity ($\alpha$) values, which allows for fast and differentiable rendering with standard alpha compositing \cite{porter1984compositing}. Furthermore, the soft alpha enables high-quality view synthesis with smooth transitions, semi-transparencies, realistic reflections, and specular highlights. Moreover, MPIs can model occluded content for a range of views that grows linearly in the number of planes \cite{srinivasan2019pushing}.
Importantly, MPI-based methods do not require per-scene optimization and thus lend themselves to NVS in the wild.

Most notable works in this line of research include the seminal paper on stereo magnification \cite{zhou2018stereo}, the theoretical analysis in \cite{srinivasan2019pushing}, the local MPI fusion proposed in \cite{mildenhall2019local}, the extension to spherical representations in \cite{broxton2020immersive}, the adaptive plane placement proposed in \cite{navarro2022deep}, as well as the current state-of-the-art method, DeepView \cite{flynn2019deepview}, which models MPI learning as an inverse problem, solved by a sophisticated variant of learned gradient descent.\footnote{For completeness, we note that there is a growing line of work generating MPIs from single image inputs but they typically achieve inferior quality compared to the multi-camera settings mentioned in Section \ref{sec:intro} (see
e.g. \citep{tucker2020single,han2022single}).} In a concurrent line of research, several methods propose to enrich the rigid MPI plane representation by adding per-entry depth values, which results in deformable layers that can adapt to the scene geometry. This approach, referred to as layered depth images (LDI) or multilayer images (MLI) goes back to \citep{shade1998layered} and has recently been proposed for in the wild NVS \citep{khakhulin2022stereo,solovev2023self}. MLIs promise more compact scene representations, but as we point out in Section \ref{sec:method}, the main bottleneck of any layer-based method does not lie in rendering from the intermediate representation but in generating it in the first place. Here, the two lines of research do not differ.
\begin{figure*}[ht]
    \centering
    \includegraphics[width=0.99\textwidth]{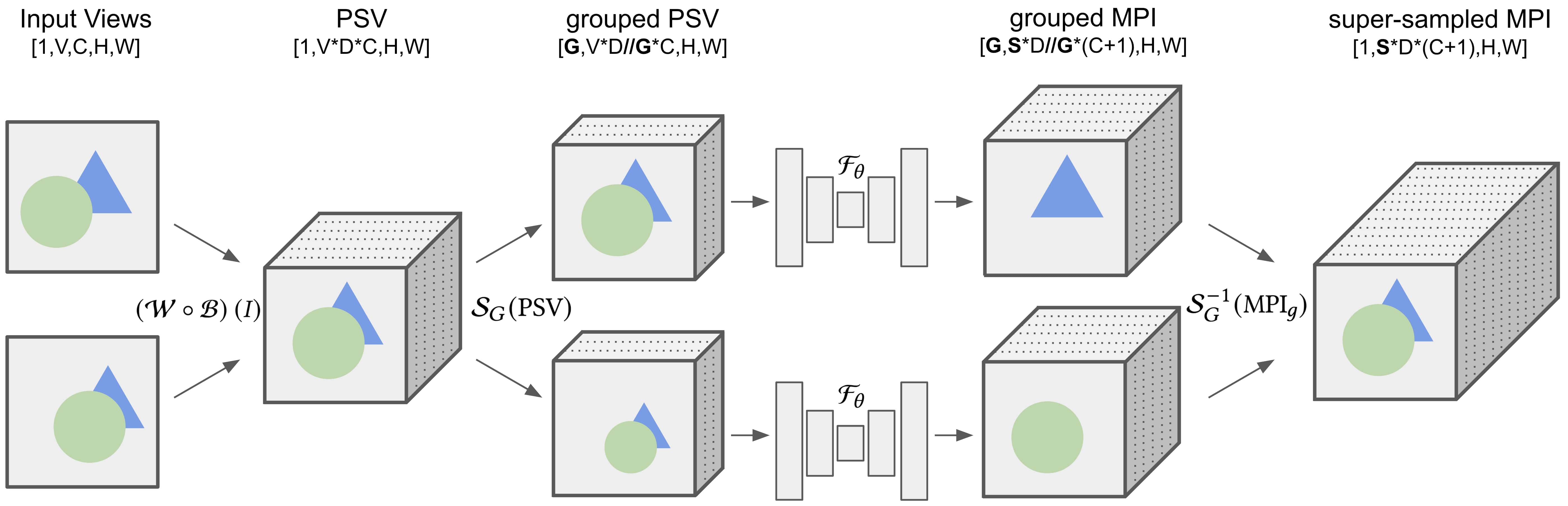}
    \caption{\textbf{Efficient MPI generation} Given a set of input views, our method constructs a plane sweep volume (PSV) with $D$ input planes, which are grouped into $G$ sets of size $D/G$ (assuming $D\mod~G =0$). Each set is then processed \textit{independently} by a fully convolutional neural network $\mathcal{F}_\theta$ with shared parameters $\theta$. The grouping allows for an optimal speed-performance trade-off, considering only \textit{local} cross-plane dependencies. Furthermore, the network super-samples its input tensor by a factor of $S$, predicting an MPI with $S \cdot D$ planes and thereby saving $\mathcal{O}((S-1)\cdot D)$ computations when generating the PSV.}
    \label{fig:overview}
\end{figure*}
All of the above methods perform planar reprojections of the input views assuming different depth values to create a so-called Plane Sweep Volume (PSV), which allows the network to process any given scene independently of the specific camera calibrations and relative rotations/translations between the input cameras. Since the PSV usually consists of a double-digit number of planes, its creation is time-consuming and furthermore, its magnitude necessitates the use of either 2D convolutional neural networks with high numbers of channels (e.g. \citep{zhou2018stereo,flynn2019deepview}) or networks with slow three-dimensional convolutional kernels (e.g. \cite{mildenhall2019local,srinivasan2019pushing}). Due to this computational constraint, current techniques face difficulties in producing and rendering layered scene representations in real time. In Section \ref{sec:method}, we put forth two adaptations to the above-mentioned conventional 
methods that significantly reduce their runtime requirements.\looseness=-1
\paragraph{Implicit scene representations}
Explicit representations often have large memory footprints and poor scalability with image resolution. An alternative approach is to implicitly store light fields as functions of spatial location and viewing angle. Neural scene representations, such as NeRF \cite{mildenhall2021nerf}, optimize continuous scene representations using MLPs, sometimes combined with sparse explicit representations (e.g~\cite{mueller2022instant,cao2023hexplane}). These methods excel in view-dependent effects, non-Lambertian surfaces, and complex objects but are traditionally limited to static scenes, per-scene optimization, and slow rendering due to numerous MLP evaluations per pixel.\looseness=-1

Recent developments have led to faster training \cite{muller2022instant}, increased rendering frame rate \cite{garbin2021fastnerf,reiser2023merf,yariv2023bakedsdf,chen2022mobilenerf,rojas2023re}, dynamic scene modelling \cite{pumarola2021d,cao2023hexplane}, and reduced input camera requirements \cite{muller2022autorf}. Some techniques avoid per-scene optimization \citep{chen2021mvsnerf,jain2021putting,martin2021nerf,yu2021pixelnerf} by learning shared priors. For example, IBRNet \cite{wang2021ibrnet} learns a generic view interpolation function that generalizes to unseen scenes. However, this method still requires expensive ray sampling and achieves the best performance only with scene-specific fine-tuning.

\paragraph{Further related works}
NeX \cite{wizadwongsa2021nex} proposes a hybrid of MPI and neural radiance fields, resulting in high-quality renderings in real time. Yet, this method requires a high number of input images as well as a lengthy per-scene optimization of the applied basis functions and thus is not suitable for in-the-wild NVS. A distinct solution explicitly developed for NVS on VR headsets is NeuralPassthrough \cite{xiao2022neuralpassthrough}. This approach utilizes simple image-based rendering based on forward warping with learned stereo-depth. As a result, it provides fast view synthesis but cannot effectively model reflections and semi-transparencies. Furthermore, it cannot inpaint disoccluded regions, opting instead to fill these disoccluded pixels with the smoothed colors of the local vicinity. Clearly, this limits the applicability of \cite{xiao2022neuralpassthrough} to use cases with very small camera offsets of just a few centimeters.

\section{Method}\label{sec:method}

\subsection{Setting}
MPI methods usually generate novel views by transforming input camera views into an intermediate, layered scene representation from which new perspectives can be rendered. In particular, given a set of $V$ input views $\{I_v\}_{v=1}^V$, where each image $I_v\in\mathbb{R}^{3,H,W}$, layered scene representations are generally obtained by first broadcasting the input views $D$ times along a new depth dimension $\mathcal{B}:\mathbb{R}^{3,H,W}\rightarrow\mathbb{R}^{D,3,H,W}$ and then using inverse homography (details in Sect. \ref{sec:homography}) to warp the repeated input images to the MPI camera $\mathcal{W}:\mathbb{R}^{D,3,H,W}\rightarrow\mathbb{R}^{D,3,H,W}$ assuming planar projections spaced linearly in disparity. The resulting tensor, named PSV, aligns at every disparity level all regions from all input images whose rays originated from the same corresponding disparity. This property allows scene understanding by patch comparison, a task for which convolutional neural networks are a natural fit.

Thus, the PSV is processed by a mapping $\mathcal{F}_\theta:\mathbb{R}^{B,D,V,3,H,W}\rightarrow\mathbb{R}^{B,D,4,H,W}$ (where $B$ constitutes the batch dimension) parameterized with learnable parameters $\theta$, which predicts a volumetric scene representation consisting of fronto-parallel $RGB\alpha$ images (MPIs) or deformable layers of $RGBd\alpha$ \cite{khakhulin2022stereo}. In the MPI case, novel views can be rendered via over-operation \cite{porter1984compositing}:

\begin{equation}\label{eq:overop}
I_{\text{target}} = \sum_{d=1}^D \left(\text{RGB}_{d}\alpha_d \prod_{j=0}^{d-1}\left(1-\alpha_j\right)\right),
\end{equation}
where $\text{RGB}_{d}$ and $\alpha_d$ are the color and opacity values of the MPI at layer $d$. The term $\prod_{j=0}^{d-1}\left(1-\alpha_j\right)$ constitutes the net transmittance of layer $d$. This operation is efficient and fully differentiable, which enables learning $\mathcal{F}_\theta$ purely based on image supervision.

Existing methods differ in where exactly the MPI camera is placed and how the mapping $\mathcal{F}_\theta$ is constructed and learned. However, what is common to all methods is the rather low inference speed due to the need to process the highly redundant PSV with a large neural network. The present method presents two novel input processing paradigms for improving this very aspect. An overview is presented in Figure \ref{fig:1}. In what follows, we focus on \textit{target}-centered MPI images (as in \cite{flynn2016deepstereo}), which we identify as an adequate setting for dynamic NVS in the wild. Nonetheless, as detailed in Section \ref{sec:ablations} our contributions also work in the setting of static, warped MPIs. In general, we highlight that our contributions can be integrated into any method that relies on processing PSVs, encompassing MLI methods. Finally, it is also conceivable to extend our approach to stereo-depth methods based on cost volumes (e.g. \cite{gu2020cascade}).

\subsection{Methodology}
We propose two novel paradigms for processing PSV tensors that leverage the fragmented nature of MPI-based NVS as well as redundancies in the PSV commonly used as input. Our approaches overcome the runtime limitations of existing methods and enable real-time generation of novel views in the wild. An overview of our method is presented in Figure \ref{fig:overview}.

Recall that the PSV is of shape $[B,D,V,3,H,W]$, where usually $B=1$ during inference and $B>1$ during training. There are two obvious ways of processing the PSV: (a) Most methods merge depth planes into the $RGB$ channel dimension, which results in network inputs with triple-digit channels ($3\cdot D\cdot V$) as typical values of $D$ range between $32$ \cite{ghosh2021liveview} and $80$ \cite{flynn2019deepview}. This necessitates the use of large neural networks to prevent the information bottleneck that would arise from the large input PSV tensor (for example, channels in the network employed the seminal works of \cite{zhou2018stereo} range from $99$ in the first layer to $1024$ in the last layer).
(b) The works of \cite{flynn2016deepstereo, flynn2019deepview} and \cite{ghosh2021liveview} (also in \cite{han2022single} for single image MPIs) present notable exceptions. There, depth planes are processed independently by merging the input view dimension $V$ into the $RGB$ channel dimension, but the depth dimension $D$ is instead merged with the batch dimension $B$. In this way, the number of input channels to the network is drastically reduced to ($3\cdot V$), which allows for much smaller networks. Yet, this process is inherently slow because the network needs to be queried $D$ times. Furthermore, when processing each depth plane independently, the network cannot access cross-depth information. This can decrease the consistency between the predicted MPI planes, which in turn may lead to aliasing artifacts \cite{srinivasan2019pushing}.

We show that the best performance-latency trade-off is achieved by processing depth planes in groups, as opposed to (a) all as one PSV or (b) independently plane-by-plane. Interestingly, we find that modeling only interactions of neighboring planes (e.g. planes within a group) is sufficient to achieve state-of-the-art performance while drastically reducing the computation time. Furthermore, inspired by results from single view novel-view synthesis with MPIs such as \cite{han2022single}, we demonstrate that performance can be further improved by letting $\mathcal{F}_\theta$ super-sample the PSV into MPIs with larger numbers of depth planes, which is significantly more efficient than increasing both input (PSV) and output (MPI) planes alike. The following two paragraphs present our contributions in more detail.

\subsubsection{Plane grouping}

\begin{figure}[h!]
    \centering
    \includegraphics[width=0.4\textwidth]{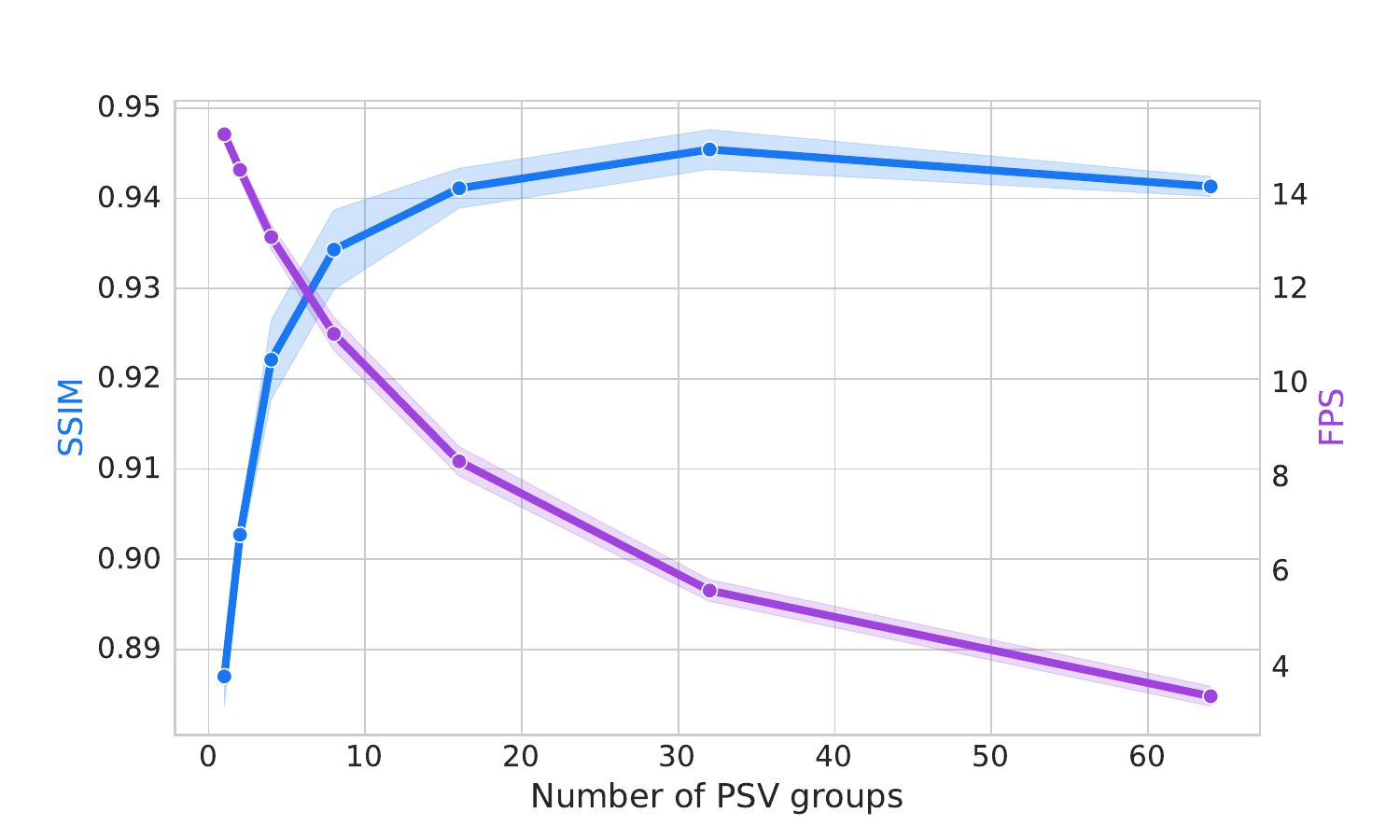}
    \caption{\textbf{Plane grouping}: SSIM and FPS (mean and standard deviation of five independently trained models for each x-axis value) on Spaces over the number of PSV groups (i.e. number of forward passes) given $64$ input planes. Performance increases with the number of groups as the information bottleneck in early layers decreases and more computation is invested per plane. Simultaneously, the inference speed drops since the number of forward passes increases with the number of groups. The simple four-level U-Net from Table \ref{table:u-net} was used as backbone.} \label{fig:pgrouping}
\end{figure}

Given the PSV computed as $\mathcal{W}(\mathcal{B}(\{I_v\}_{v=1}^{V}))$, we obtain the input for our neural network $\mathcal{F}_\theta$ by dividing the PSV tensor along the depth dimension into $G$ sets of $D/G$ planes, assuming $D \mod G=0$. Each set is then considered as a sample in the batch dimension and the depth, view, and color channels are merged (as in other MPI methods), resulting in the transformation $\mathcal{G}:\mathbb{R}^{B,D,V,3,H,W}\rightarrow\mathbb{R}^{B\cdot G,D/G\cdot V\cdot3,H,W}$. Compared to approach (a) introduced above, this reduces the information bottleneck in the first layer by a factor of $G$. At the same time, it also reduces the number of forward passes of (b) by a factor of $G$.

This is illustrated in Figure \ref{fig:pgrouping}. As can be seen, the rendering speed decreases in the number of groups as expected. At the same time, performance increases with $G$ to the point of having $32$ groups (with two planes each). Compared with lower values of $G$, this indicates that more compute on local context is better than having more global context but less compute per plane. At the same time, comparing to $G=64$ shows that having at least some local context is better than processing planes completely independently. Notably, most existing approaches are either (a) on the far left (e.g.~\cite{zhou2018stereo,broxton2020immersive,solovev2023self,navarro2022deep}) or (b) on the far right (e.g .~\cite{flynn2016deepstereo,ghosh2021liveview,flynn2019deepview}). Our results show that both approaches are sub-optimal. In fact, the best performance-latency trade-off lies between these two extremes.\footnote{Note that the specific intersection depicted here does not indicate an optimum as it depends on the scaling of the y-axes.} Importantly, our plane grouping approach allows for considerable flexibility, as the specific selection of $G$ enables a simple adjustment to meet runtime and quality requirements, even dynamically at inference time.

\subsubsection{Plane super-sampling}
The performance of MPI methods is generally correlated with the number of MPI planes used for rendering \cite{flynn2019deepview, srinivasan2019pushing}. Traditionally, increasing the number of input planes in the PSV and output MPI planes has been done with one-to-one correspondence. However, this approach results in higher computational costs as the time required to generate the PSV is proportional to the number of depth planes as $\Omega(D)$. Taking inspiration from single view MPI methods (e.g. \cite{tucker2020single,han2022single}), we here show that such strict correspondence is actually not necessary. Instead, redundancies in the PSV can be leveraged by the neural network to generate better results with super-sampled MPIs.

\begin{figure}[h]
    \centering
    \includegraphics[width=0.4\textwidth]{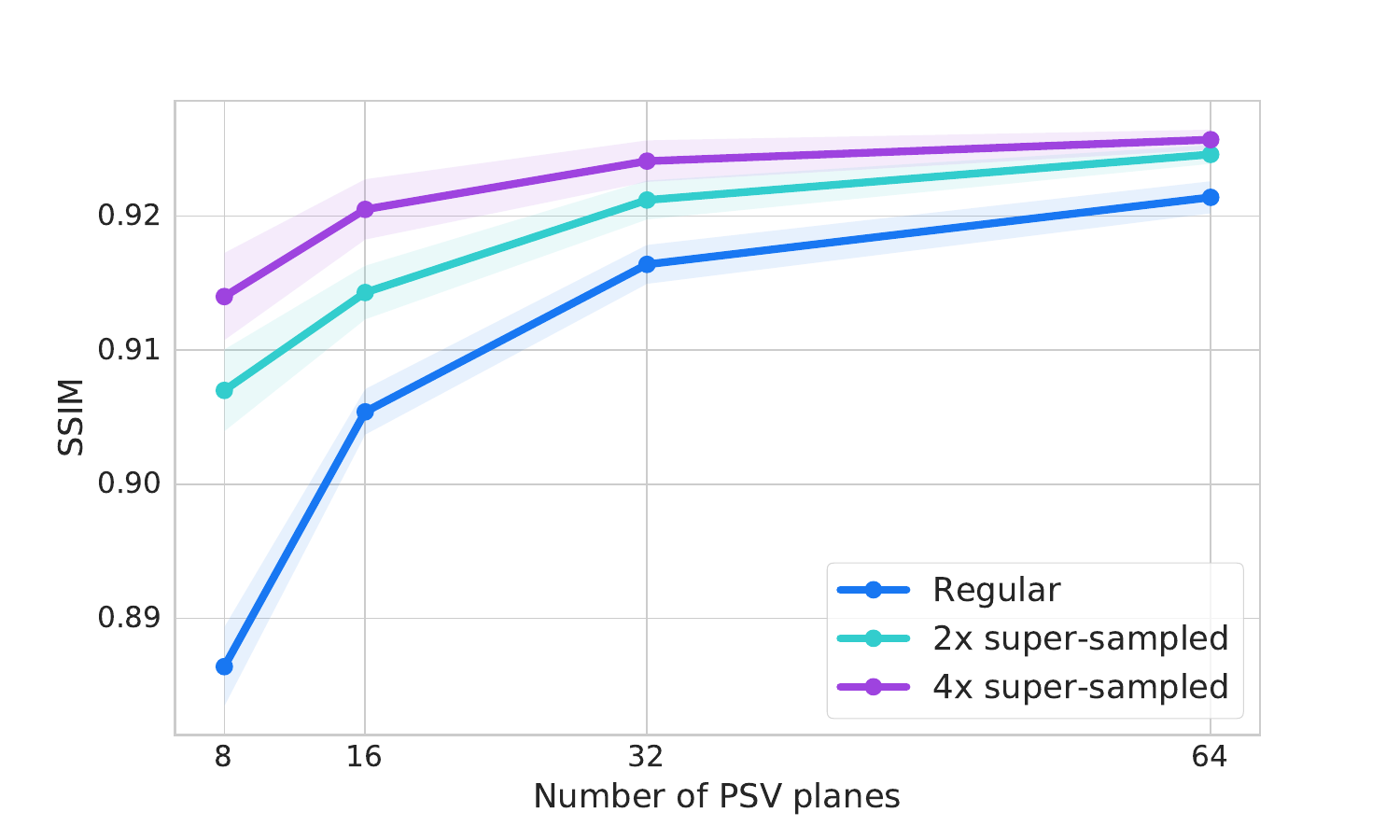}
    \caption{\textbf{Plane super-sampling}: SSIM (mean and standard deviation of five independently trained models for each mark) on Spaces over number of PSV (i.e. input) planes for the regular approach (where number of PSV planes equals number of MPI planes) and the super-sampling approach with factors two and four. Super-sampling clearly improves performance and is almost on par with regular models of corresponding MPI density. A simple four-level U-Net network (Tbl.~\ref{table:u-net}) was used as backbone.}
    \label{fig:psampling}
\end{figure}

As can be seen, the network $\mathcal{F}_\theta$ can predict super-sampled MPIs of almost equal quality as the standard one. In particular, the performance of MPI methods can be enhanced at marginal computational cost\footnote{Only the number of output channels of the last layer of the neural network changes.} by letting the network $\mathcal{F}_\theta$ leverage redundancies in the PSV to predict super-sampled MPIs. As can be seen in Figure \ref{fig:psampling}, this approach yields significantly better results than the regular method (compare lines \textit{vertically}), while achieving comparable performance to regular models with equivalent numbers of MPI planes (compare lines \textit{horizontally}). This result is significant given the elevated computational cost of generating PSV planes. For instance, generating a PSV of $4$ input cameras at $464 \times 800$ takes $24.21$ms for $32$, $43.6$ms for $64$ and $85.25$ms for 128 depth planes on an A100 GPU (see Figure \ref{fig:psv_timings} for more details)

\subsection{Training Details}\label{sec:training_details}
In order to show the above-mentioned flexibility of our approach, we design and train three different models, each parameterizing $\mathcal{F}_\theta$ with a fully convolutional encoder-decoder network. We consider two backbones: (i) the simple 4-level U-Net architecture presented in \cite{ghosh2021liveview} (details in Table \ref{table:u-net}) and (ii) a variant of the former where we replace the blocks of each level with the corresponding blocks of the ImageNet pre-trained ConvNeXt tiny \citep{liu2022convnet}. Given these two backbones, we design the following three versions of our method, which we term \textit{fast MPI} (or fMPI for short):

\begin{itemize}
    \item fMPI-S: U-Net backbone, $D=16$, $G=4$, $S=2$
    \item fMPI-M: U-Net backbone, $D=32$, $G=16$, $S=2$
    \item fMPI-L: ConvNext backbone, $D=40$, $G=20$, $S=2$.
\end{itemize}
For instance, the M model receives a PSV of $D=32$ planes as input. It processes this input in $G=16$ parallel forward passes, each taking $D/G=2$ input planes and predicting $S\cdot D/G=4$ output planes. In total, the model hence outputs an MPI of $S\cdot D=64$ planes.

Following \cite{zhou2018stereo}, we do not directly predict an RGB value per MPI plane pixel but opt to take the input views as an effective prior. In particular, the network outputs only a single RGB background image per plane group alongside plane-specific view weights (one for each input camera and for the background image). The final MPI RGB value is then a softmax weighted combination of the background image and the input image colors from the corresponding pixels in the PSV (see Eq.~\ref{eq:RGB_app}).

\newcommand{\bigImgWidth}{0.2865}
\newcommand{\smallImgWidth}{0.14}

\begin{figure*}[ht]
    \centering
    \setlength{\tabcolsep}{1.2pt}

    \begin{tabular}{c c c c c c}
        \multicolumn{1}{c}{\textbf{}} & \multicolumn{1}{c}{SIMPLI} & \multicolumn{1}{c}{DeepView} & \multicolumn{1}{c}{fMPI-M} & \multicolumn{1}{c}{fMPI-L} & \multicolumn{1}{c}{Ground truth} \\
        \multirow{2}{*}[6em]{
            \includegraphics[width=\bigImgWidth\linewidth, trim=1 1 1 1, clip]{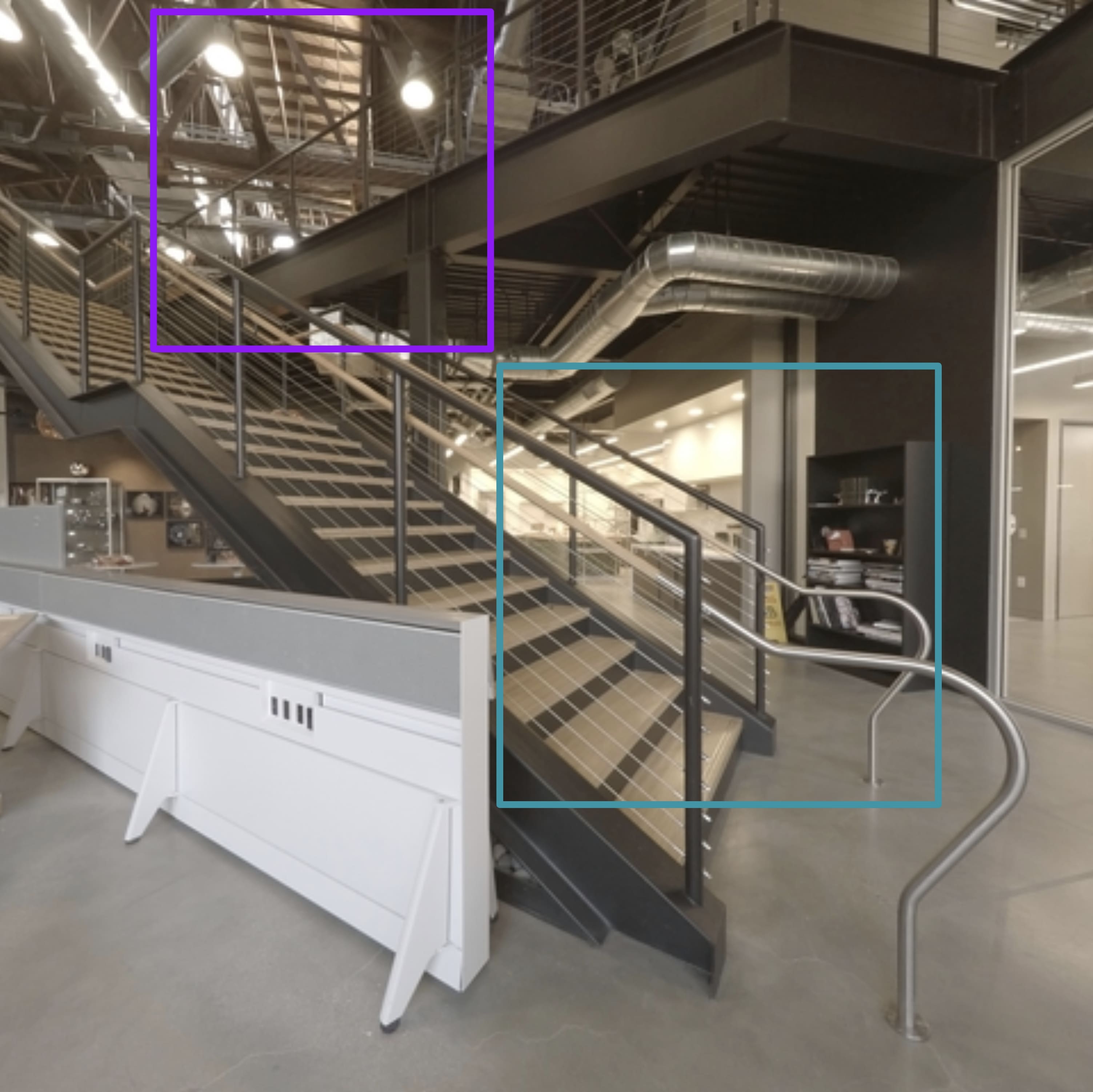}
        } &
        \includegraphics[width=\smallImgWidth\linewidth, trim=1 1 1 1, clip]{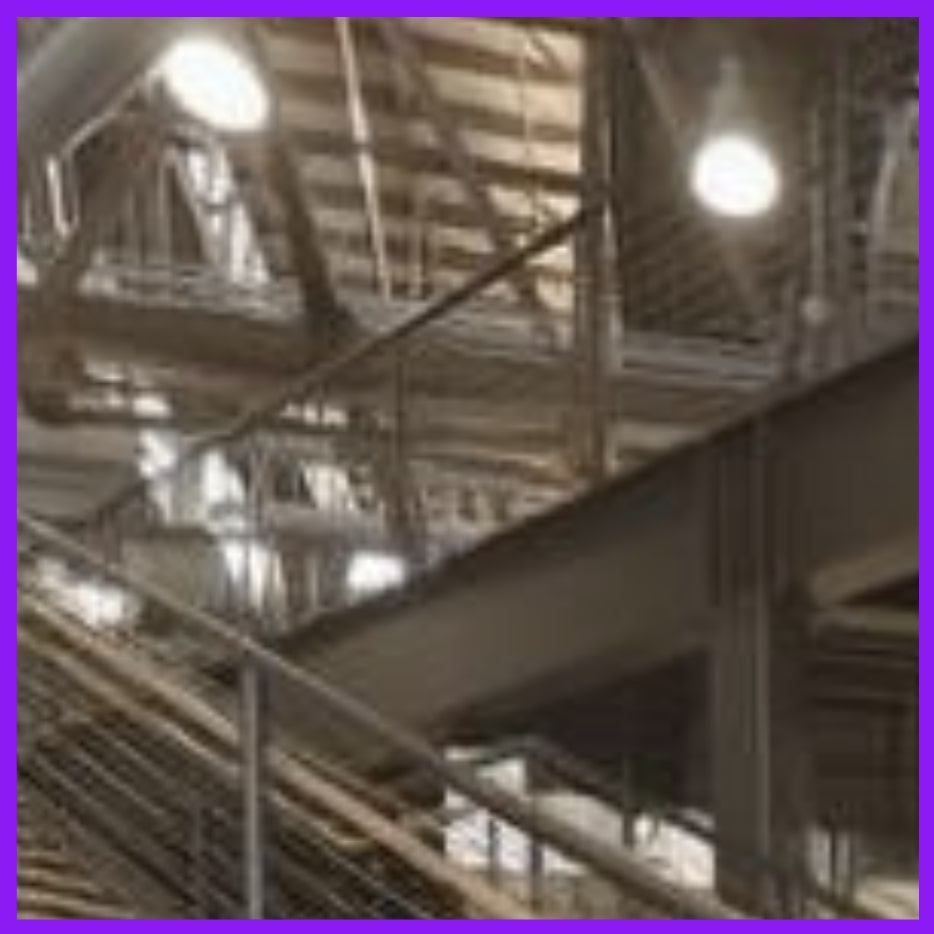} &
        \includegraphics[width=\smallImgWidth\linewidth, trim=1 1 1 1, clip]{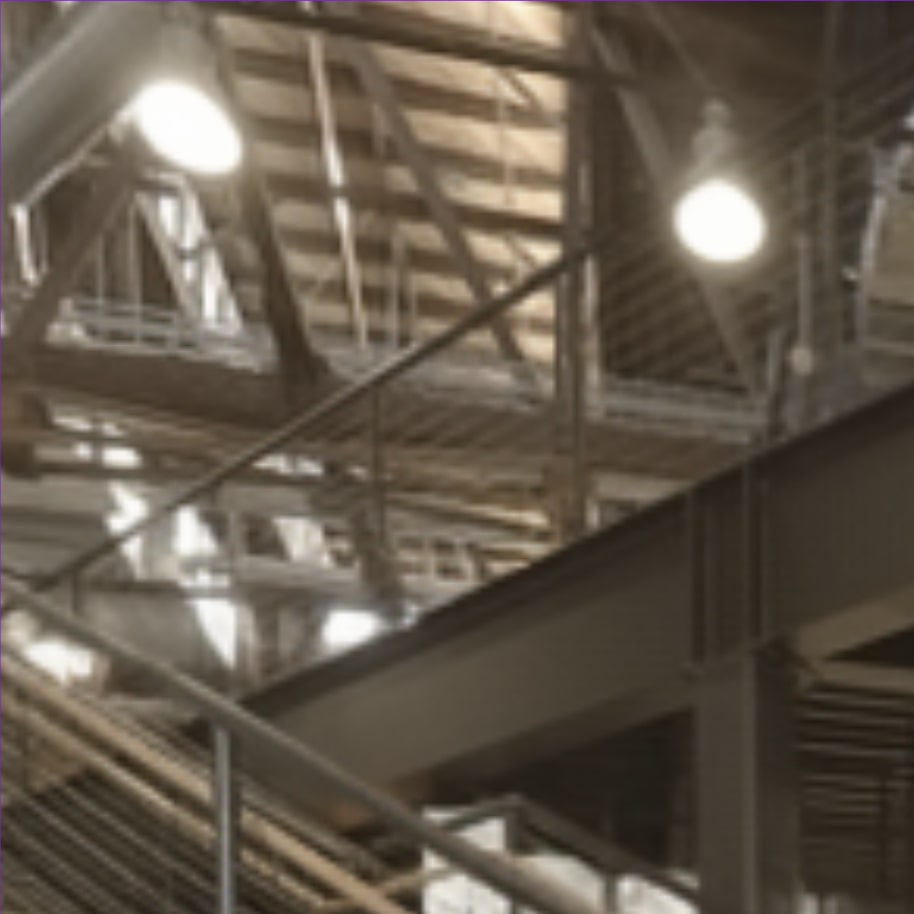} &
        \includegraphics[width=\smallImgWidth\linewidth, trim=1 1 1 1, clip]{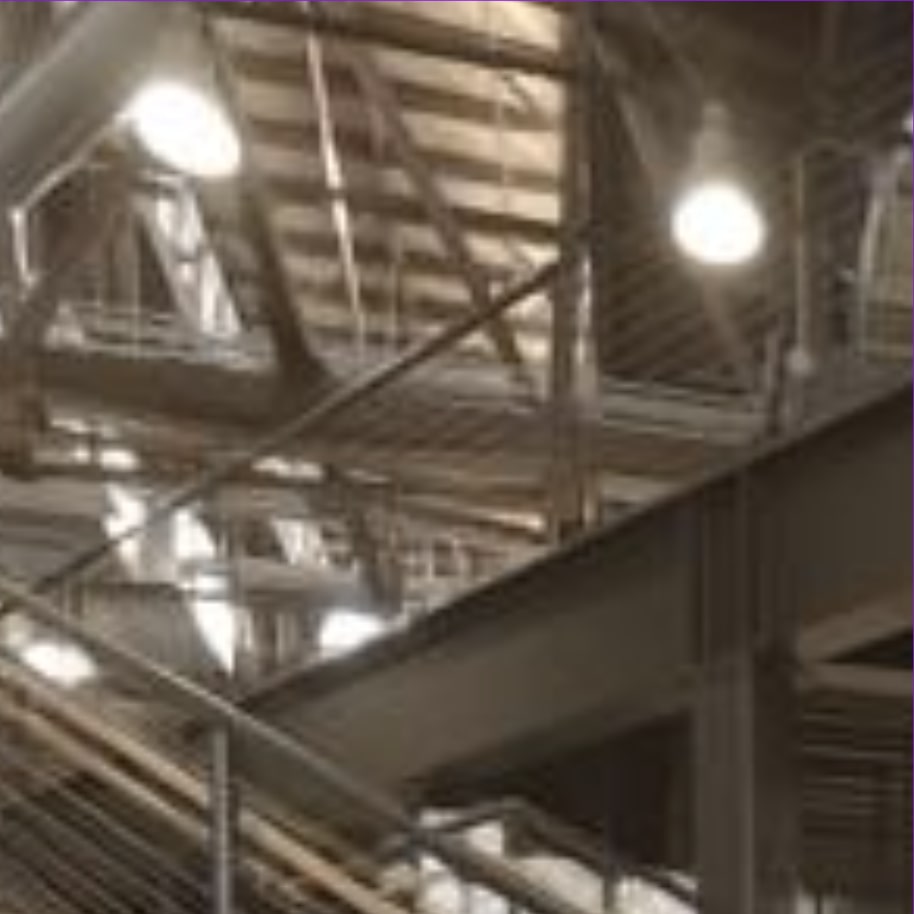} &
        \includegraphics[width=\smallImgWidth\linewidth, trim=1 1 1 1, clip]{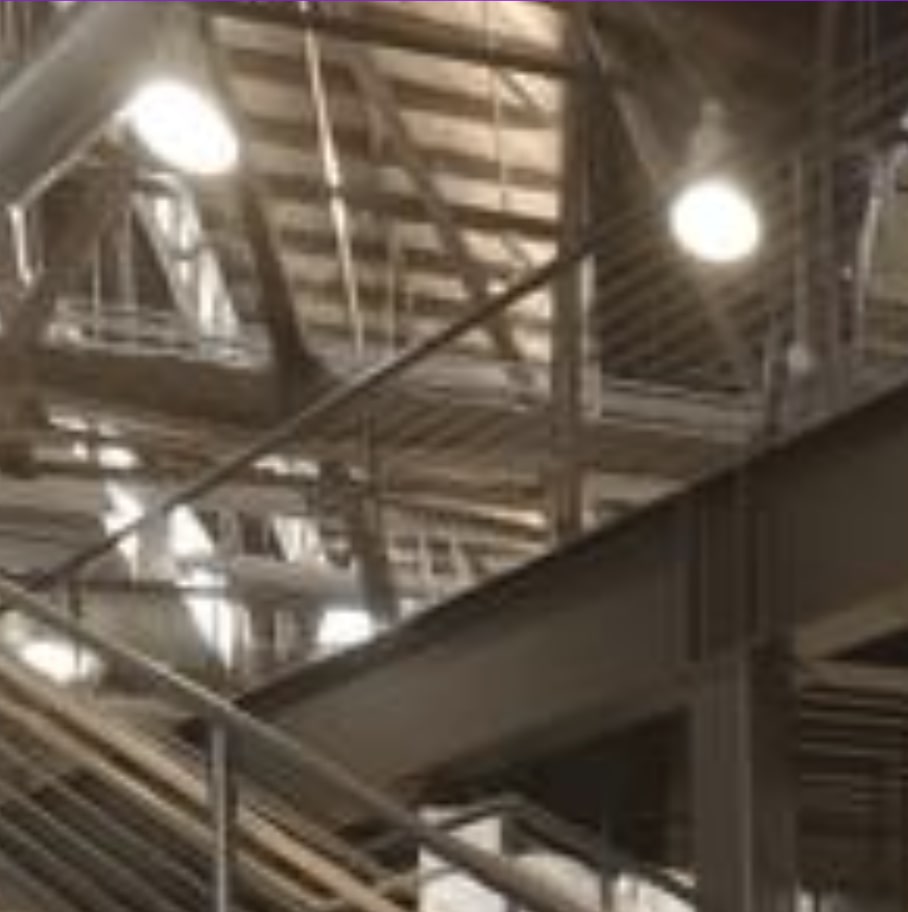} &
        \includegraphics[width=\smallImgWidth\linewidth, trim=1 1 1 1, clip]{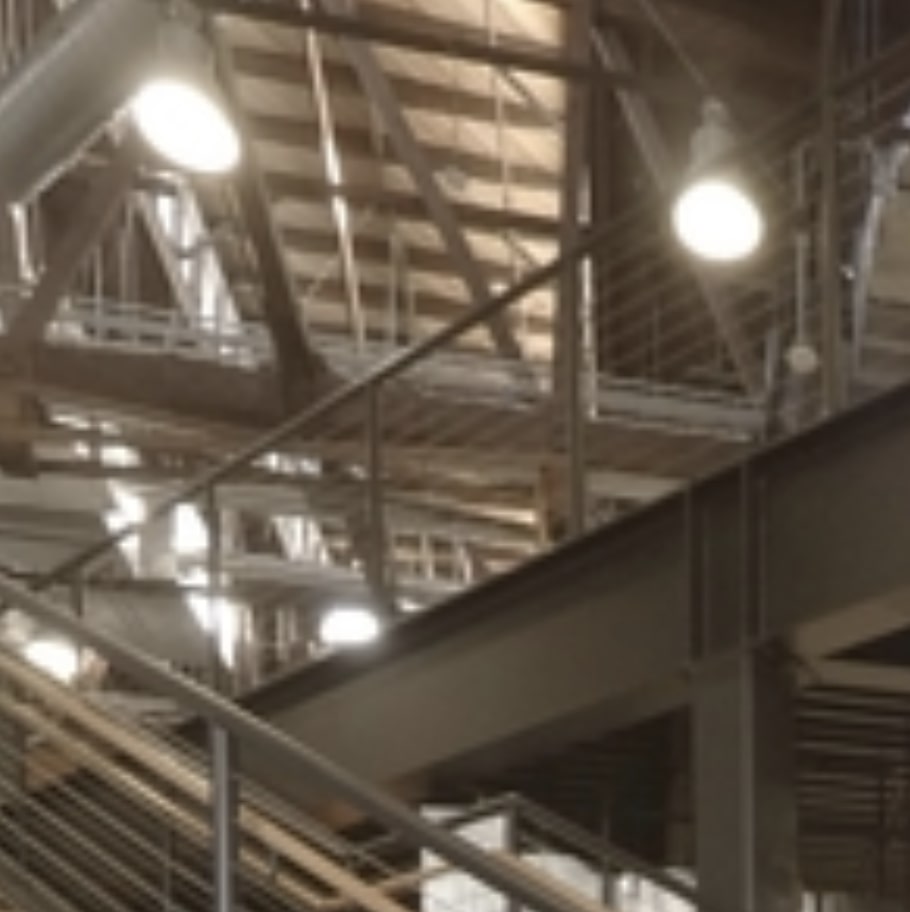}  \\
        &
        \includegraphics[width=\smallImgWidth\linewidth, trim=1 1 1 1, clip]{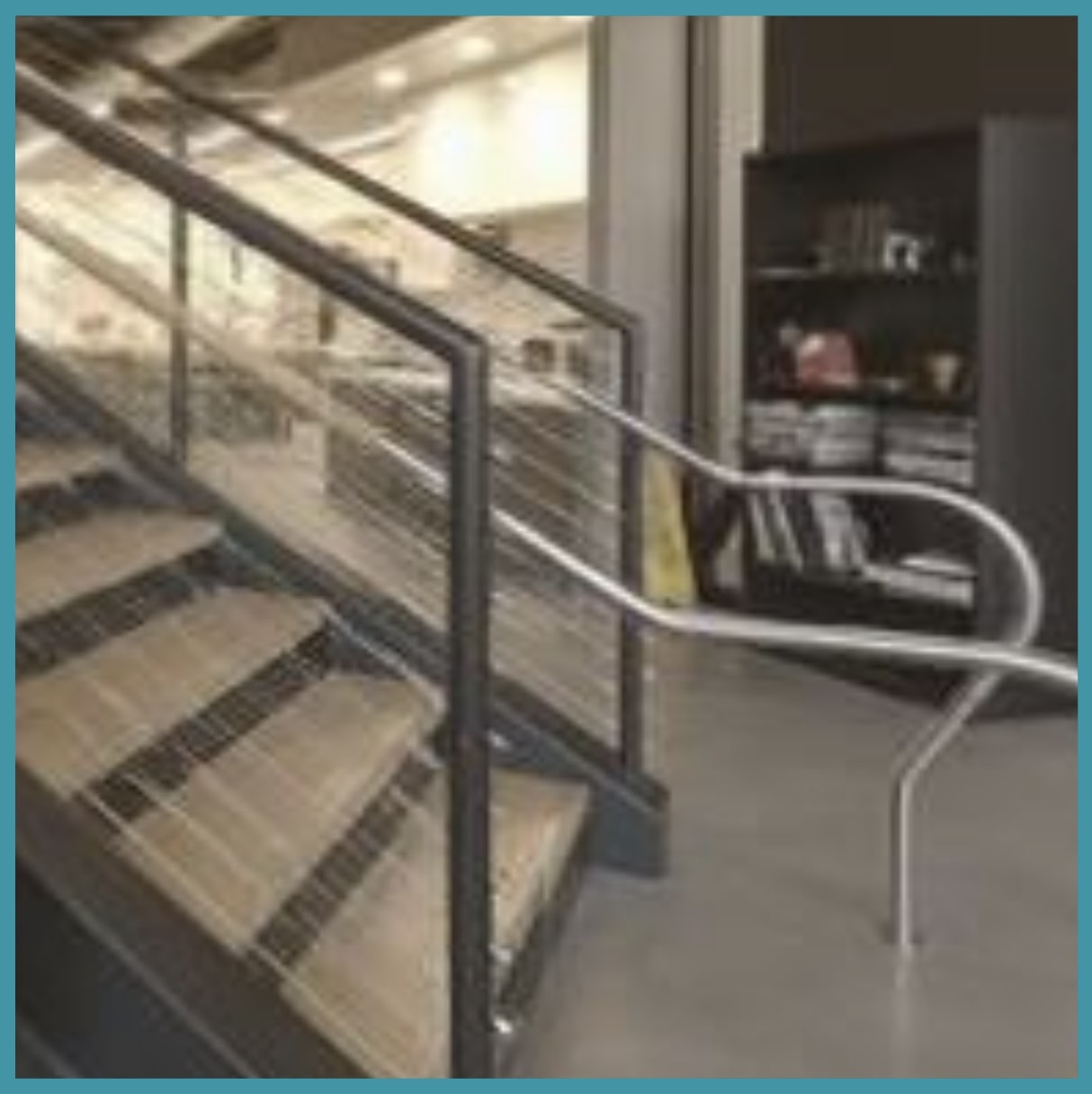} &
        \includegraphics[width=\smallImgWidth\linewidth, trim=1 1 1 1, clip]{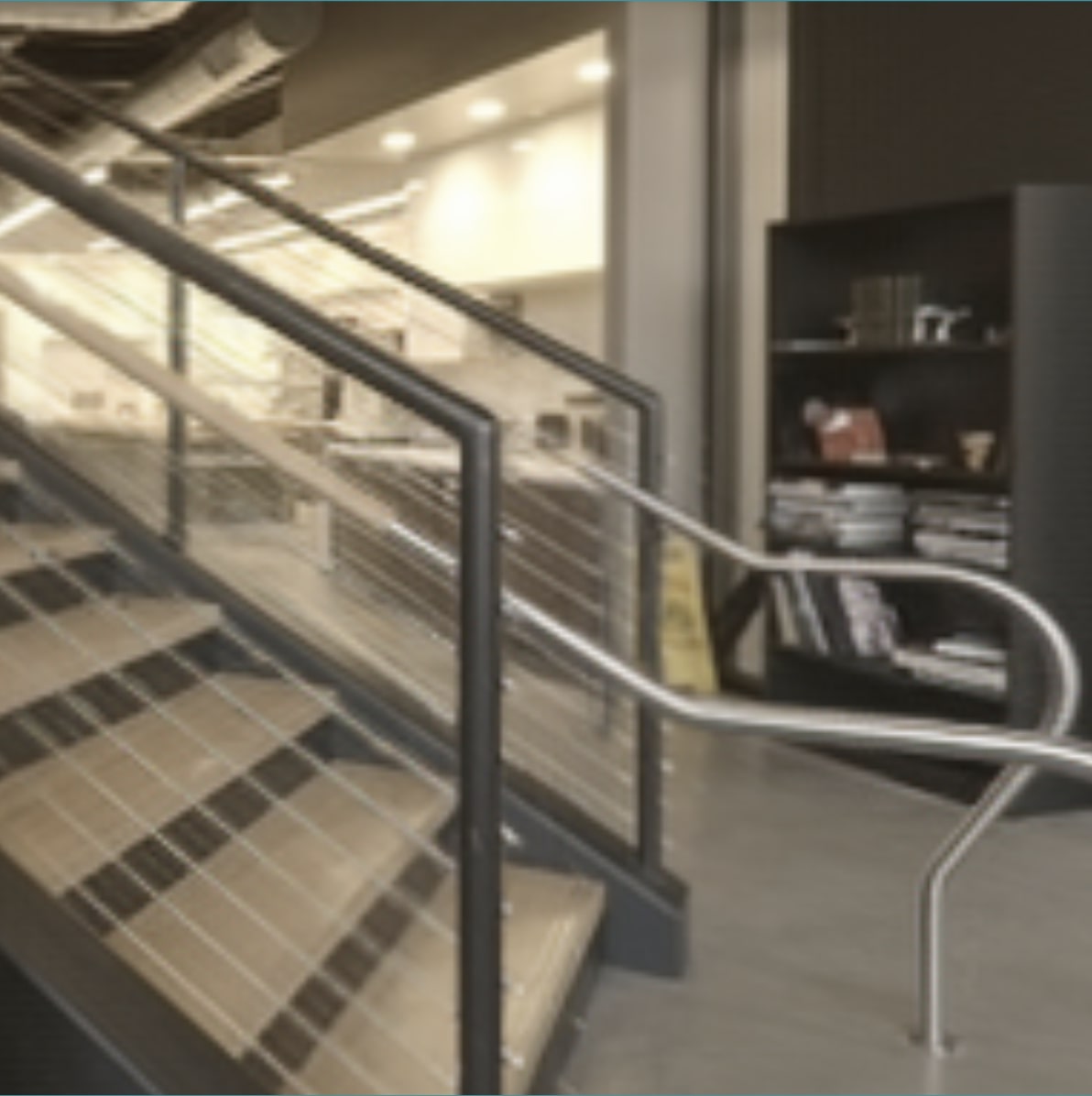} &
        \includegraphics[width=\smallImgWidth\linewidth, trim=1 1 1 1, clip]{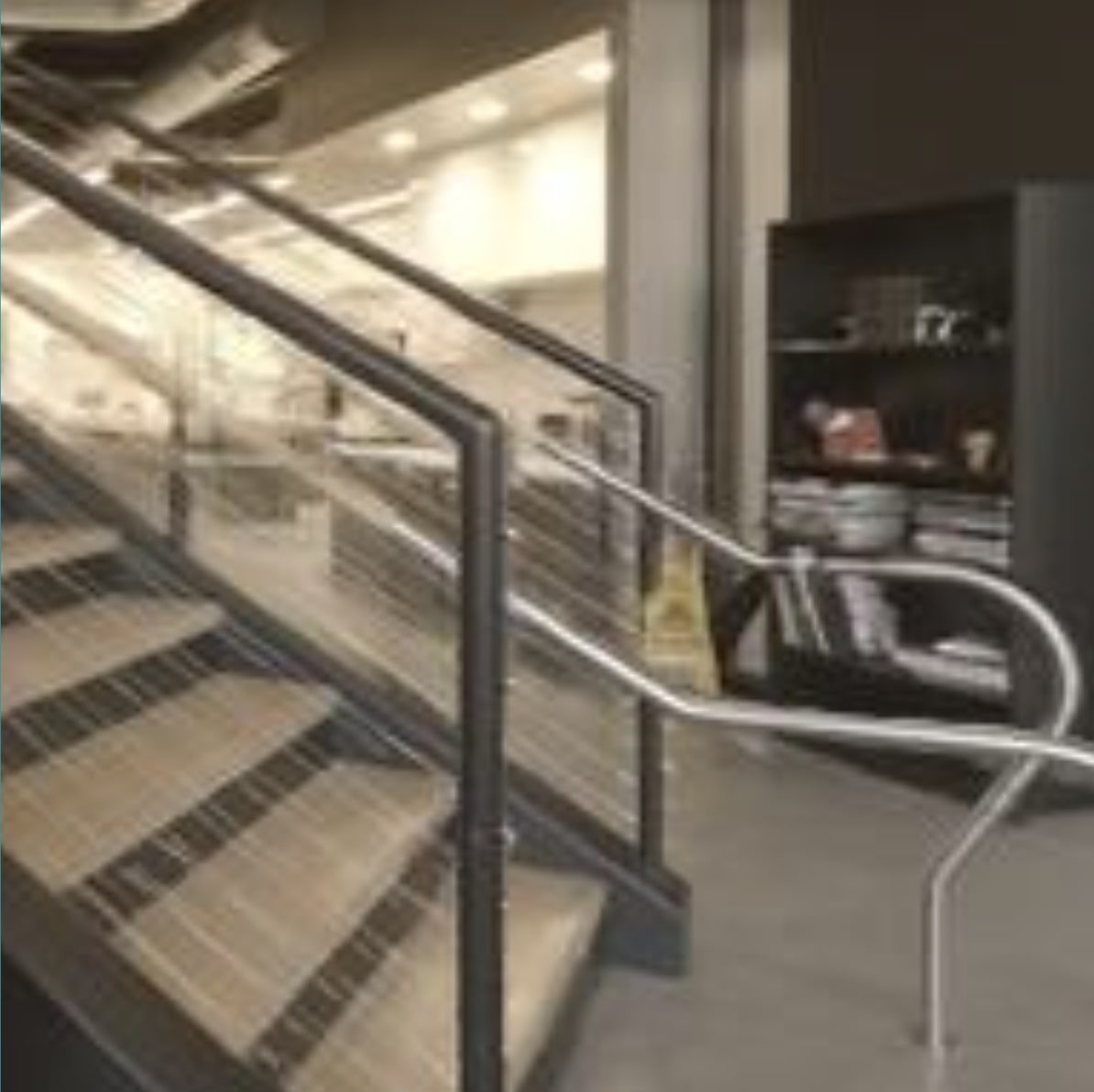} &
        \includegraphics[width=\smallImgWidth\linewidth, trim=1 1 1 1, clip]{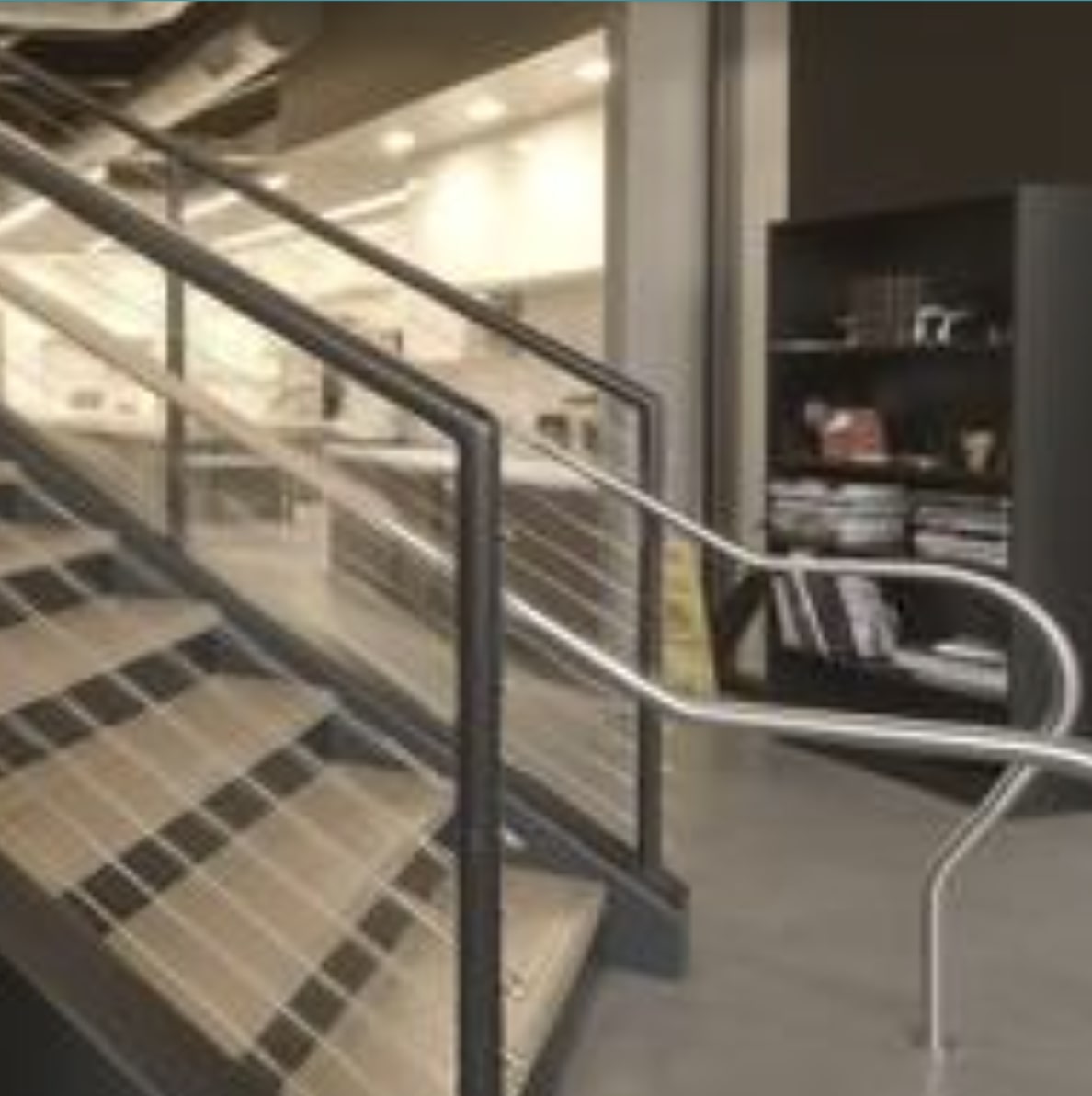} &
        \includegraphics[width=\smallImgWidth\linewidth, trim=1 1 1 1, clip]{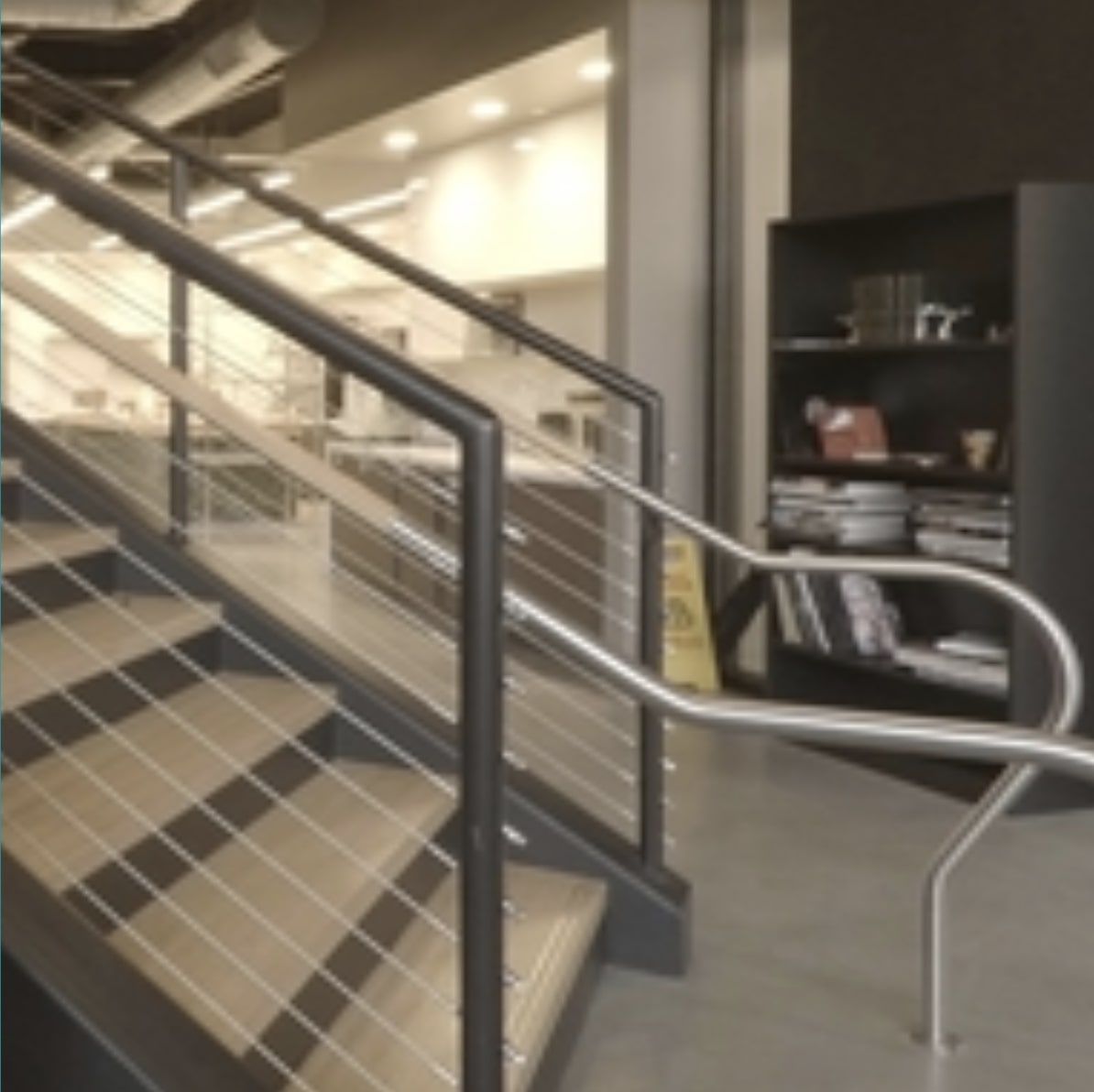} \\

             \multicolumn{1}{c}{\textbf{}} & \multicolumn{1}{c}{SIMPLI} & \multicolumn{1}{c}{IBRNet} & \multicolumn{1}{c}{fMPI-M} & \multicolumn{1}{c}{fMPI-L} & \multicolumn{1}{c}{Ground truth} \\

\multirow{2}{*}[6.075em]{
            \includegraphics[width=\bigImgWidth\linewidth, trim=1 1 1 1, clip]{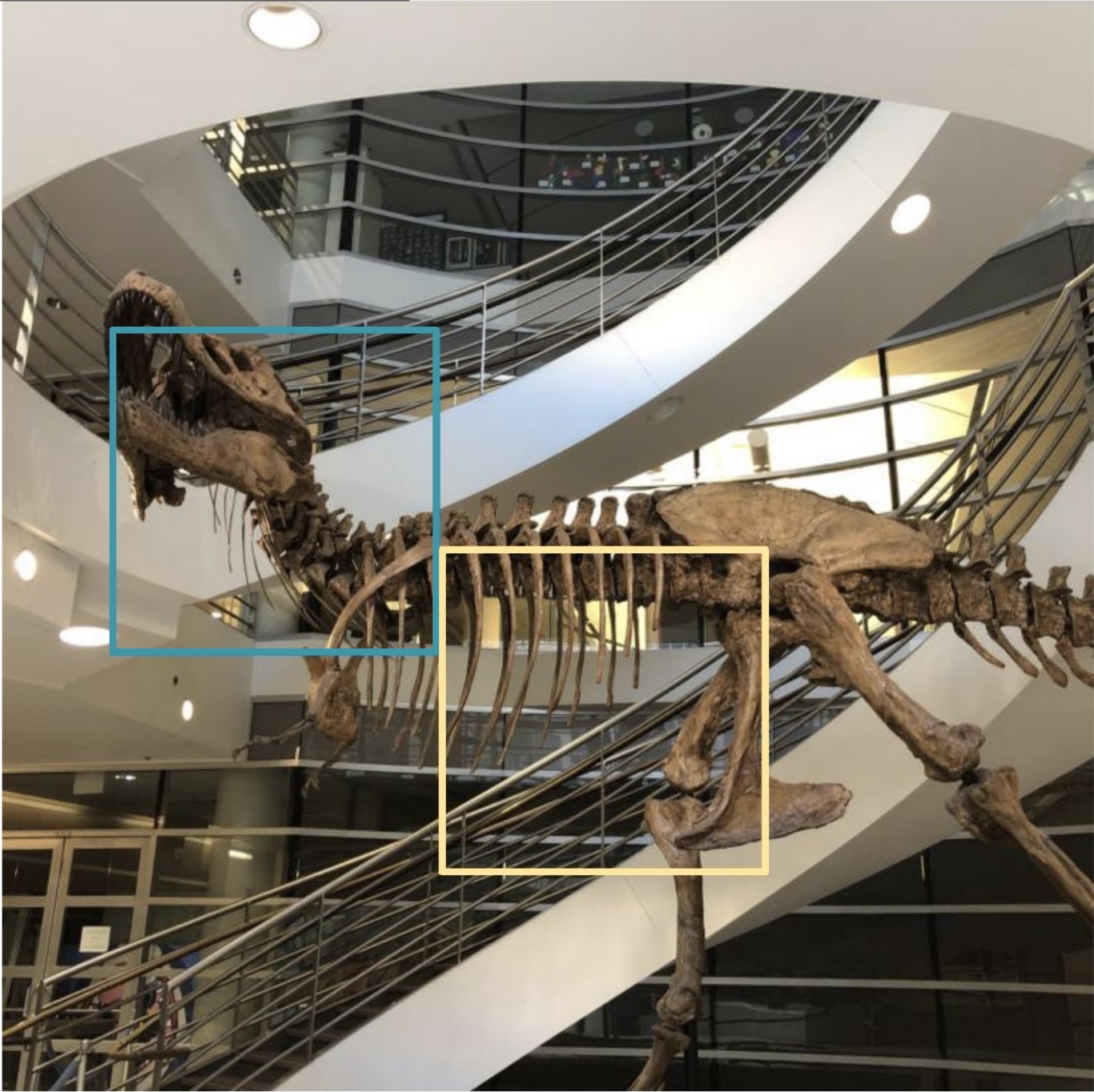}
        } &
        \includegraphics[width=\smallImgWidth\linewidth, trim=1 1 1 1, clip]{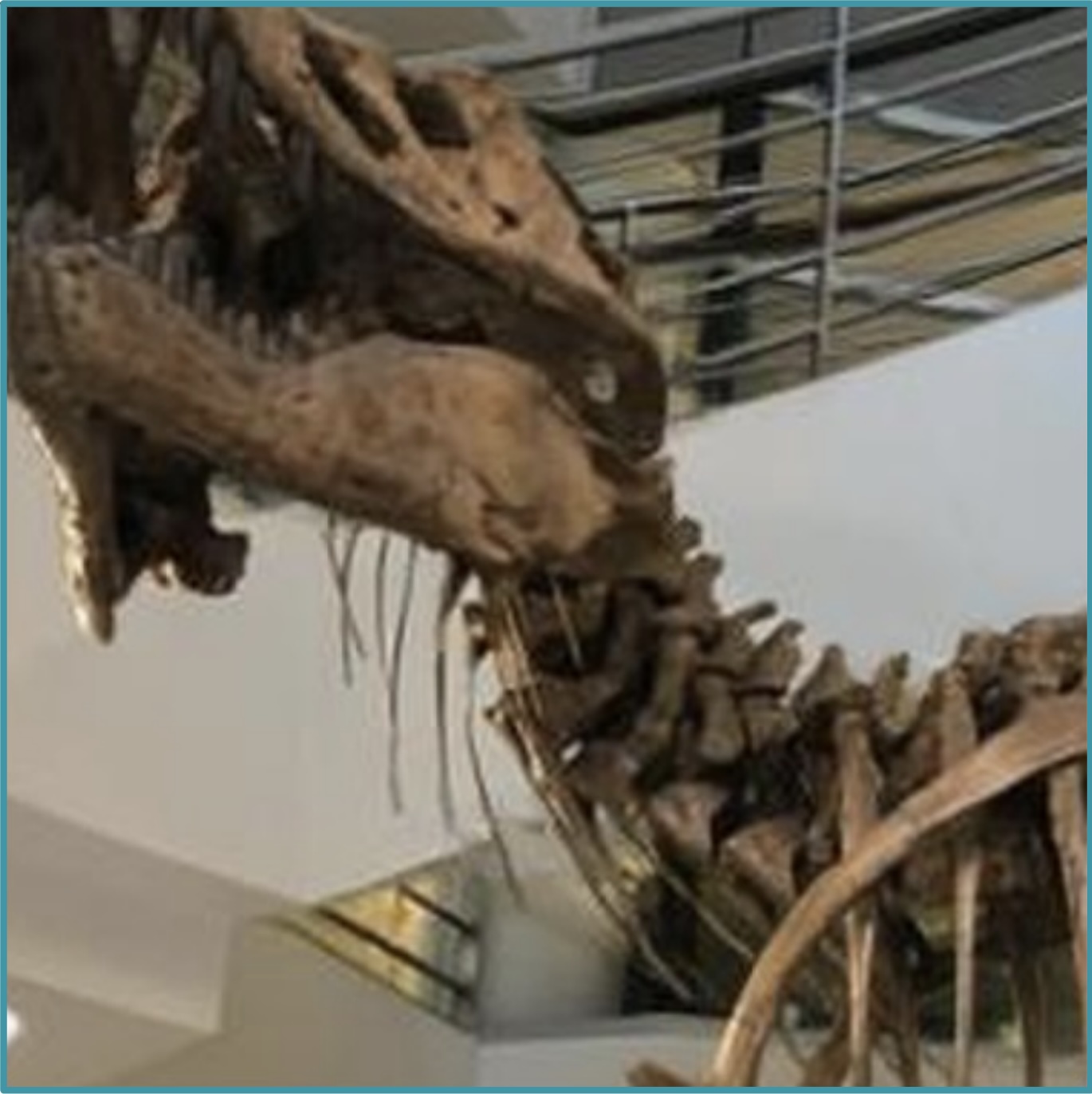} &
        \includegraphics[width=\smallImgWidth\linewidth, trim=1 1 1 1, clip]{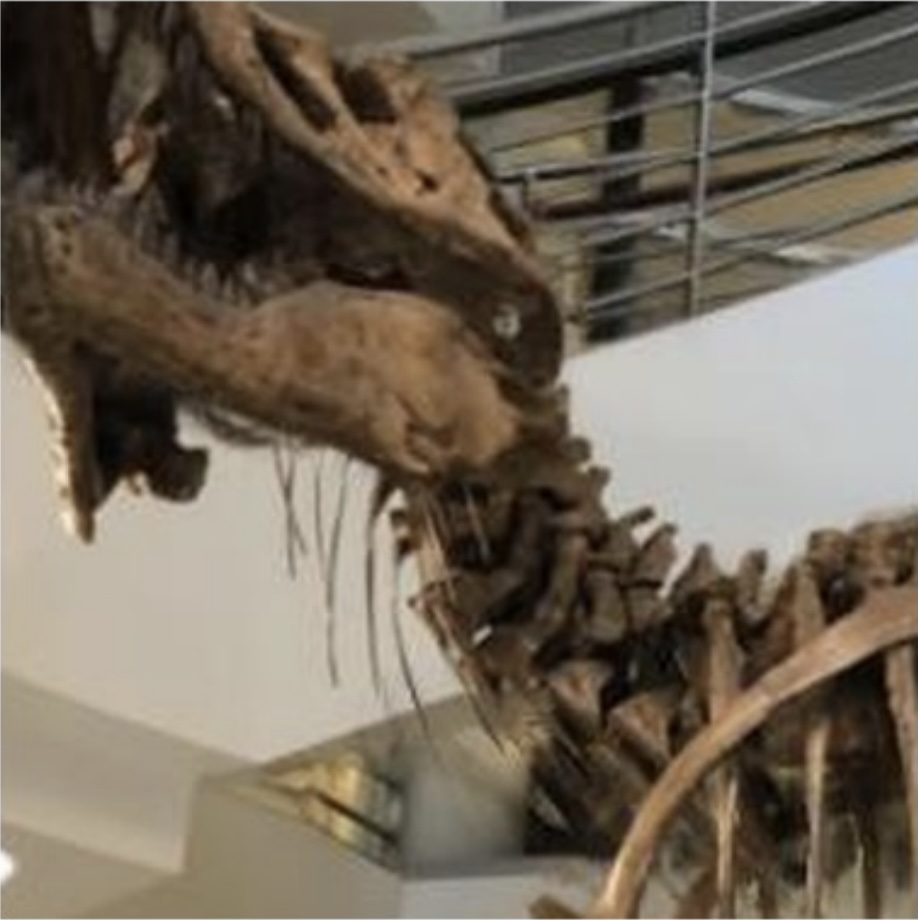} &
        \includegraphics[width=\smallImgWidth\linewidth, trim=1 1 1 1, clip]{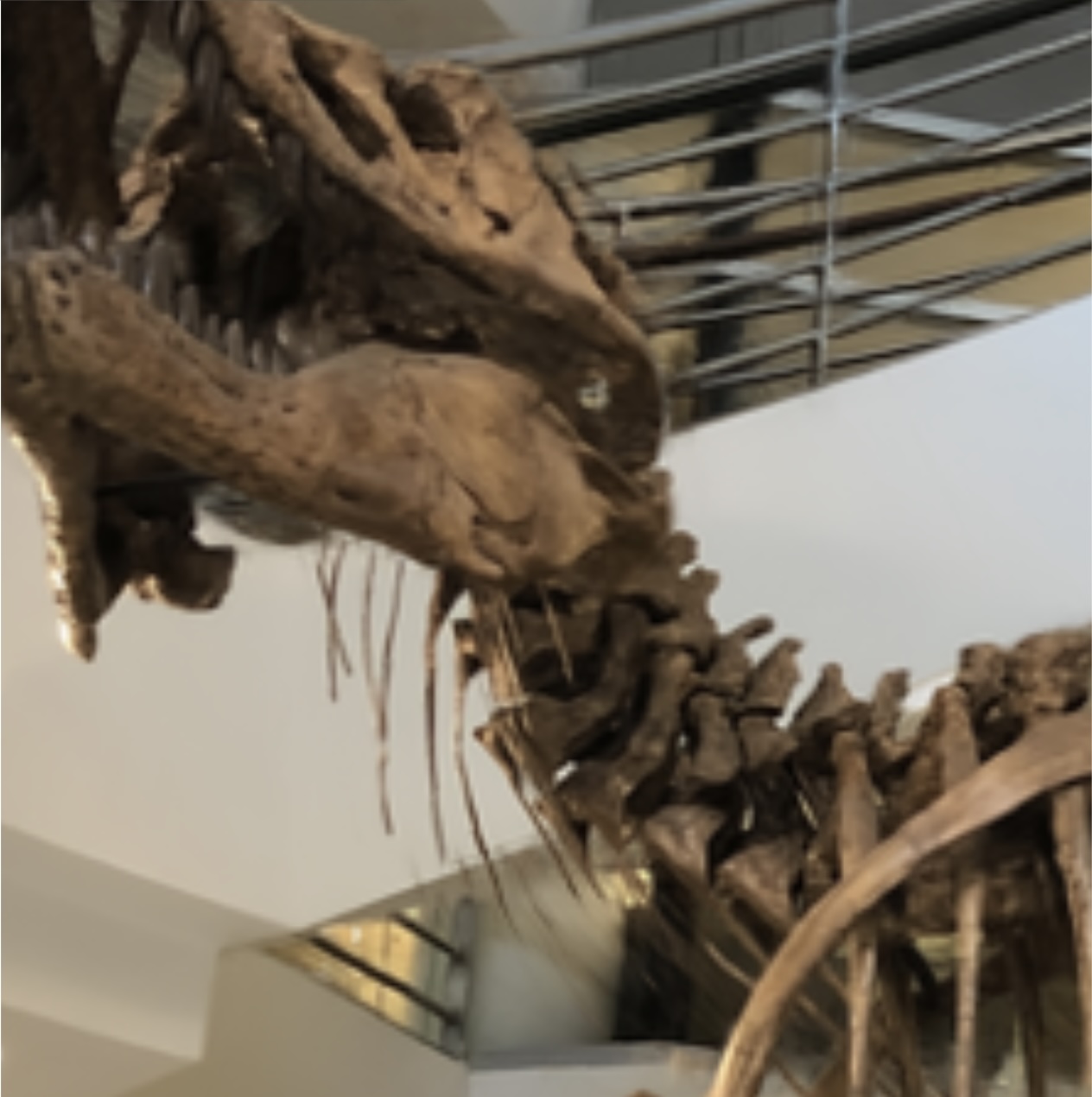} &
        \includegraphics[width=\smallImgWidth\linewidth, trim=1 1 1 1, clip]{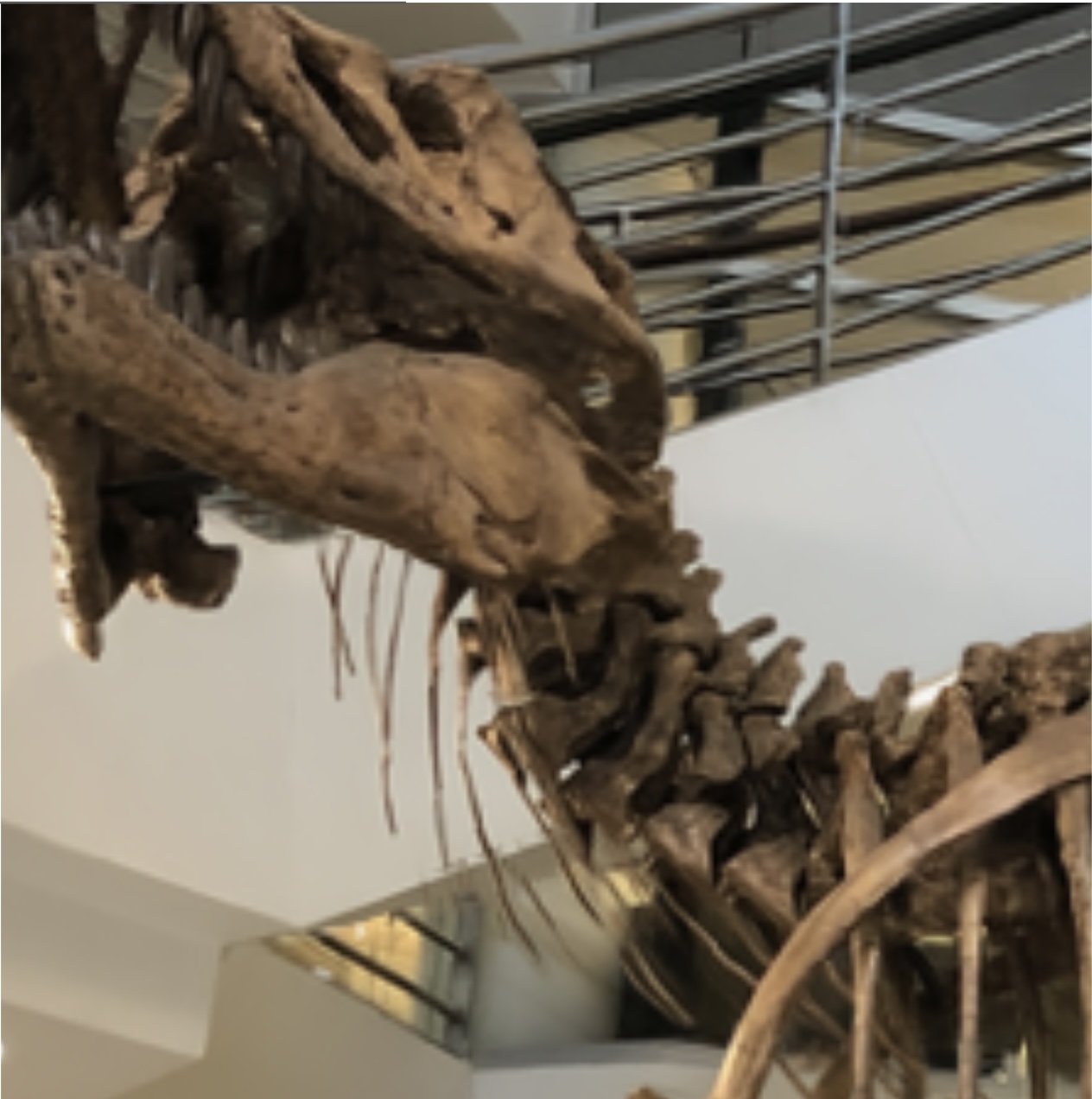} &
        \includegraphics[width=\smallImgWidth\linewidth, trim=1 1 1 1, clip]{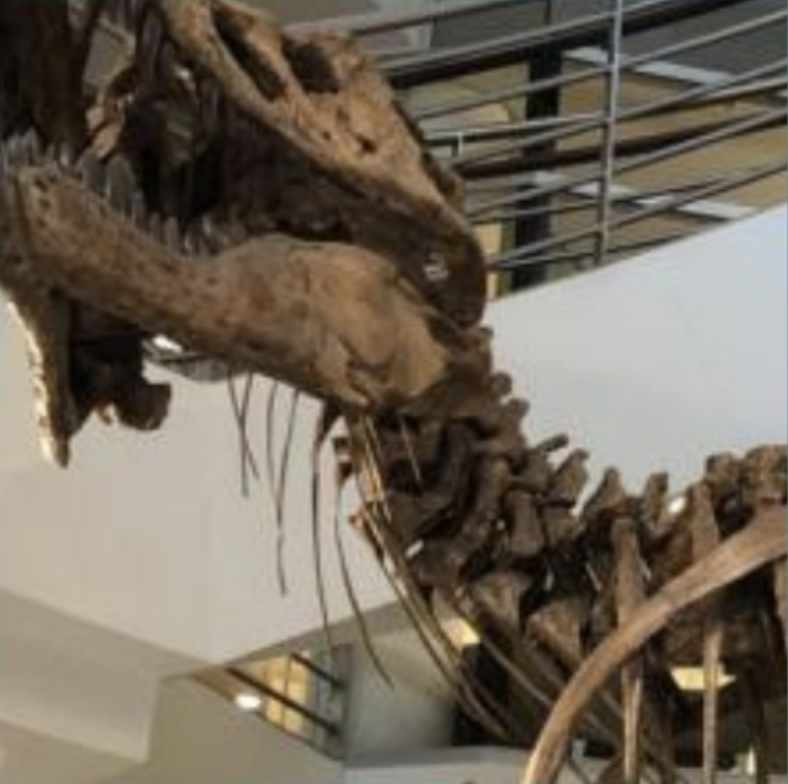}  \\
        &
        \includegraphics[width=\smallImgWidth\linewidth, trim=1 1 1 1, clip]{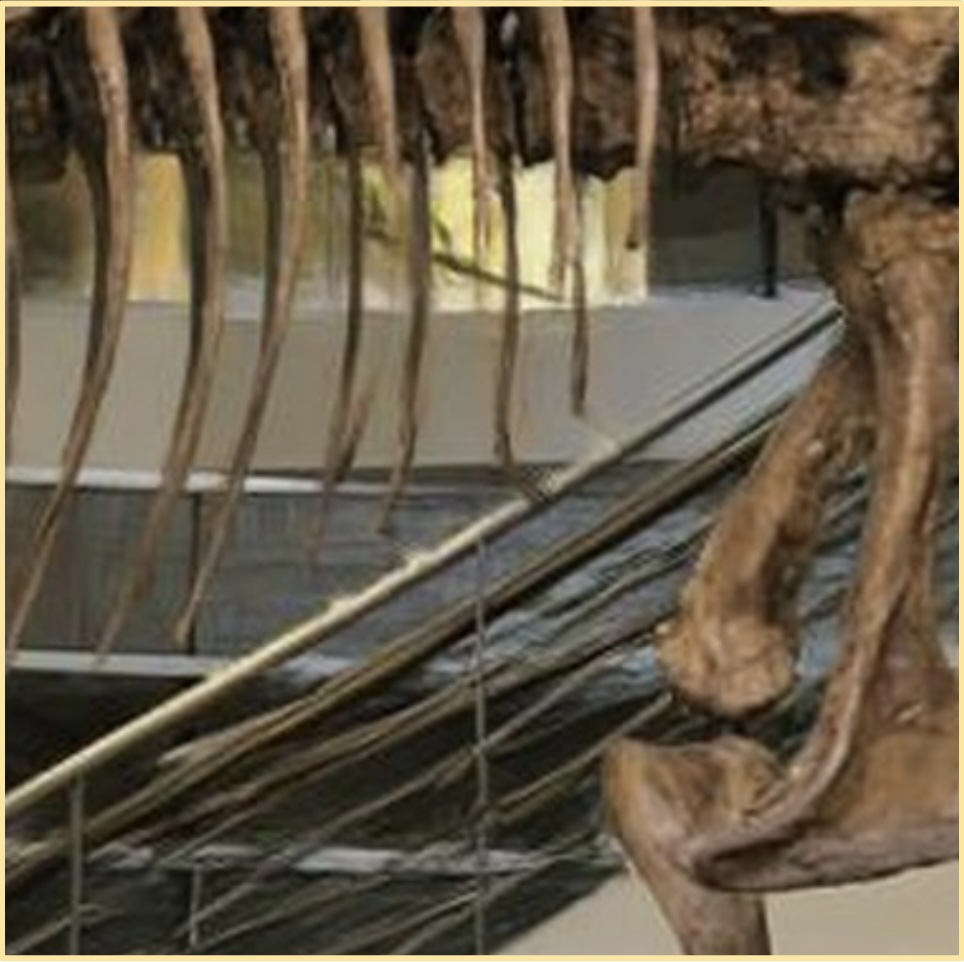} &
        \includegraphics[width=\smallImgWidth\linewidth, trim=1 1 1 1, clip]{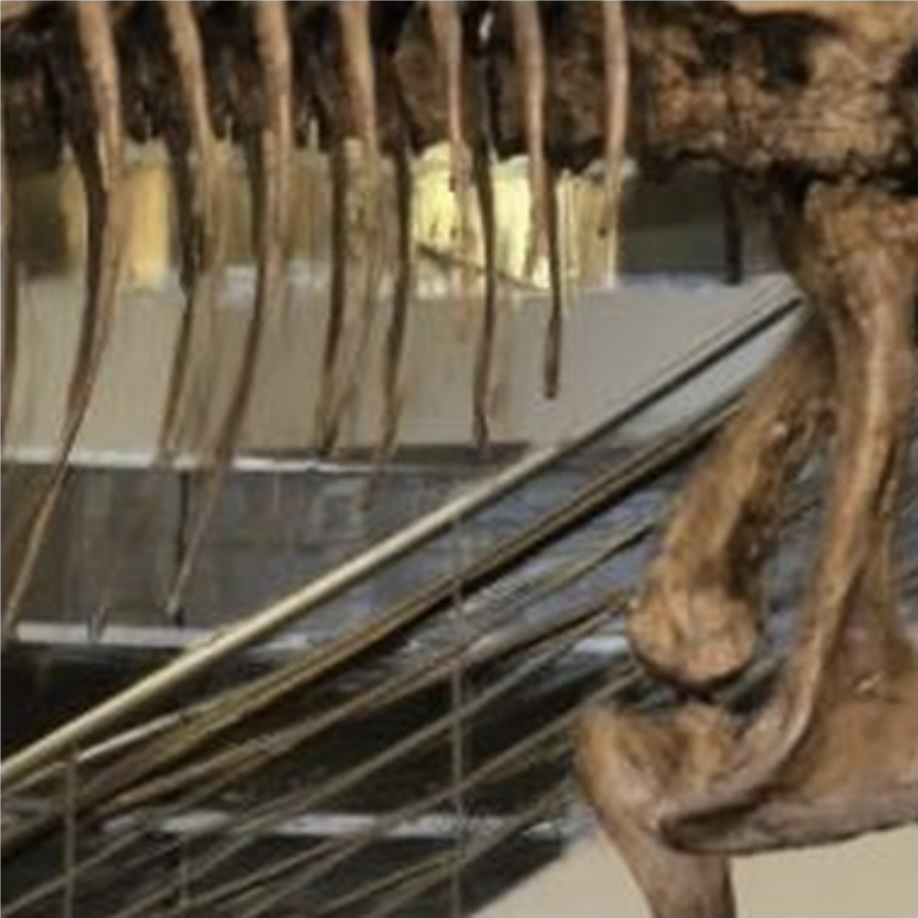} &
        \includegraphics[width=\smallImgWidth\linewidth, trim=1 1 1 1, clip]{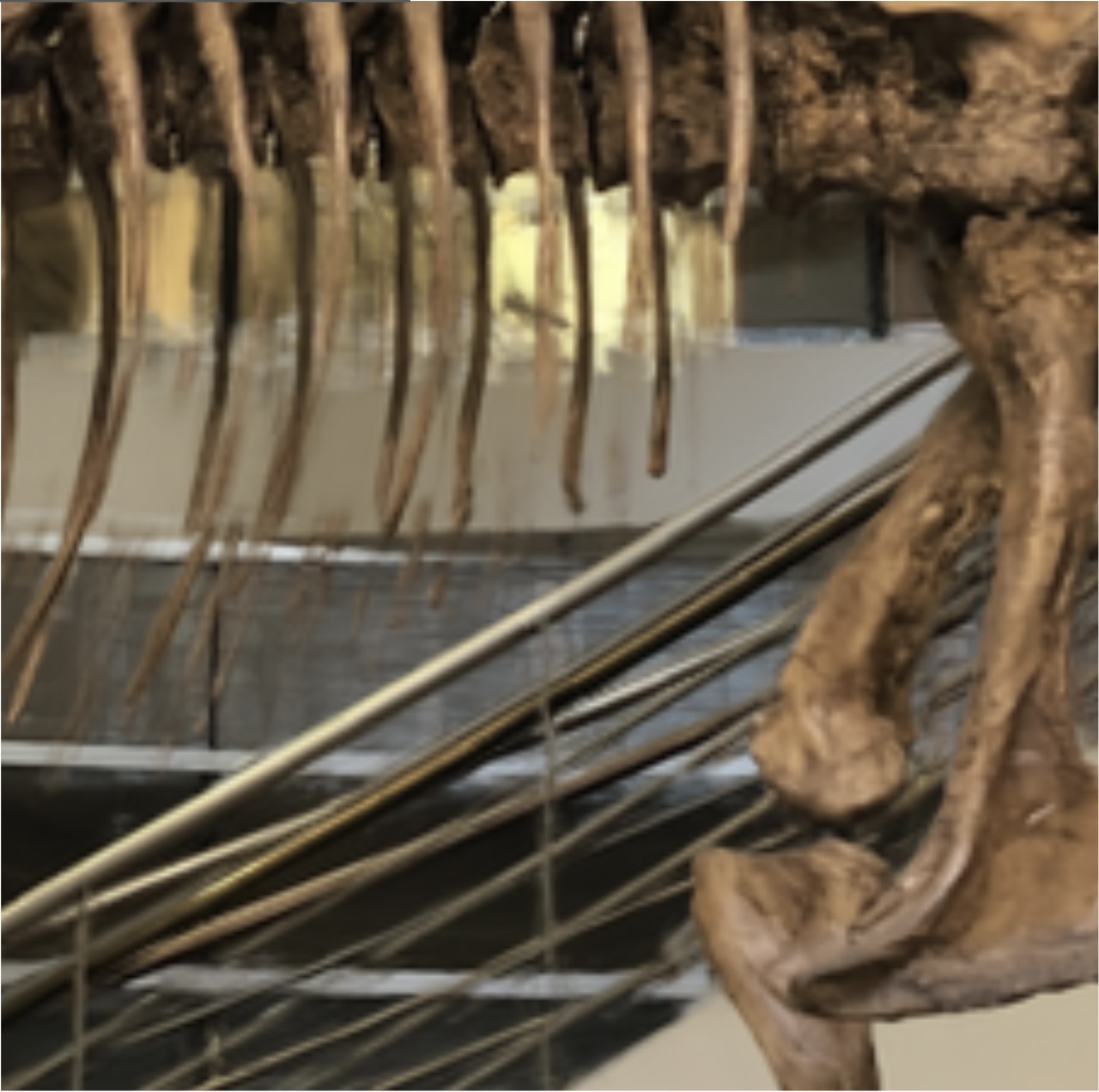} &
        \includegraphics[width=\smallImgWidth\linewidth, trim=1 1 1 1, clip]{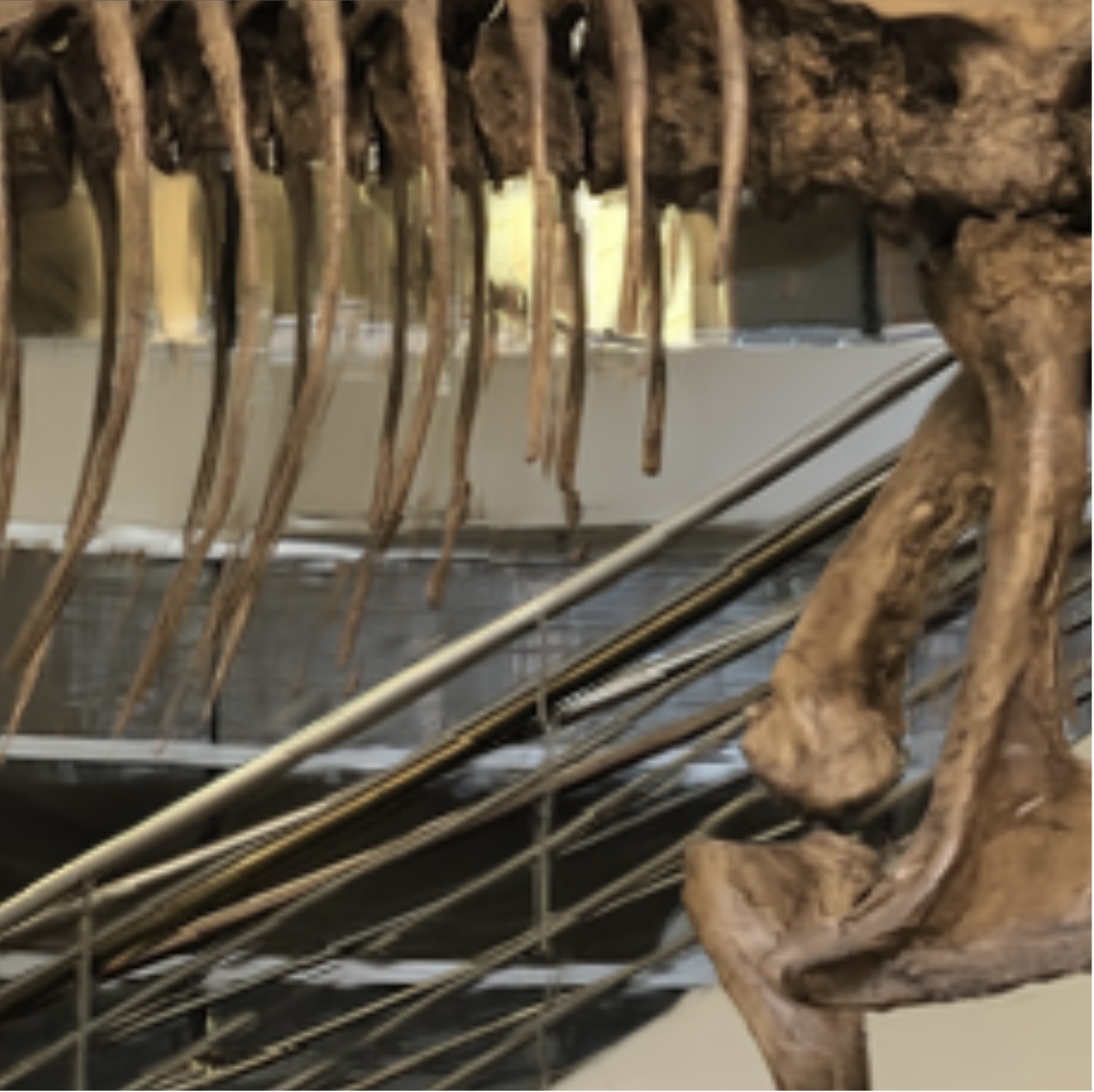} &
        \includegraphics[width=\smallImgWidth\linewidth, trim=1 1 1 1, clip]{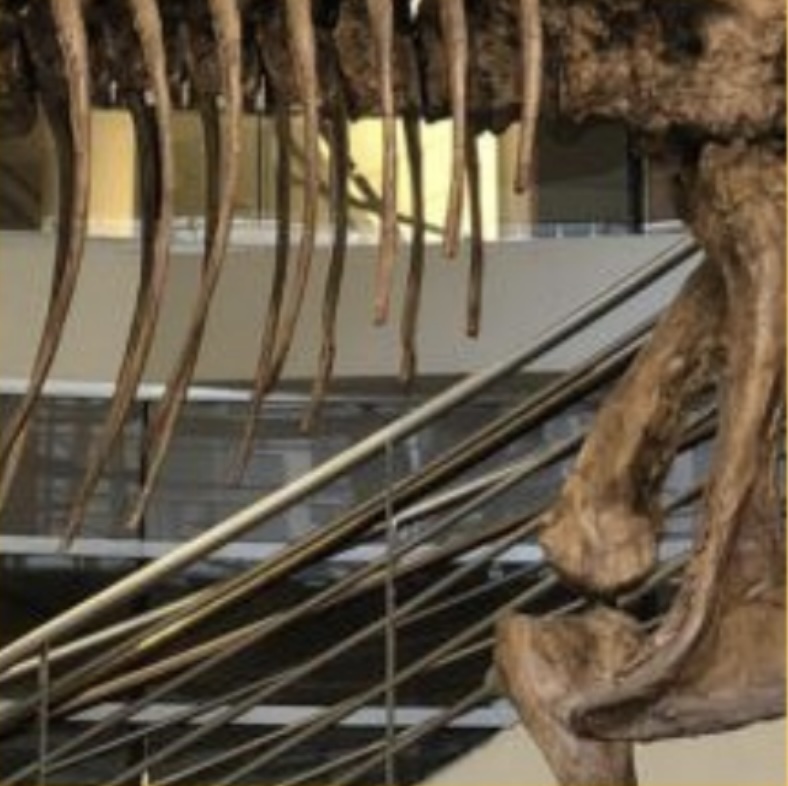}
             
    \end{tabular}
    \caption{\textbf{Visual comparison.} Results on a scene from the Spaces (top) and RFF (bottom) evaluation datasets. As can be seen, the image quality correlates well with the metrics presented in Table \ref{table:spaces} and \ref{table:rff}. Notice e.g. the strings of the handrails or the bones of the dinosaur.}
    \label{fig:all_images}
\end{figure*}

Since the described rendering approach is fully differentiable, it conveniently allows for training $\mathcal{F}_\theta$ solely based on image supervision. Toward this end, we construct and minimize the following loss function:

\begin{align}
\mathcal{L}\left(Y,\hat{Y}_\theta\right) &:= \| Y-\hat{Y}_\theta\|_1 + \text{SSIM}\left(Y,\hat{Y}_\theta\right) \nonumber \\
& +\lambda \|\text{VGG}(Y)-\text{VGG}(\hat{Y}_\theta)\|_1,
\end{align}

where $Y$ and $\hat{Y}_\theta$ constitute the ground truth and predicted target views. SSIM stands for the structural similarity index measure \cite{wang2004image} and the term $\|\text{VGG}(Y)-\text{VGG}(\hat{Y}_\theta)\|_1$ represents the so-called perceptual loss (LPIPS), calculated from the image embeddings given by the first four layers of a VGG19 network \citep{zhang2018unreasonable}. We choose $\lambda=0.01$ and minimize $\mathcal{L}$ with respect to $\theta$ under the classical empirical risk minimization regime, using the Lion optimizer \citep{chen2023symbolic} with learning rate $lr=0.00009$ as well as $\beta_1=0.99$ and $\beta_2=0.90$. Furthermore, we employ a learning rate decay factor of $10$ for the last $20\%$ (of $150'000$) iterations. As usual, we train on patches (in our case size $352 \times 352$) instead of full images. During each step, a random target view is chosen for each training sample. All experiments are performed on a set of eight NVIDIA A100 GPUs with $40$GB memory. Due to computational constraints, we did not grid-search any of the hyper-parameters.

\section{Results}\label{sec:results}
We present results on two publicly available NVS datasets, which are standard for benchmarking novel view synthesis methods. These datasets are not exclusively tied to NVS in the wild, but for a fair comparison, we will limit the methods considered in our benchmarks to those able to generalize to unseen scenes.\footnote{Excluding implicit and hybrid methods like NeX \cite{wizadwongsa2021nex}, which achieves outstanding performance (e.g. $0.7\%$ better SSIM on Spaces than DeepView) but is constrained to per-scene pre-training and higher numbers of input views.} At the time of writing, the best-performing methods include DeepView~\cite{flynn2019deepview}, LiveView~\cite{ghosh2021liveview}, SIMPLI~\cite{solovev2023self} and IBRNet~\cite{wang2021ibrnet} (see Section \ref{sec:related} for more details). At least one of the former methods beats other common benchmarks like Soft3D~\cite{penner2017soft}, LLFF~\cite{mildenhall2019local} and NeRF~\cite{mildenhall2021nerf} in terms of both runtime and performance. We furthermore include StereoMag~\cite{zhou2018stereo} into our benchmark as it is currently the fastest model.

\subsection{Spaces}

We first consider the Spaces dataset published by DeepView \cite{flynn2019deepview}. This dataset consists of 100 indoor and outdoor scenes captured 5 to 14 times from different viewpoints, using a fixed rig of 16 forward-facing cameras. 90 scenes are used for training and 10 scenes are held-out for evaluation. Each image has a resolution of $480 \times 800$. \cite{flynn2019deepview} presents four different camera setups. We here focus on the most challenging one called "4-views, large baseline", which has four inputs (arranged in a rectangle of approximate size $40 \times 25cm$) and 8 target cameras.

As can be seen in Table \ref{table:spaces}, fMPI-L achieves state-of-the-art in SSIM and PSNR, while being $\sim 50 \times$ faster than DeepView. This suggests that target-centered MPIs can be learned without sophisticated learned gradient descent algorithms when using local cross-plane context and a powerful backbone. Furthermore, fMPI-M performs comparably to LiveView but is $\sim3.5 \times$ faster. Finally, our fastest variant beats the current forerunner in terms of both performance and speed \cite{zhou2018stereo}. Figure \ref{fig:appendix_spaces} shows that fMPI-S can achieve good quality renderings from four input cameras with over $25$FPS (see Figure~\ref{fig:breakdowns} for detailed runtime breakdown).\footnote{We note that, for a fair comparison, we time all models in fp32 and \textit{omit} any runtime optimizations like compiling, operator fusion, kernel auto-tuning, static graph freezing, etc.}
\begin{table}[h]
\begin{tabular}{l|lllr}
               & SSIM $\uparrow$ & PSNR $\uparrow$ & LPIPS $\downarrow$ &  ms $\downarrow$ \\ \hline\hline
StereoMag-40    & 0.867           & -               & -                  & 121  \\
SIMPLI-8      & 0.9225          & 30.47           & 0.106             & 2'959  \\
LiveView-64     & 0.9475          & 31.52      & 0.099        & 375 \\
DeepView-80     & 0.9544          & 32.439         & \textbf{0.076}             & ~50'000 \\
fMPI-S         & 0.9178          & 29.57           & 0.128             & \textbf{34}  \\
fMPI-M         & 0.9426          & 31.13           & 0.084             & 101 \\
fMPI-L         & \textbf{0.9655} & \textbf{33.691} & 0.080     & 1'079

\end{tabular}
\caption{\textbf{Spaces}: Quantitative results and speed benchmark. For SIMPLI, LiveView, and DeepView, we obtained the synthesized images from the authors to recompute 
metrics using the open-sourced code from \cite{flynn2019deepview}. We obtained all runtimes on a single NVIDIA A100 GPU (mean of 30 runs) in fp32 and without any inference optimizations. The only exception is DeepView, for which we took timings from \cite{flynn2019deepview} (comparably high runtimes for DeepView were re-computed in \cite{solovev2023self}).}
\label{table:spaces}
\end{table}
\subsection{Real Forward-Facing} Additionally, we train and evaluate on the \textit{Real Forward-Facing} dataset, introduced by~\cite{mildenhall2019local}. This dataset is composed of 48 static indoor and outdoor scenes (40 for training and 8 for evaluation) with 20 to 62 images each from handheld cellphone captures. Camera poses are computed using the COLMAP structure from motion implementation \cite{schoenberger2016sfm}. $\tfrac{1}{8}$ of the images in the test scenes are held out as target views for the test set. Evaluation is performed by selecting the five closest images to the target view as input views. Table \ref{table:rff} shows a quantitative comparison of \textit{fast MPI} and previous methods on this dataset. We observe that fMPI-M and fMPI-L outperform previous methods by a large margin. fMPI-S performs on par with SIMPLI while offering 100x faster runtime speed, enabling real-time NVS in the wild. This raises doubts on the marginal value of more compact scene representations, as suggested by works on multi-layer images and layered depth images.\footnote{One can of course argue that MLI-based scene representations are more memory efficient, but we emphasize that even in this case, the plane sweep volume and the neural network activations are still memory heavy.}


\begin{table}[]
\begin{tabular}{l|lllr}
            & SSIM $\uparrow$ & PSNR $\uparrow$ & LPIPS $\downarrow$ &  ms $\downarrow$ \\ \hline\hline
IBRNet      & 0.73            & 22.69           & 0.19               & 1'563 \\
DeepView-80 & 0.76            & 23.11           & 0.13               & 40'887 \\
SIMPLI-8    & 0.79            & 23.58           & 0.11               & 2'237 \\
fMPI-S      & 0.792           & 22.73          & 0.13             & \textbf{19}  \\
fMPI-M      & 0.840           & 23.60          & 0.10              & 54 \\
fMPI-L      & \textbf{0.862}  & \textbf{24.71} & \textbf{0.08}      & 429
\end{tabular}
\caption{\textbf{Real Forward-Facing}: Quantitative results and speed on a NVIDIA A100 GPU. All \textit{fast MPI} models are trained solely on \textit{Real Forward-Facing} training data at a resolution of $378 \times 504$ pixels. IBRNet, DeepView, and SIMPLI results taken from \cite{solovev2023self}.}
\label{table:rff}
\end{table}

\subsection{Ablations}\label{sec:ablations} 
To substantiate the claims of our two major contributions, ablations on plane grouping and super-sampling are depicted in Fig.~\ref{fig:pgrouping} and Fig.~\ref{fig:psampling}. Regarding the former, we find that plane grouping not only offers a simple speed-performance trade-off but more importantly  Fig.~\ref{fig:pgrouping} shows that processing PSV planes in groups yields strictly better performance than either of the two currently common approaches (namely plane by plane or joint processing). Regarding the latter, Fig.~\ref{fig:psampling} shows that super/sub-sampling planes offers a simple, yet powerful, option for practitioners to gain performance/save runtime with a given MPI method. This claim is substantiated by the fact that fMPI-M matches the performance of LiveView despite using only half the number of PSV planes (32 vs 64), which yields 3x speed-ups using the same network architecture.

In what follows, we review further design choices of fMPI. First, we have chosen to generate \textit{target-centered MPIs}. These can be generated on the fly and are thus suitable for dynamic scene content. However, warping static MPIs to novel target views may yield temporal consistency across views. For in the-wild applications with dynamic scene content, we foresee that the best approach will be a hybrid method, re-computing MPIs at given intervals and using view-consistent homography warpings in between. From a runtime perspective, Sect.~\ref{sec:homography} shows that our innovations allow for generating novel MPIs at the target view at no more than $2$x the latency of warping a static MPI to a novel view.\footnote{Compare the time for generating 32 MPI planes on the left of Fig.~\ref{fig:psv_timings} (19.6ms) to the runtime of fMPI-S (34ms), which also gives 32 MPI planes} From a quality perspective, we find similar performance. Namely, the SSIM of fMPI-M on spaces drops by only $0.011$ when generating a single MPI (at the center of the camera rig) and warping it to all nine target views, instead of generating one MPI per target view.\looseness=-1

Second, as described in Section~\ref{sec:training_details}, in addition to input view weights, $\mathcal{F}_\theta$ also outputs a single RGB image for each plane group, to facilitate inpainting. Although the benefits of this approach are relatively marginal, (increase in SSIM of $0.0027$ for fMPI-M on spaces) we opted to retain this feature due to the minimal computational overhead ($\sim 3.5ms$).

Third, to isolate the impact of the larger backbone of fMPI-L from the fact the these weights where pre-trained on ImageNet, we retrained this model from scratch with random initialization. Interestingly, we observed that the spaces SSIM score declined by $0.006$, indicating that approximately $26.2\%$ of the performance gap between our M and L models is attributable to pre-training.

\section{Limitations and future work}
In this study, we presented two novel input processing paradigms for layer-based NVS methods that significantly improve their runtime requirements. Our approach is highly flexible, enabling a trade-off between performance and speed while outperforming state-of-the-art methods on public benchmarks. Notably, it is also very general and can benefit layer-based NVS methods of all sorts. Despite these improvements, some limitations and interesting areas for future work remain. Firstly, temporal consistency is not enforced in our method, which could lead to inconsistencies when synthesizing novel views in videos. We thus consider the integration of our innovations into an MPI pipeline for online video generation an interesting follow-up. Secondly, developing optimized code for homography warping and optimizing network inference times, for example by employing advanced techniques such as horizontal and vertical operator fusion, kernel auto-tuning and dynamic memory management (as done in optimization libraries like \citep{vanholder2016efficient} and \citep{AITemplate}), has the potential to yield significant speedups. Thirdly, more efficient backbone architectures should further enhance the real-time capabilities of our method. In this regard, inspirations can be drawn from the vast literature on efficient segmentation models (see e.g.~\citep{mehta2018espnet}). Lastly, depth-based losses could further improve rendering quality when ground truth depth is available.

Importantly, we did not consider the memory requirement of methods based on layered representations. Generally, these methods have high memory footprint, which can pose limitations in resource-constrained environments. Exploring ways to reduce this footprint presents an intriguing avenue for future research. Finally, integrating our approach with adaptive positioning of the MPI planes in space (e.g. \cite{navarro2022deep}), rather than fixing them at inverse disparity levels, is likely to enhance performance in practical applications and could lead to additional runtime savings.

{
    \small
    \bibliographystyle{ieeenat_fullname}
    \bibliography{sample-bibliography}
}

\appendix

\section{Appendix}

\subsection{Details on PSV generation}\label{sec:homography}
Generating the plane sweep volume involves inversely warping each input RGB onto each of the PSV planes, placed at the target perspective. For every depth value $d_i$, we consider the geometry of the input view being transformed (i.e., the source) as $n \cdot x + a = 0$. Here, $n$ represents the plane normal, and $x = [u_s, v_s, 1]^T$ signifies the homogeneous coordinates of the source pixel. The rigid 3D transformation matrix that maps the source to the target camera is determined by a 3D rotation, $R$, and a translation, $t$. Additionally, the intrinsic parameters of the source and target cameras are represented by $k_s$ and $k_t$, respectively. For each pixel $(u_t, v_t)$ in the target PSV plane, we employ a standard inverse homography \cite{hartley2003multiple} to derive

\begin{equation}
\begin{bmatrix}
u_s \\
v_s \\
1
\end{bmatrix}
\sim k_s \underbrace{\left(R^\intercal + \frac{R^\intercal tn R^\intercal}{-\left(d_i+nR^\intercal t\right)}\right)}_{H_i}
 k_t^{-1}
\begin{bmatrix}
u_t \\
v_t \\
1
\end{bmatrix},
\end{equation}
where we omitted the plane index $i$ on $u,v,R$ and $t$.

As a result, we can determine the color and alpha values for each target pixel $[u_t, v_t]$ by referring to its corresponding coordinates $[u_s, v_s]$ in the source image. Three high-level steps are needed to implement this logic:

\begin{itemize}
    \item Mapping: We first compute the homography matrix $H_i$ for each PSV plane and each input camera.
    \item Coordinate transformation: Second, we calculate homogeneous coordinates and apply the previously computed homography matrices to obtain the warped grid. Finally, the warped grid is converted back to the Cartesian coordinate system.
\item View sampling: Finally, we sample from the input views at the warped grid coordinates, employing bilinear interpolation for non-integer indices. 
\end{itemize}

\begin{figure}[h]
\centering
\begin{subfigure}[b]{0.49\columnwidth}
\includegraphics[trim={1cm 0.5cm 2cm 2.5cm},clip,width=\textwidth]{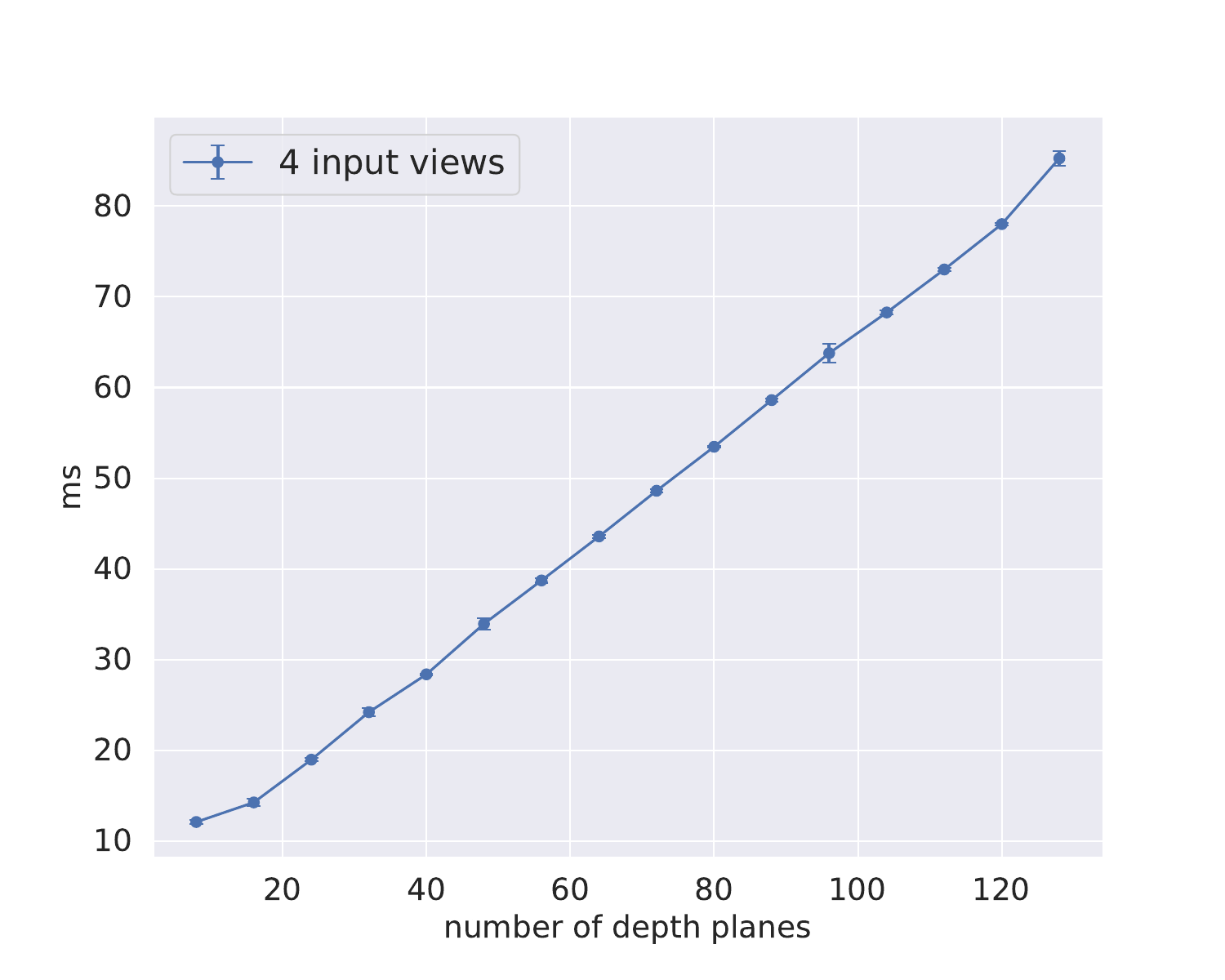}
\label{fig:1a}
\end{subfigure}
\hfill
\begin{subfigure}[b]{0.49\columnwidth}
\includegraphics[trim={1cm 0.5cm 2cm 2.5cm},clip,width=\textwidth]{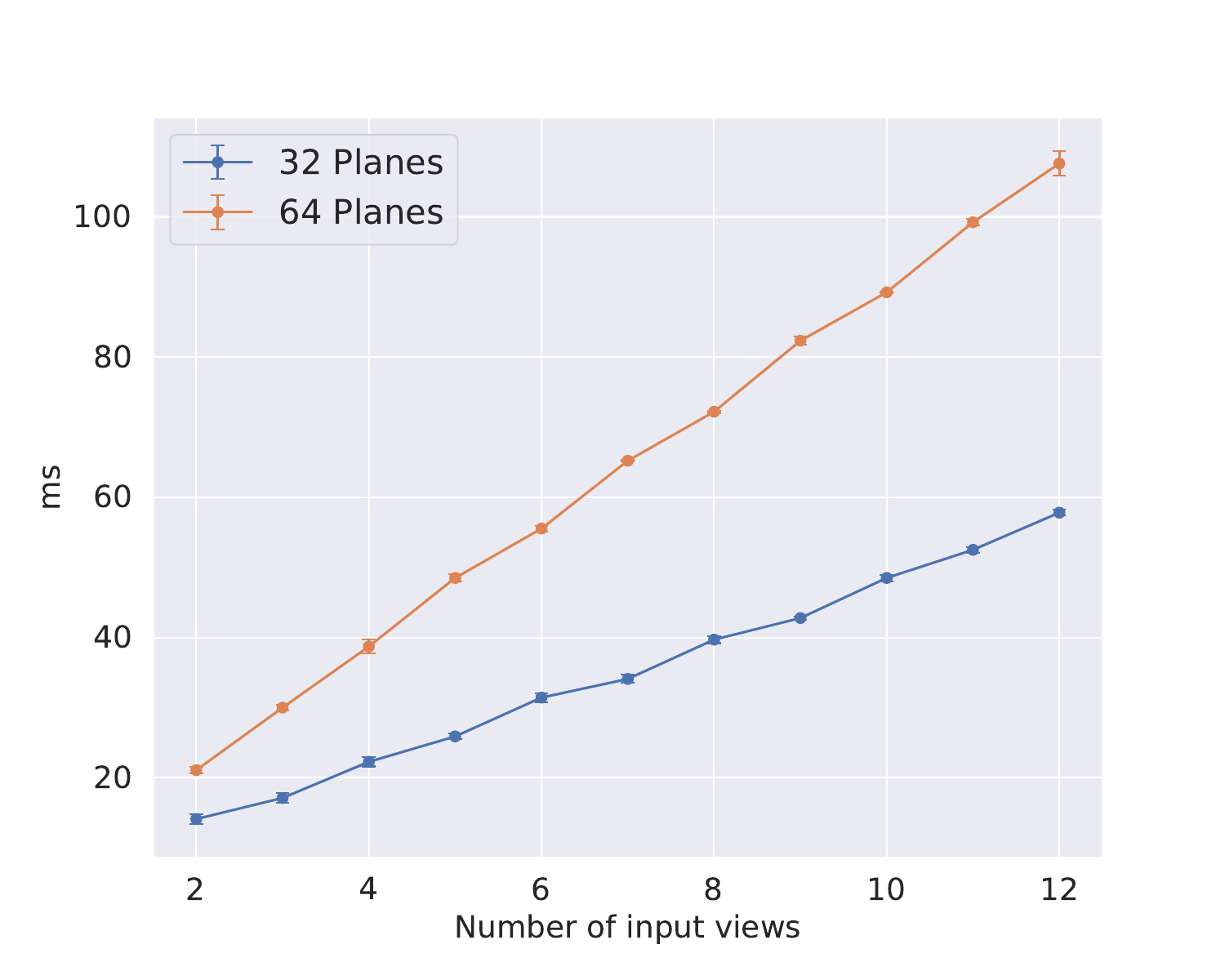}
\label{fig:1b}
\end{subfigure}
\caption{Time to generate the plane sweep volume over the number of depth planes (left) and the number of input views (right) for resolution $464\times800$. Four input views were assumed on the left. Mean and standard deviation of 30 runs on an A100 GPU.}
\label{fig:psv_timings}
\end{figure}

\begin{figure}[h]
\centering
\begin{subfigure}[b]{0.49\columnwidth}
\includegraphics[trim={0.5cm 1cm 2cm 1cm},clip,width=\textwidth]{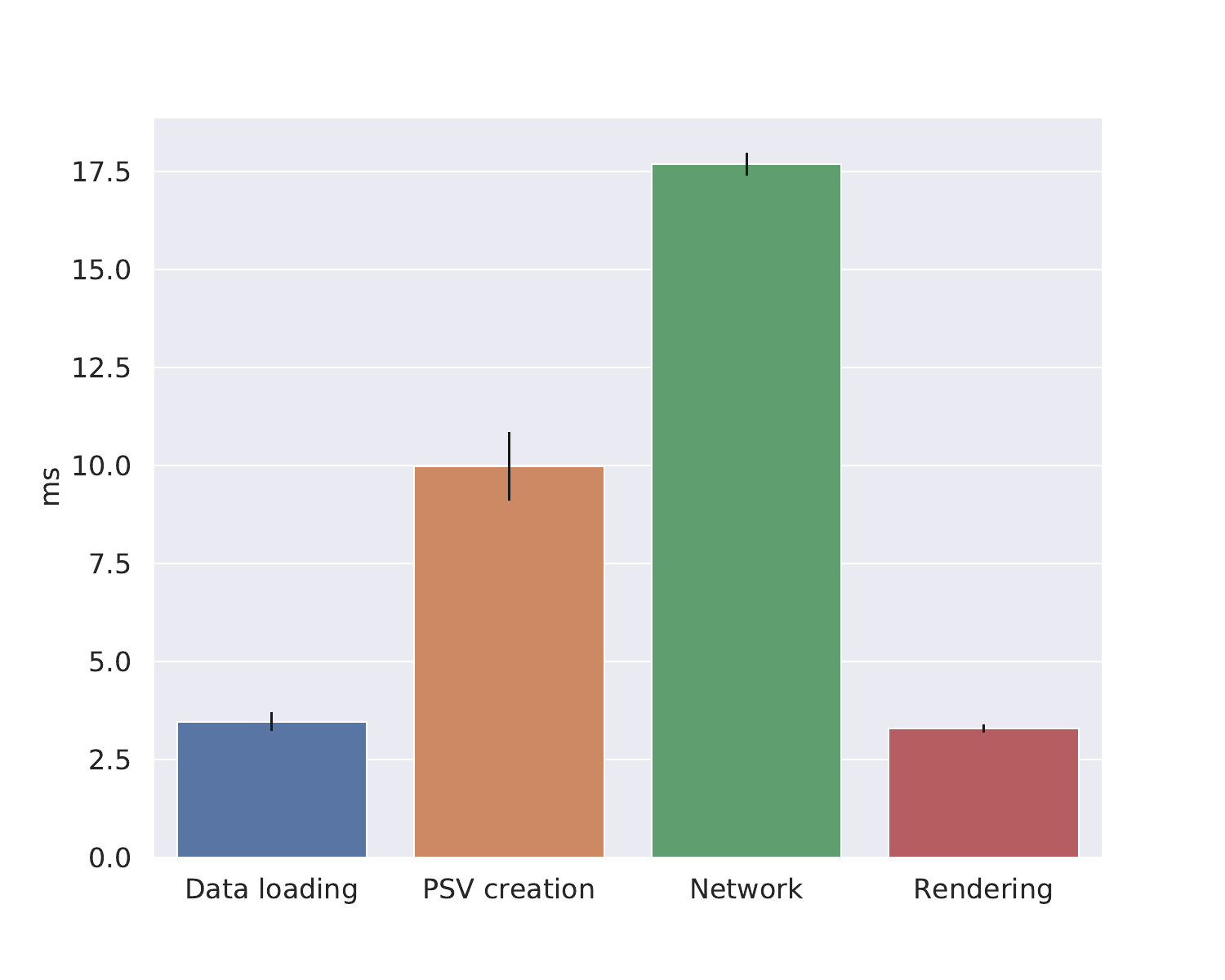}
\label{fig:1a}
\end{subfigure}
\hfill
\begin{subfigure}[b]{0.49\columnwidth}
\includegraphics[trim={0.5cm 1cm 2cm 1cm},clip,width=\textwidth]{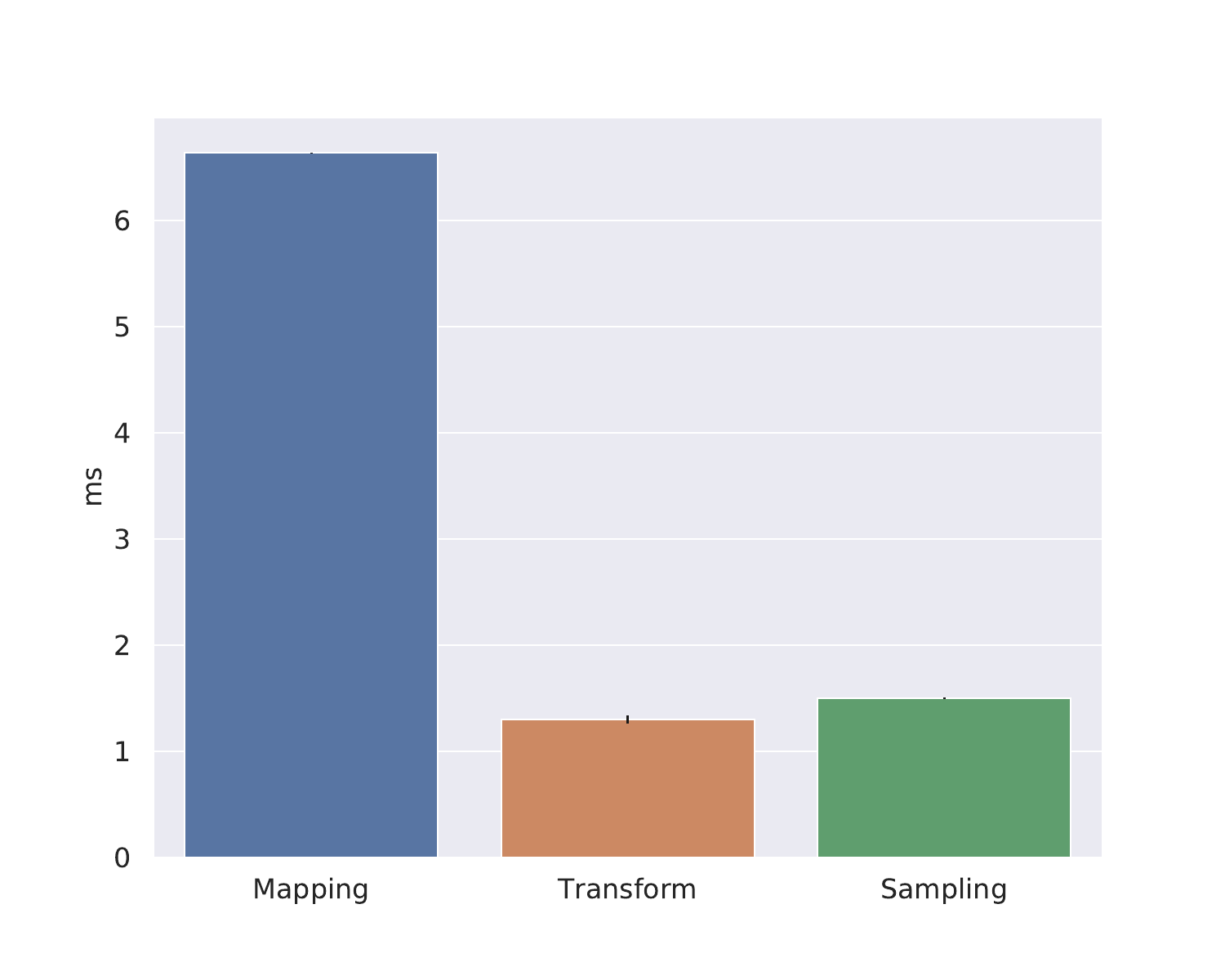}
\label{fig:1b}
\end{subfigure}
\caption{Runtime breakdown of f-MPI S on spaces (left) and breakdown of PSV generation process (right) in the same setting, i.e. $16$ planes at resolution $464 \times 800$.}
\label{fig:breakdowns}
\end{figure}

Figure \ref{fig:psv_timings} illustrates that the total PSV generation time increases linearly with both the number of planes and the number of input views. Moreover, Figure \ref{fig:breakdowns} provides a representative example of how this total time is divided into the three previously discussed steps, specifically for the case of fMPI-M in the spaces setting (four cameras at $464x800$ ). It is worth noting that for applications that have a fixed camera setup, such as Passthrough on VR headsets, the initial step can be pre-calculated, thereby resulting in substantial time savings.

\subsection{Details on backbone and MPI generation}\label{sec:details_network}
Table \ref{table:u-net} presents details of the simple U-Net backbone used in fMPI-S and fMPI-M. The network is fully convolutional with ReLU activations in all but the final layer. Simple nearest neighbor interpolation is used for upsampling. For fMPI-L, we simply mirror the four levels of ConvNext tiny \cite{liu2022convnet} to create an encoder-decoder architecture, again employing nearest neighbor interpolation for upsampling. We then adapt the first and last layers to match the input/output channels of the U-Net backbone.

For both networks, the input equals the number of planes per PSV group times the number of input images $V=\left |\{I_v\}_{v=1}^V\right|$, times three color channels. The network outputs for each of the $D/G$ planes in the group: $V-1$ view weights $w_i$, one $\alpha$ channel, three additional $RGB$ channels  $I_{V+1}$ for inpainting, as well as one additional weight $w_{V+1}$. Subsequently, the RGB value of a given pixel in a given MPI plane is obtained by the following equation:

\begin{equation}\label{eq:RGB_app}
RGB=\sum_{i=0}^{V+1}\frac{e^{w_i}}{\sum_{j=0}^{V+1}e^{w_j}} I_{i},
\end{equation}

where the weight for the $V$-th camera is implicitly defined by the softmax (i.e. $w_V:=0$).

\begin{table}[ht]
\centering
\begin{tabular}{|c|c|c|c|c|c|c|}
\hline
Layer & k & s & chns in // out & in size  & out size& from \\
\hline
conv1 & 3 & 1 & $\tfrac{D}{G}\cdot 3V$ // 16   & 1x & 1 & PSV \\
conv2 & 3 & 2 & 16  // 32                      & 1x & 2x & conv1 \\
conv3 & 3 & 2 & 32 // 64                       & 2x & 4x & conv2 \\
conv4 & 3 & 2 & 64 // 128                      & 4x & 8x & conv3 \\
conv5 & 3 & 1 & 128 // 128                     & 8x & 8x & conv4 \\
conv6 & 3 & 1 & 128 // 256                     & 8x & 8x & conv5 \\
up1   & - & - & - // -                         & 8x & 4x & conv6 \\
conv6 & 3 & 1 & 320 // 64                      & 4x & 4x & up1 + conv3 \\
up2   & - & - & - // -                         & 4x & 2x & conv6 \\
conv7 & 3 & 1 & 96 / 32                        & 2x & 2x & up2 + conv2 \\
up3   & - & - & - // -                         & 2x & 1x & conv7 \\
conv8 & 3 & 1 & 48 // 16                       & 1x & 1x & up3 + conv1 \\
conv9 & 3 & 1 & 16 //                          & 1x & 1x & conv8 \\
      &   &   & $\tfrac{D}{G} \cdot (V+1)$ + 3 & & & \\
\hline
\end{tabular}
\caption{Table of Layer Parameters for the U-Net backbone of fMPI-S and fMPI-M. Up is simple bilinear upsampling. In and out size depict the (horizontal and vertical) downsampling factor.}
\label{table:u-net}
\end{table}

\subsection{Additional results}

\newcommand{\basicImgWidth}{0.2}
\begin{figure*}
    \centering
    \setlength{\tabcolsep}{1.2pt}
    \begin{tabular}{ c c c c c}
       \multicolumn{1}{c}{SIMPLI} & \multicolumn{1}{c}{DeepView} & \multicolumn{1}{c}{fMPI-S} & \multicolumn{1}{c}{fMPI-L} & \multicolumn{1}{c}{Ground truth} \\
        \includegraphics[width=\basicImgWidth\linewidth, trim=1 1 1 1, clip]{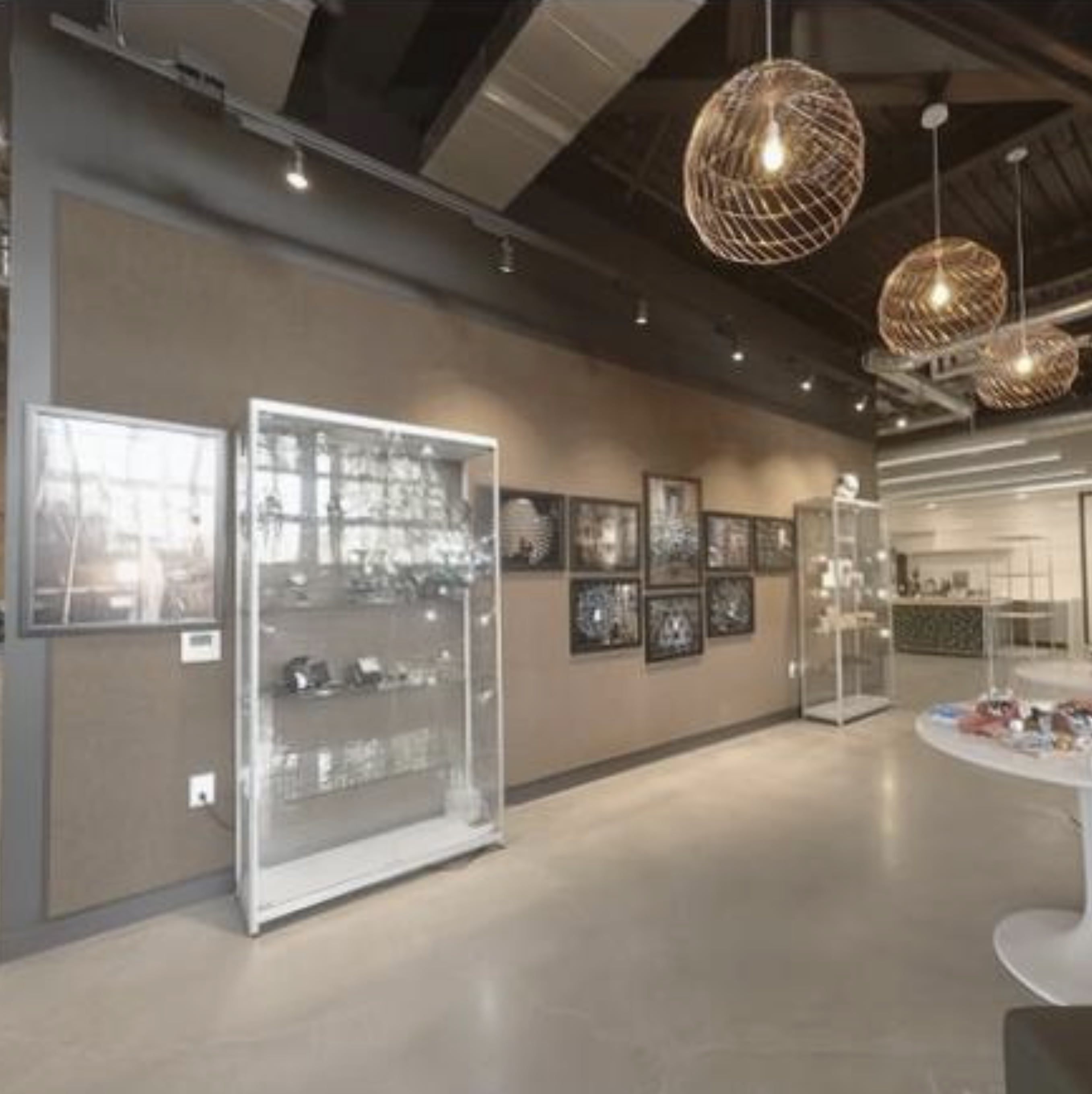} &
        \includegraphics[width=\basicImgWidth\linewidth, trim=1 1 1 1, clip]{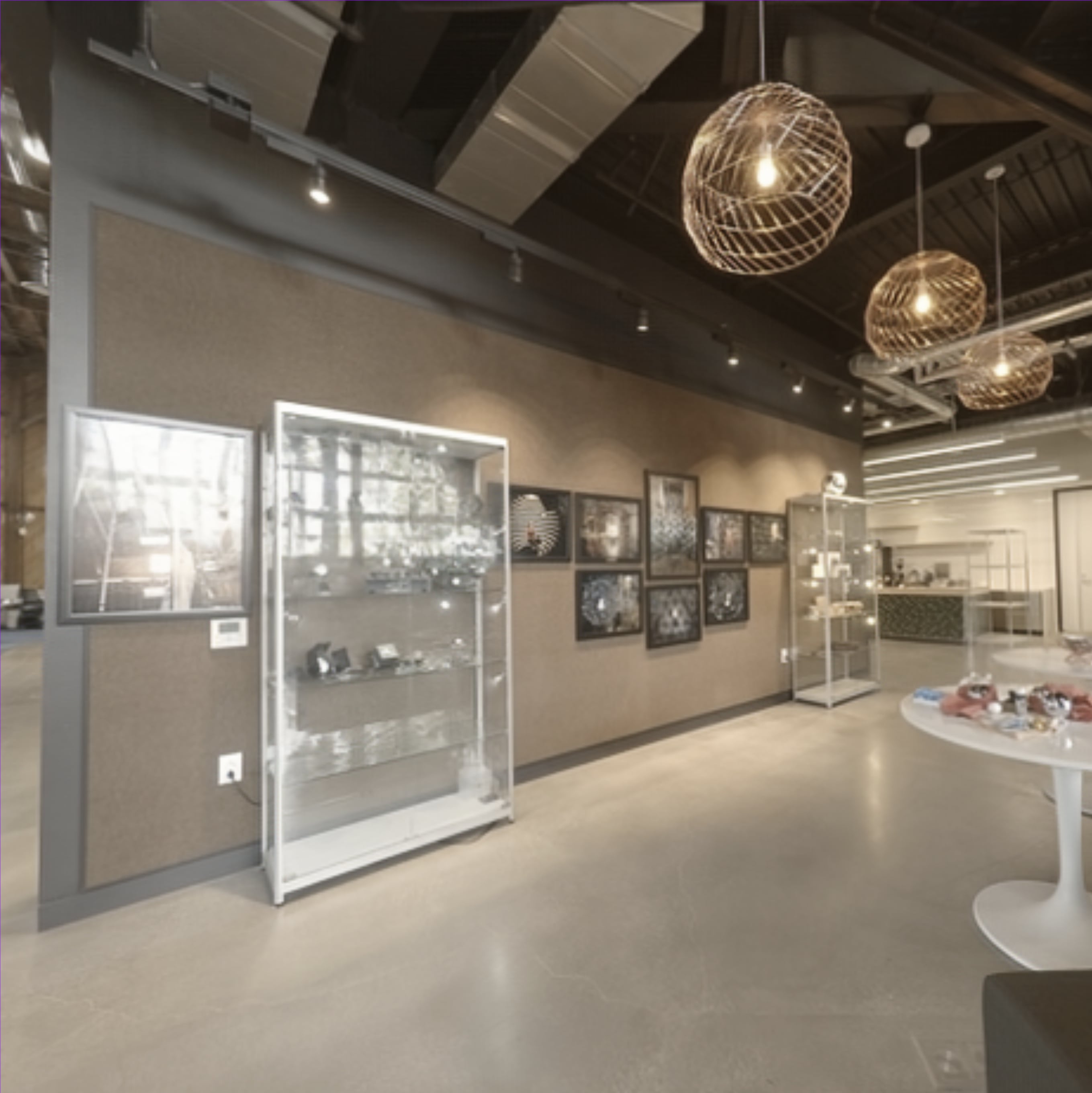} &
        \includegraphics[width=\basicImgWidth\linewidth, trim=1 1 1 1, clip]{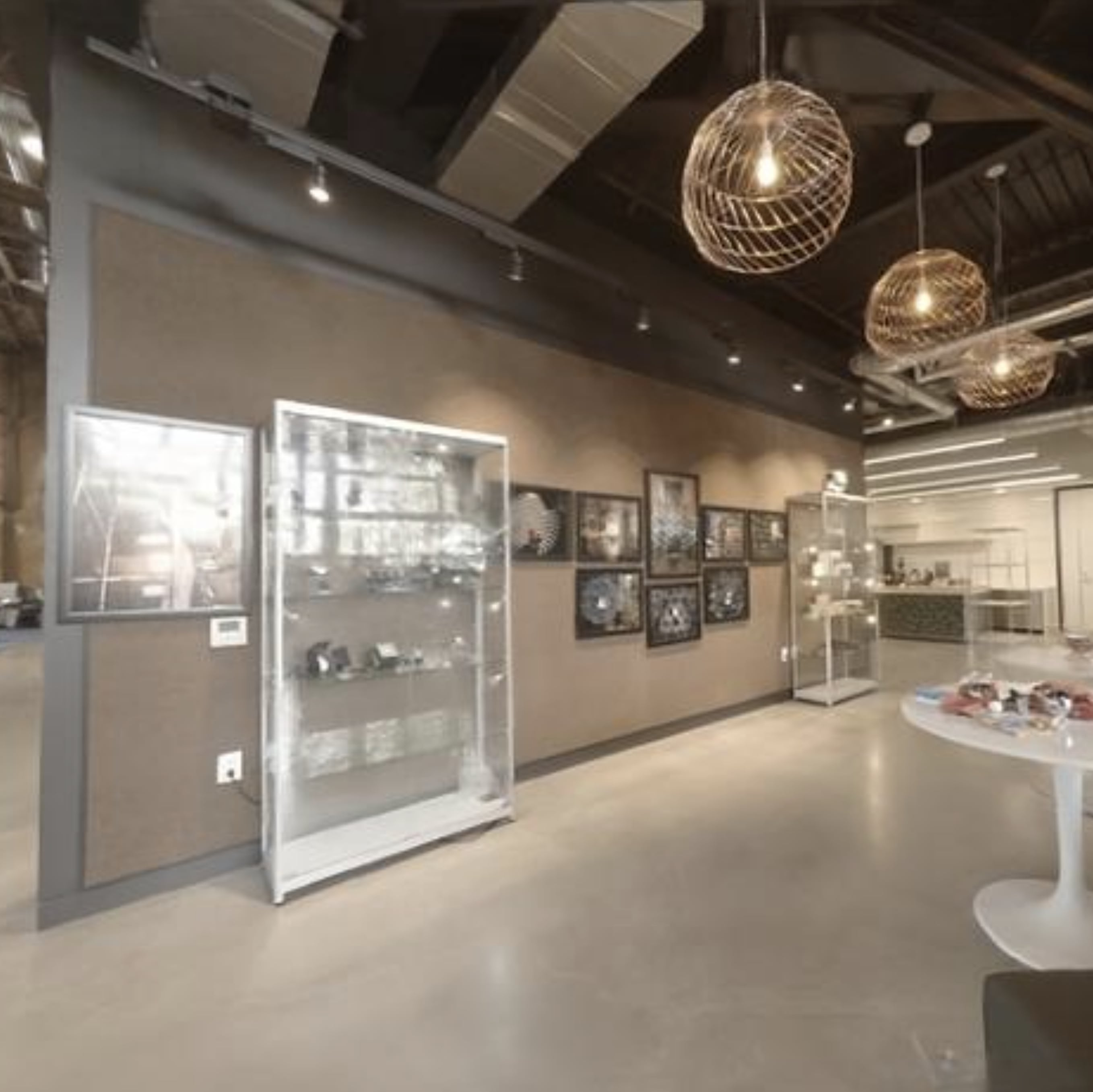} &
        \includegraphics[width=\basicImgWidth\linewidth, trim=1 1 1 1, clip]{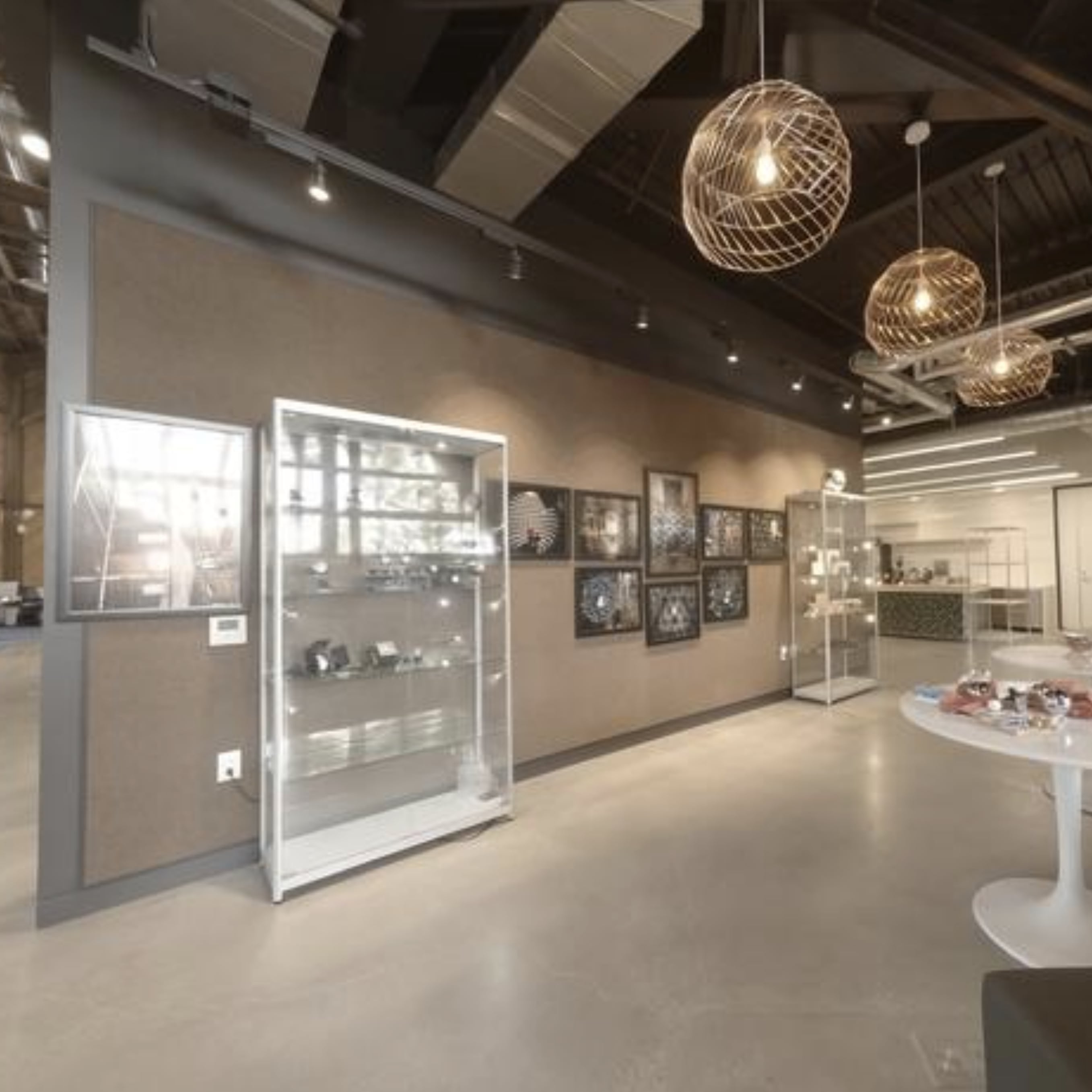} &
        \includegraphics[width=\basicImgWidth\linewidth, trim=1 1 1 1, clip]{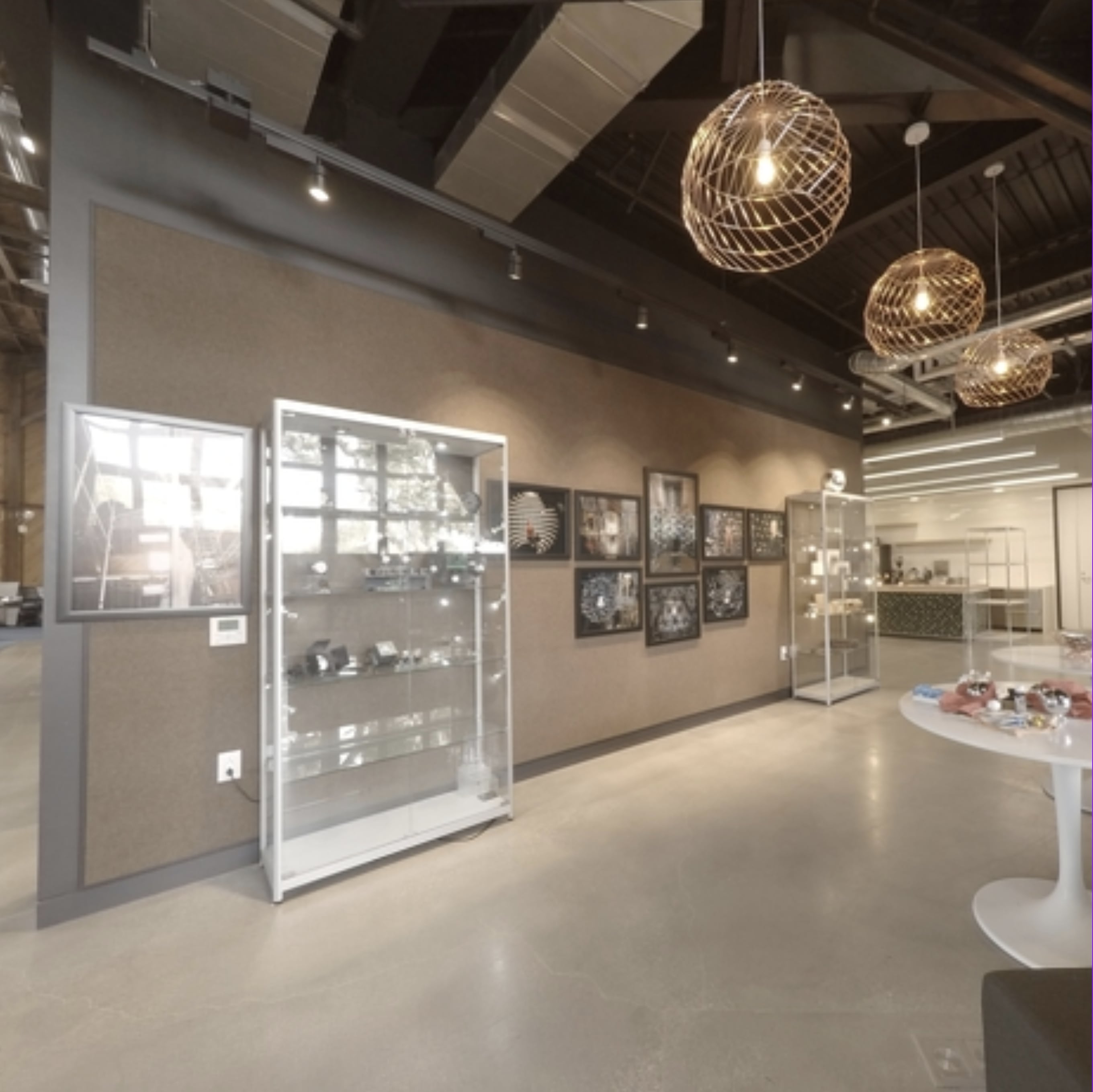}  \\
        \includegraphics[width=\basicImgWidth\linewidth, trim=1 1 1 1, clip]{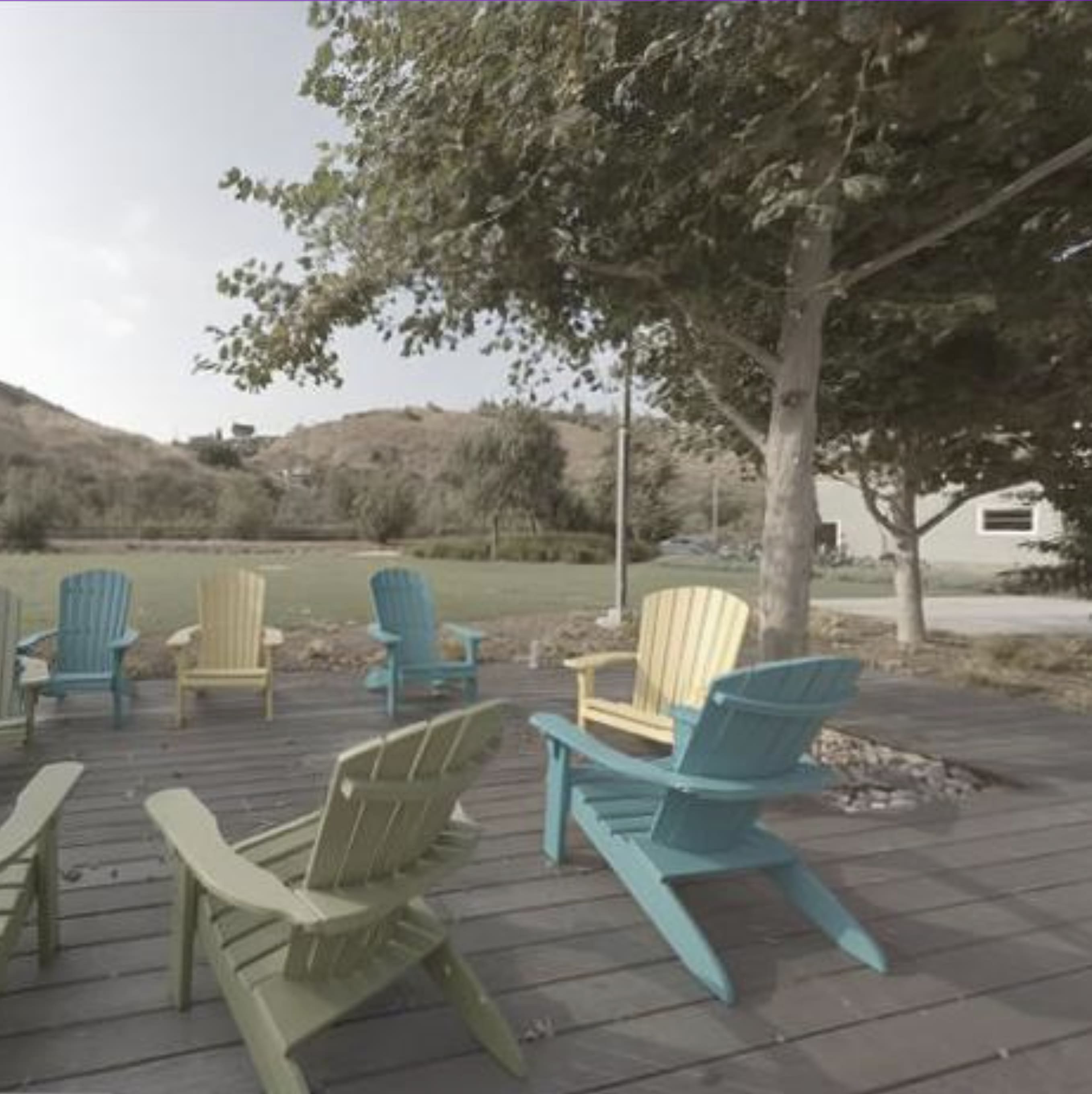} &
        \includegraphics[width=\basicImgWidth\linewidth, trim=1 1 1 1, clip]{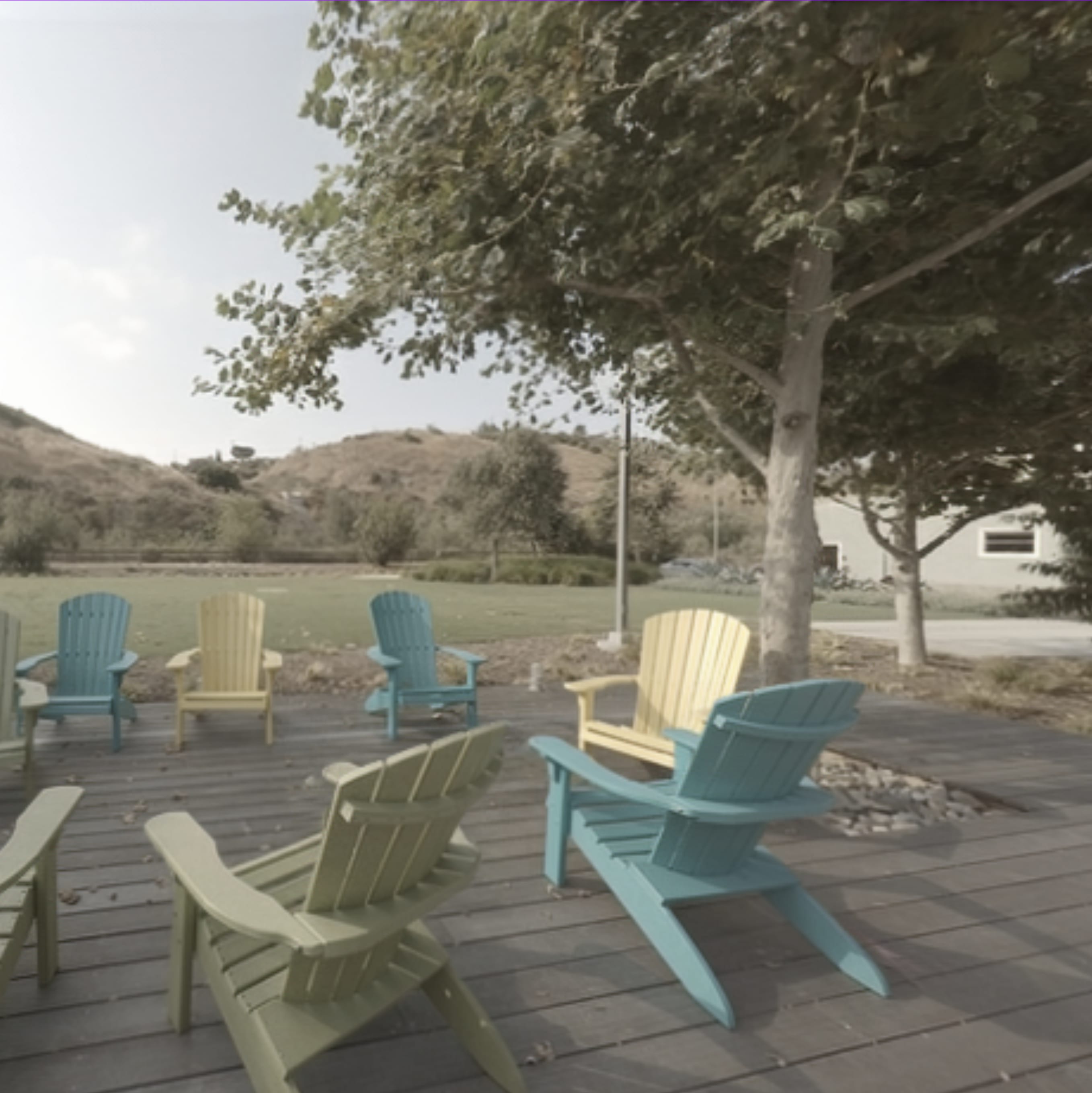} &
        \includegraphics[width=\basicImgWidth\linewidth, trim=1 1 1 1, clip]{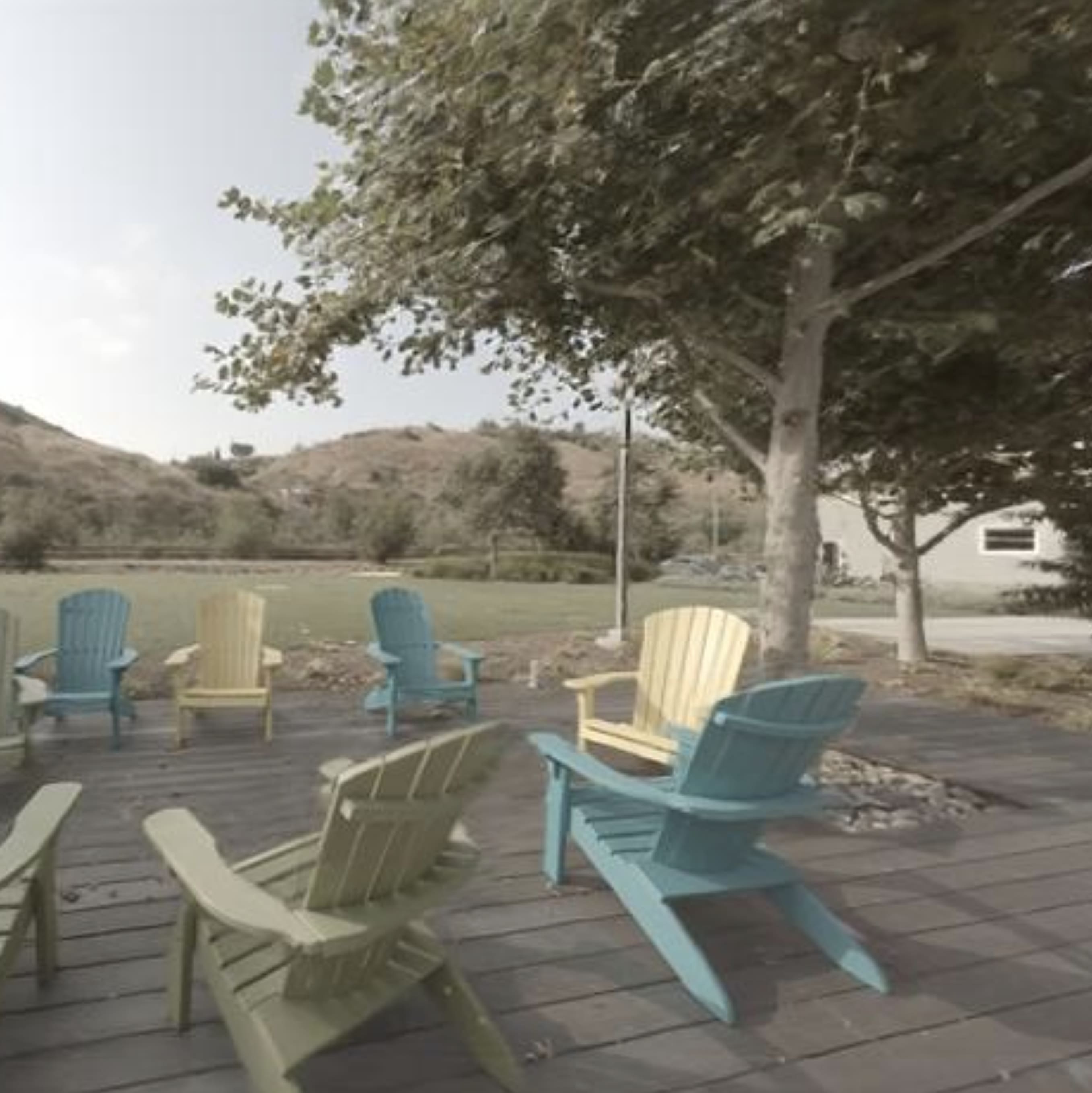} &
        \includegraphics[width=\basicImgWidth\linewidth, trim=1 1 1 1, clip]{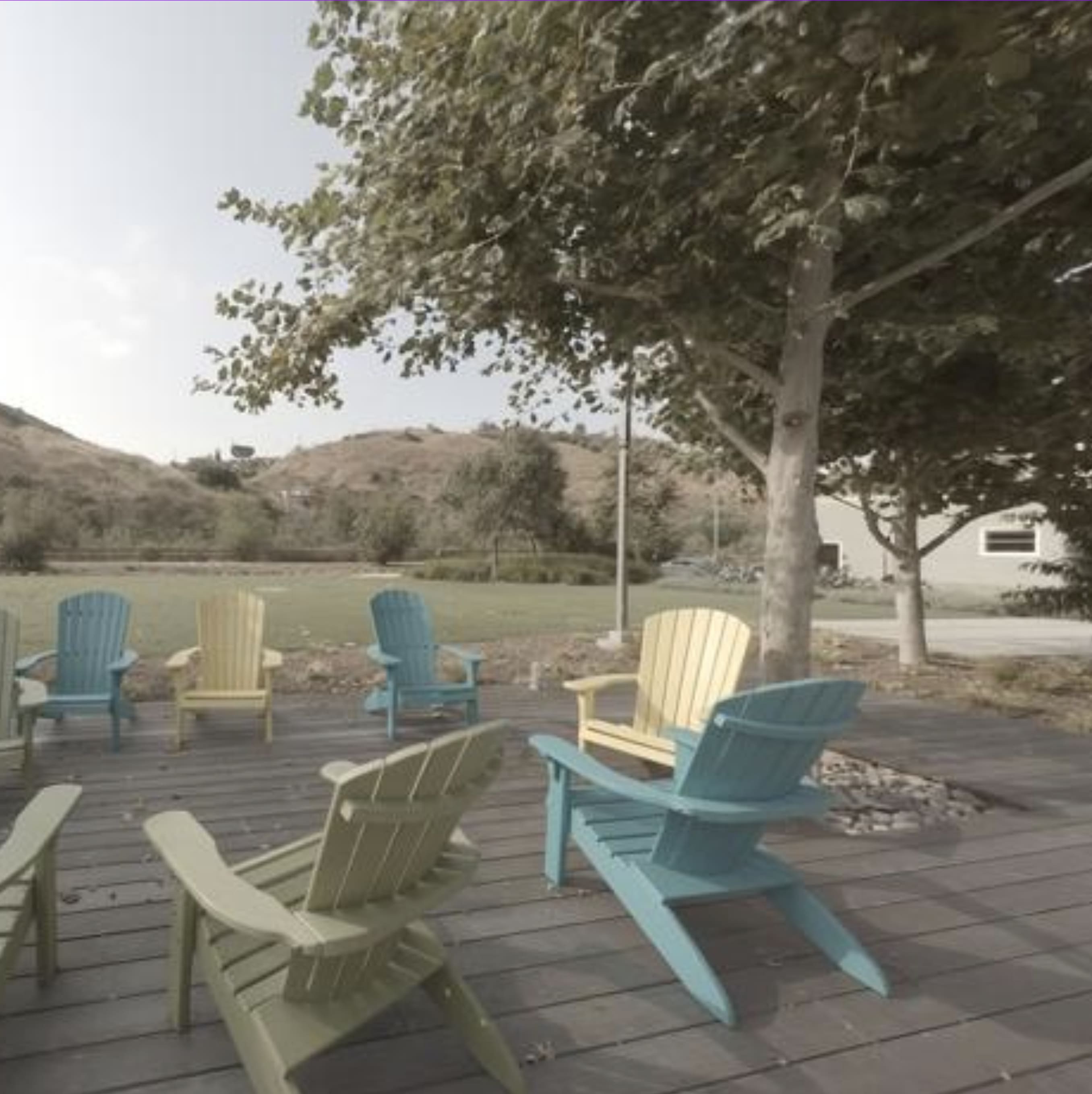} &
        \includegraphics[width=\basicImgWidth\linewidth, trim=1 1 1 1, clip]{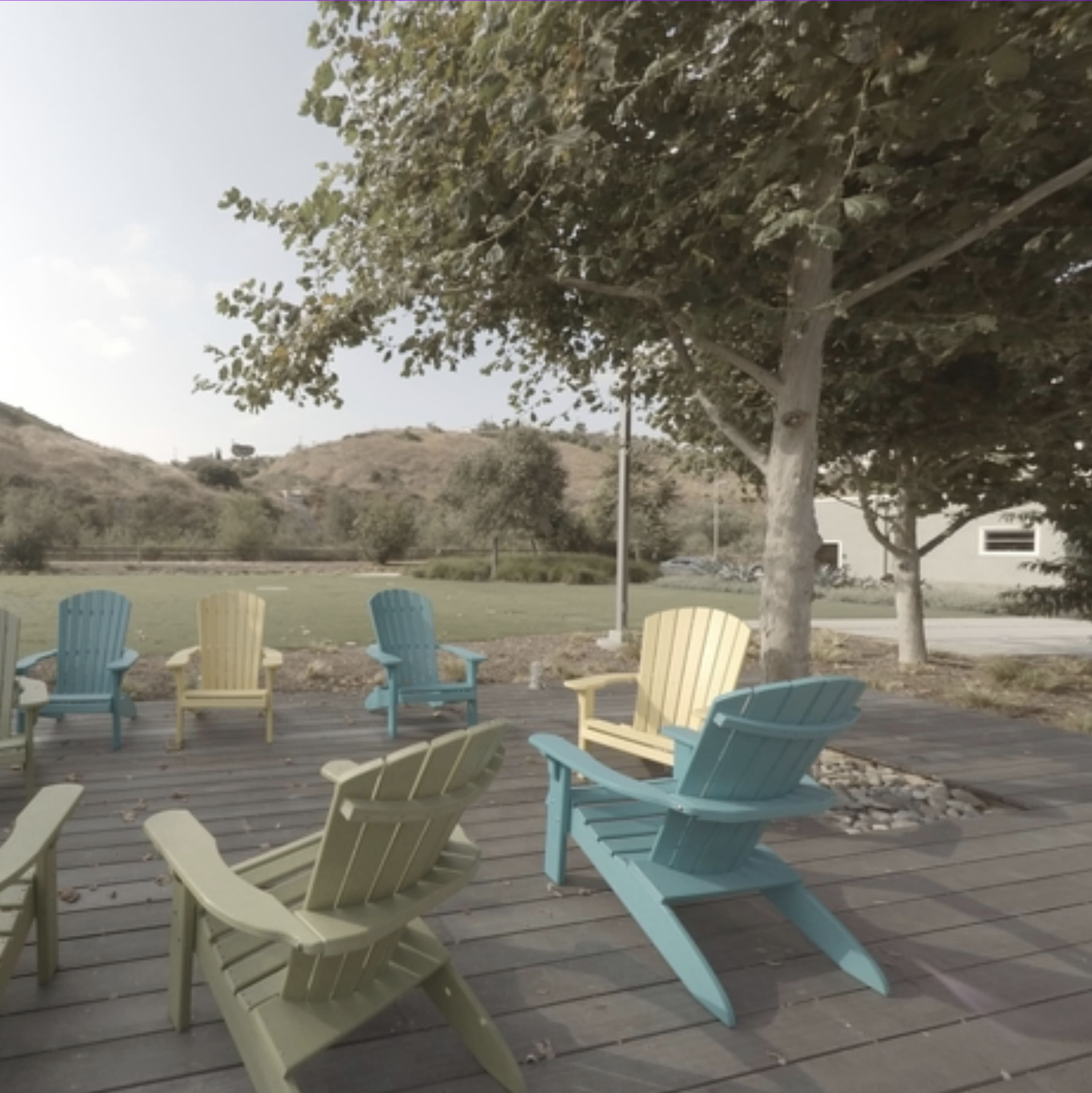}  \\
    \end{tabular}
    \caption{Additional results on spaces evaluation set. Note the sharpness of the images near the border between foreground and background objects.}
    \label{fig:appendix_spaces}
\end{figure*}

\begin{figure*}
    \centering
    \setlength{\tabcolsep}{1.2pt}
    \begin{tabular}{ c c c c}
       \multicolumn{1}{c}{fMPI-M} & \multicolumn{1}{c}{Ground truth} & \multicolumn{1}{c}{fMPI-M} & \multicolumn{1}{c}{Ground truth} \\
        \includegraphics[width=\basicImgWidth\linewidth, trim=1 1 1 1, clip]{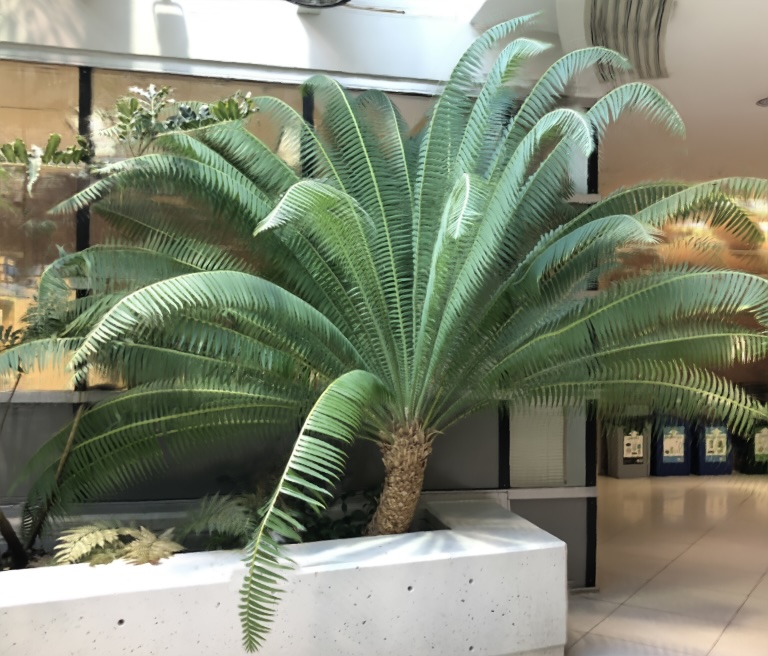} &
        \includegraphics[width=\basicImgWidth\linewidth, trim=1 1 1 1, clip]{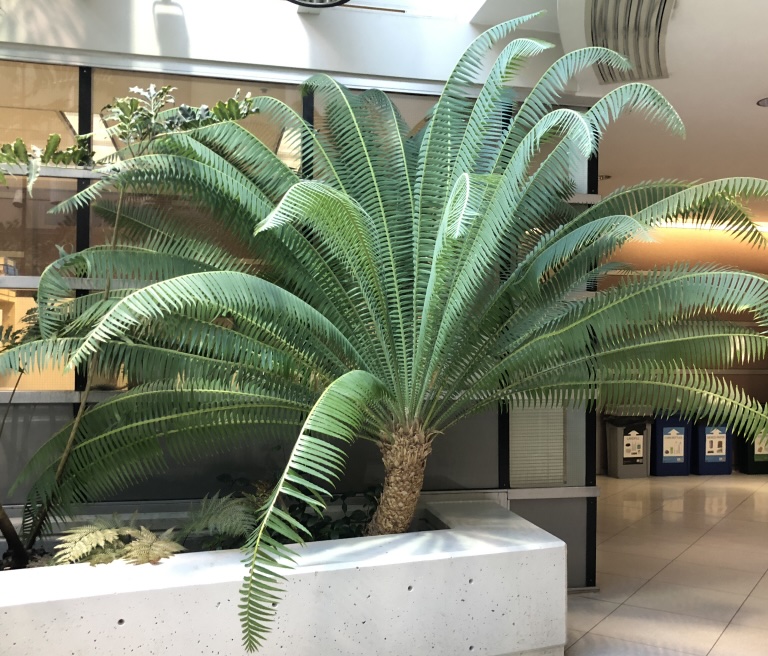} &
        \includegraphics[width=\basicImgWidth\linewidth, trim=1 1 1 1, clip]{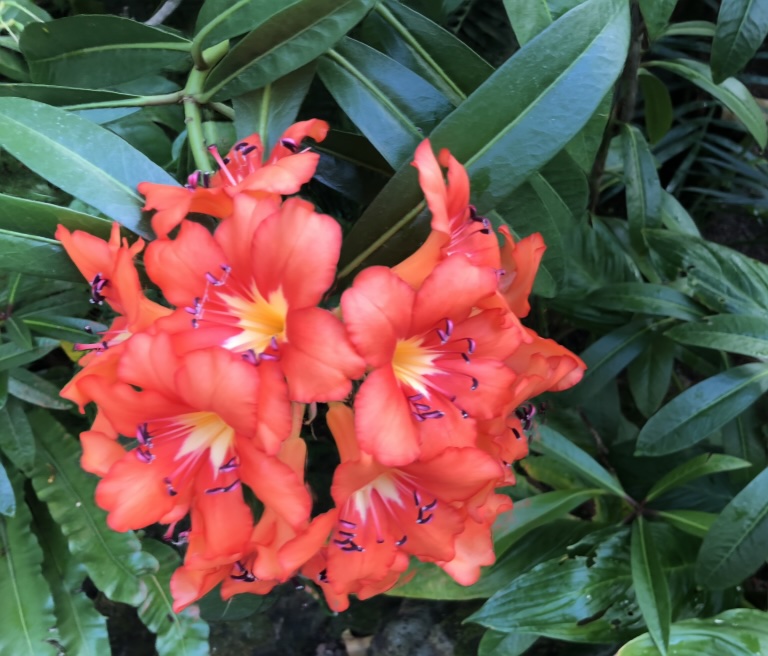} &
        \includegraphics[width=\basicImgWidth\linewidth, trim=1 1 1 1, clip]{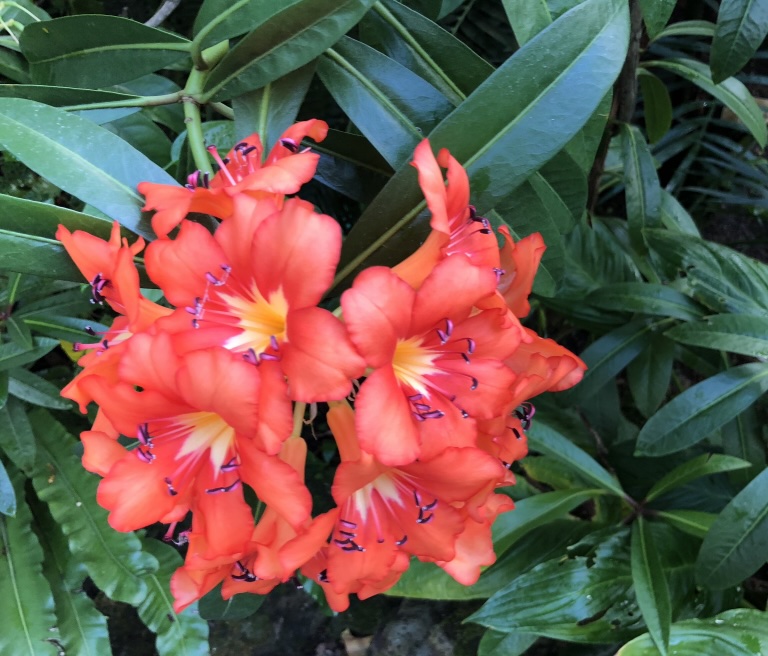} \\
        \includegraphics[width=\basicImgWidth\linewidth, trim=1 1 1 1, clip]{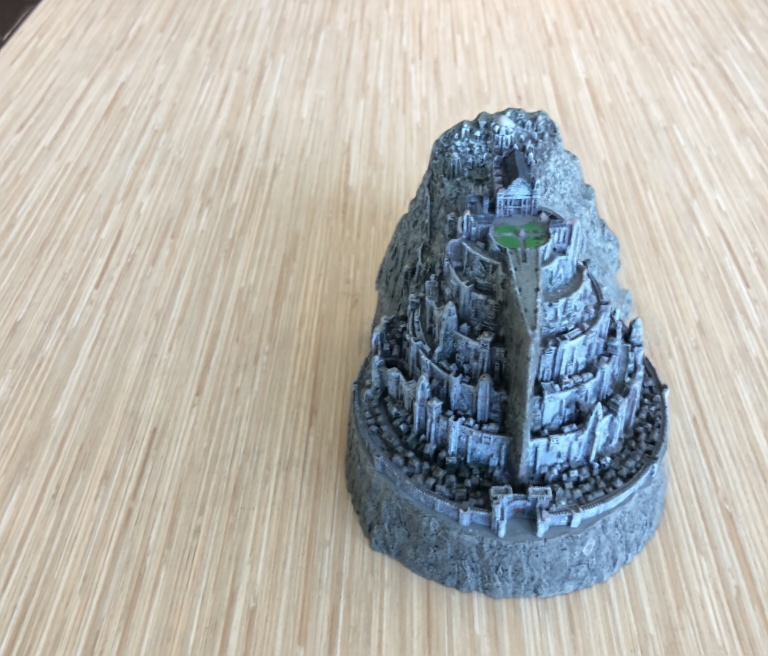} &
        \includegraphics[width=\basicImgWidth\linewidth, trim=1 1 1 1, clip]{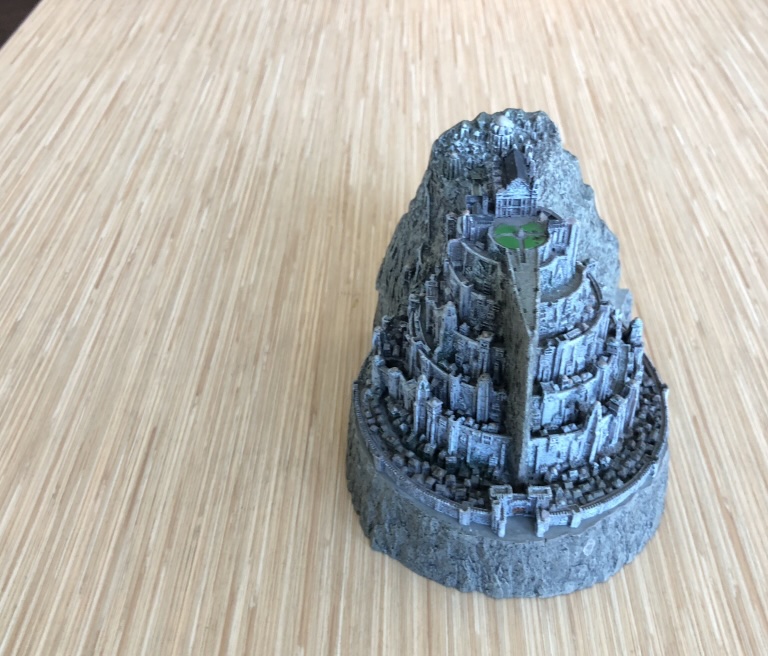} &
        \includegraphics[width=\basicImgWidth\linewidth, trim=1 1 1 1, clip]{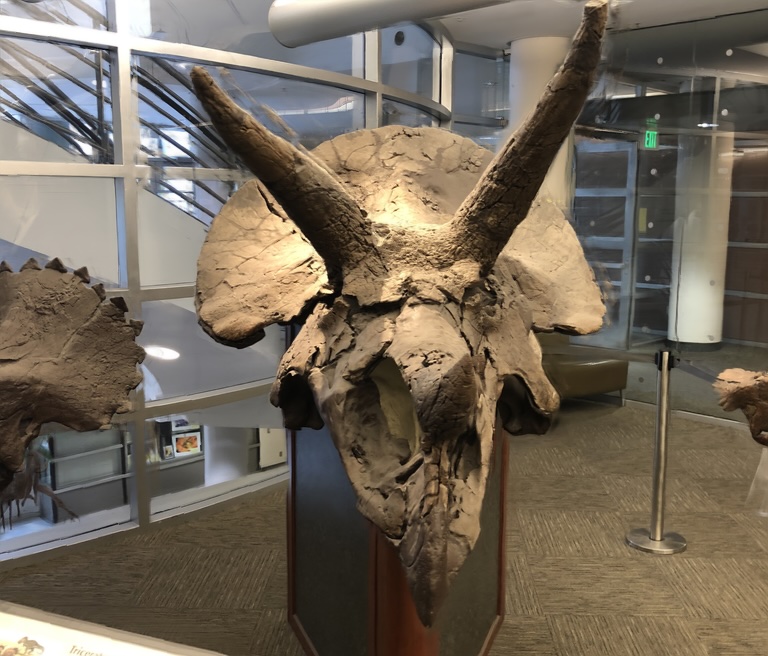} &
        \includegraphics[width=\basicImgWidth\linewidth, trim=1 1 1 1, clip]{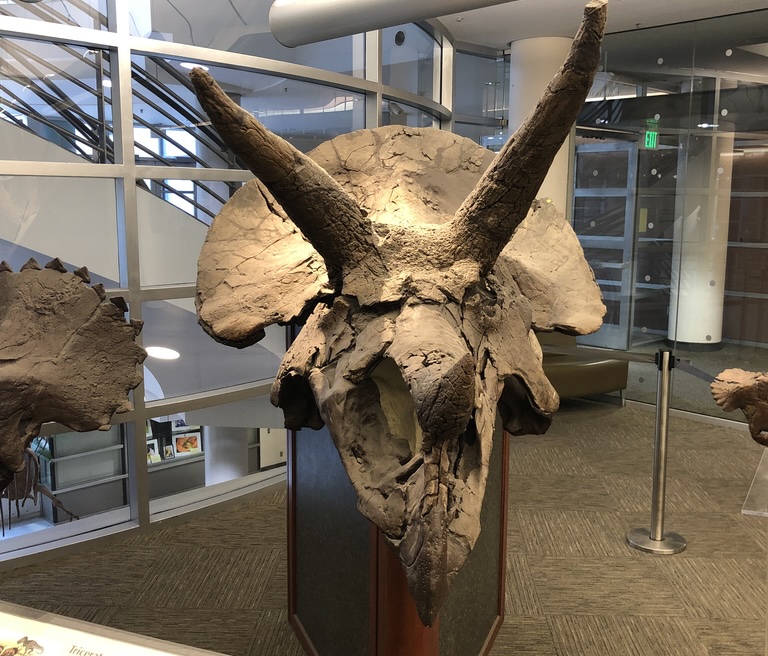} \\
        \includegraphics[width=\basicImgWidth\linewidth, trim=1 1 1 1, clip]{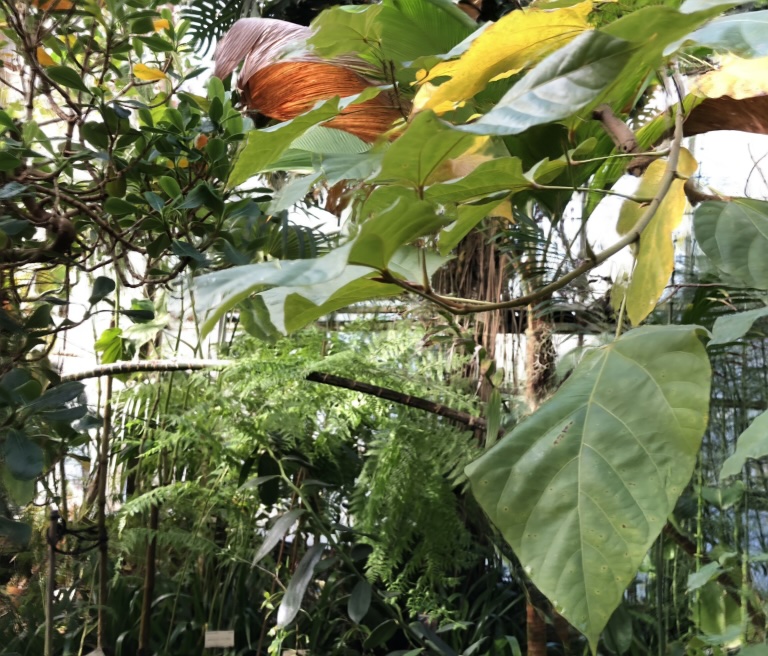} &
        \includegraphics[width=\basicImgWidth\linewidth, trim=1 1 1 1, clip]{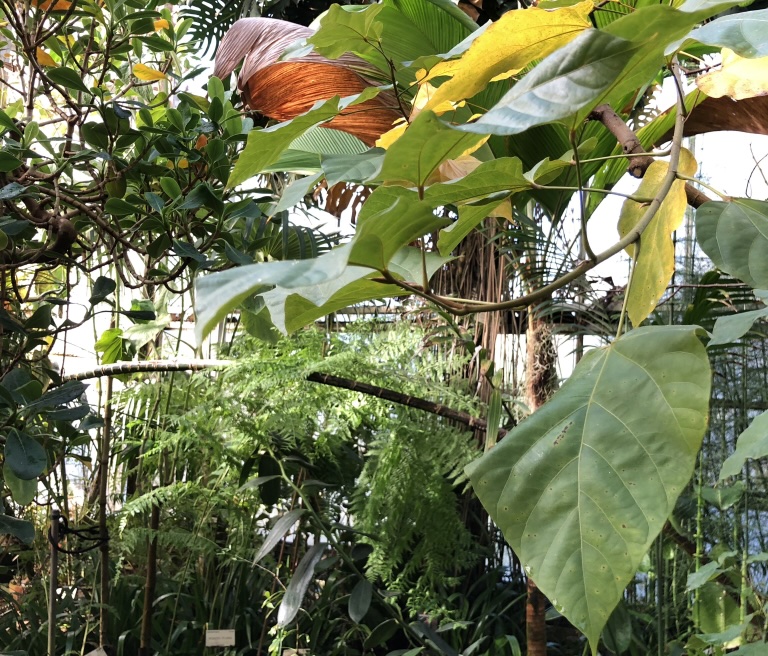} &
        \includegraphics[width=\basicImgWidth\linewidth, trim=1 1 1 1, clip]{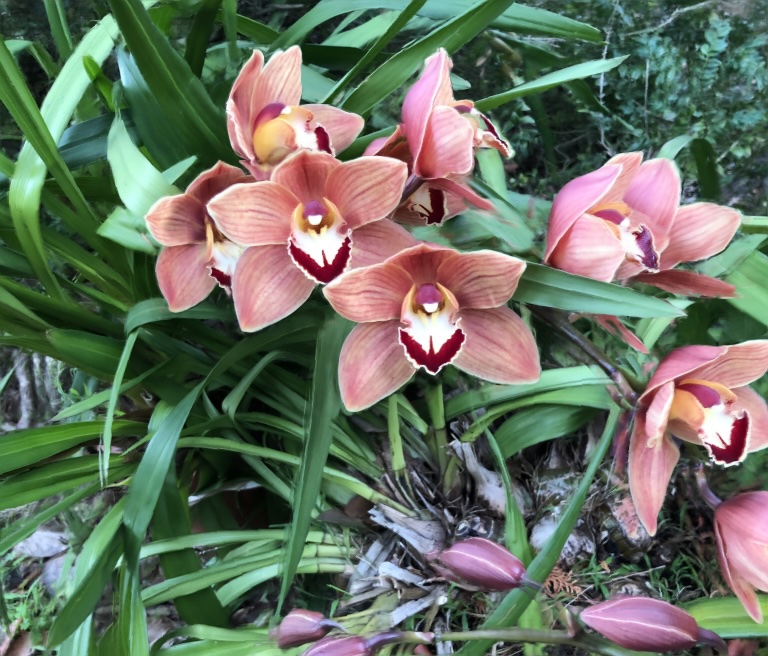} &
        \includegraphics[width=\basicImgWidth\linewidth, trim=1 1 1 1, clip]{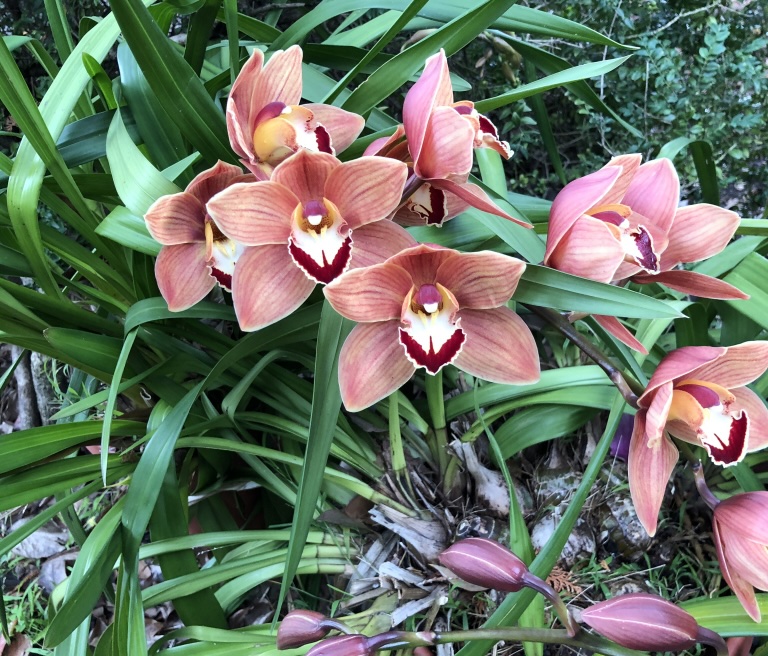} \\
        \includegraphics[width=\basicImgWidth\linewidth, trim=1 1 1 1, clip]{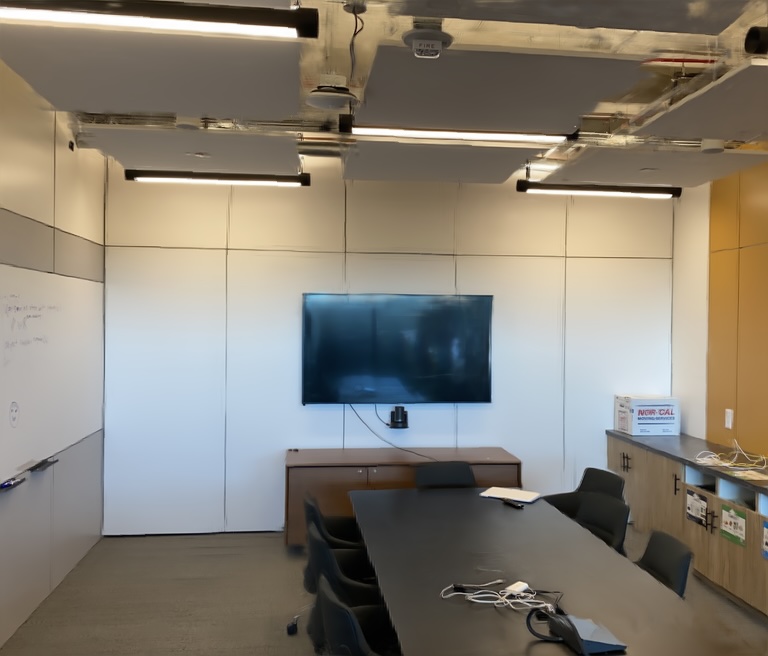} &
        \includegraphics[width=\basicImgWidth\linewidth, trim=1 1 1 1, clip]{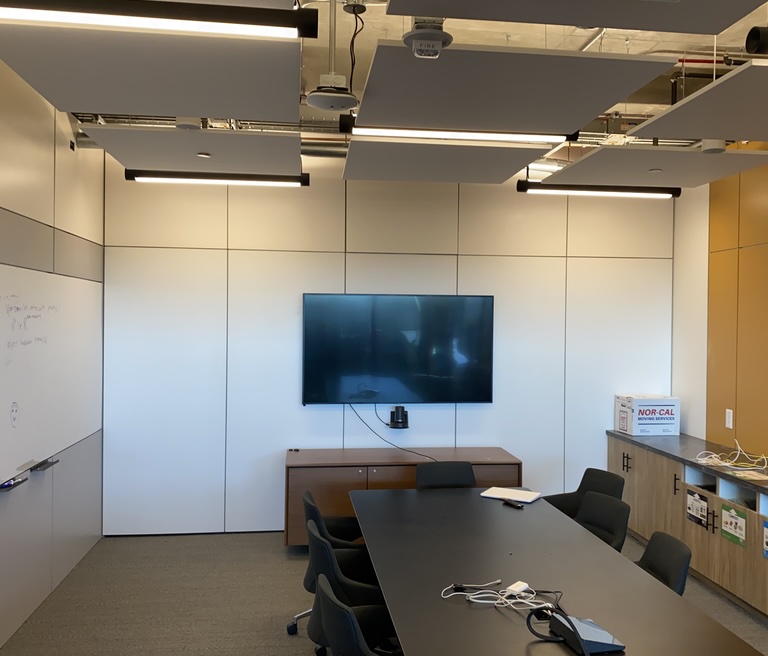} &
        \includegraphics[width=\basicImgWidth\linewidth, trim=1 1 1 1, clip]{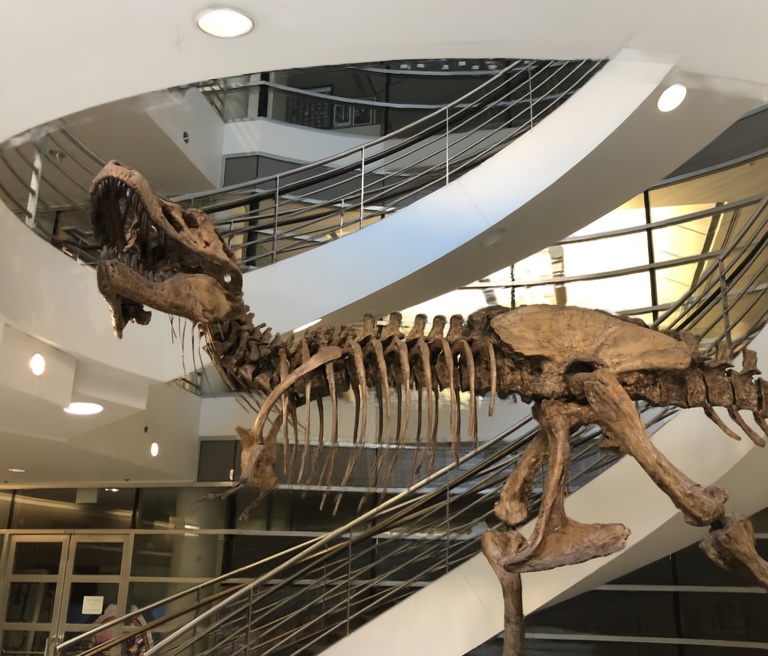} &
        \includegraphics[width=\basicImgWidth\linewidth, trim=1 1 1 1, clip]{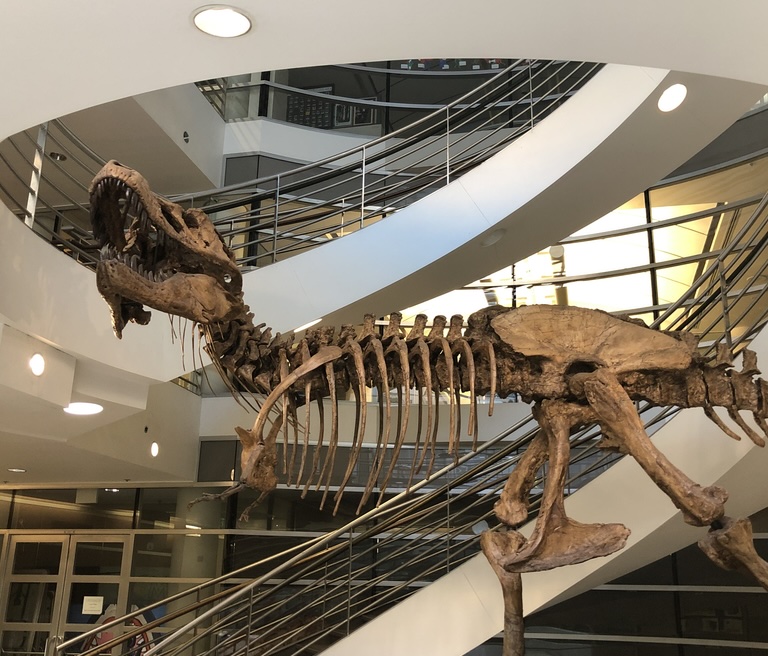} \\
    \end{tabular}
    \caption{Additional results on \textit{Real Forward Facing} evaluation set}
    \label{fig:appendix_rff}
\end{figure*}

\end{document}